\documentclass{article}

\usepackage{microtype}
\usepackage{graphicx}
\usepackage{mathtools} 
\usepackage{subcaption}
\usepackage{booktabs} 

\usepackage{hyperref}

\newcommand{\CNF}{\textsf{CNF}}
\newcommand{\DNF}{\textsf{DNF}}
\newcommand{\Rule}{\textsf{Rule}}

\usepackage[accepted]{icml2023}

\usepackage{amsmath}
\usepackage{amssymb}
\usepackage{mathtools}
\usepackage{amsthm}
\usepackage{hyperref}
\usepackage{url}
\usepackage{amsmath}
\usepackage{amssymb}
\usepackage{mathtools}
\usepackage{amsthm}
\usepackage{graphicx}
\usepackage{multirow}
\usepackage{microtype}
\usepackage{graphicx}
\usepackage{booktabs} 
\usepackage{graphicx}
\usepackage{capt-of}
\usepackage{caption}
\usepackage{booktabs}
\usepackage{varwidth}
\usepackage{svg}            
\usepackage{tikz}
\usepackage{comment}

\usepackage{amsmath,amssymb} 
\usepackage{color}
\usepackage{hyperref}   
\usepackage{booktabs}
\usepackage{multirow}
\usepackage{caption}
\usepackage{svg}  
\usepackage{wrapfig,lscape}
\usepackage{float}
\usepackage{rotating}
\usepackage{multicol}
\usepackage{listings}
\usepackage{tikz}
\usepackage{makecell}
\usepackage{comment}
\usetikzlibrary{calc}
\usetikzlibrary{decorations.pathreplacing}
\usetikzlibrary{tikzmark,decorations.pathreplacing,calligraphy}
\definecolor{jonquil}{rgb}{0.98, 0.85, 0.37}
\definecolor{babypink}{rgb}{0.96, 0.73, 0.73}
\definecolor{babypink2}{rgb}{0.96, 0.82, 0.82}
\definecolor{babyblueeyes}{rgb}{0.57, 0.73, 0.92}
\definecolor{babyblueeyes2}{rgb}{0.63, 0.79, 0.95}
\definecolor{darkseagreen}{rgb}{0.63, 0.83, 0.63}
\definecolor{darkseagreen2}{rgb}{0.63, 0.92, 0.63}
\usepackage[capitalize,noabbrev]{cleveref}

\theoremstyle{plain}

\theoremstyle{definition}

\theoremstyle{remark}

\usepackage{bm}

\usepackage[textsize=tiny]{todonotes}

\newcommand{\myparagraph}[1]{\vspace*{-0.265cm}\paragraph{#1}}
\newcommand{\myparagraphtitle}[1]{\vspace*{-0.05cm}\paragraph{#1}}
\newlength{\subseclen}
\icmltitlerunning{$\mathcal{T}$ruth $\mathcal{T}$able Net}

\begin{document}

\twocolumn[
\icmltitle{A Scalable, Interpretable, Verifiable \& Differentiable Logic Gate \\ Convolutional Neural Network Architecture From Truth Tables}



\icmlsetsymbol{equal}{*}

\begin{icmlauthorlist}
\icmlauthor{Adrien Benamira}{ntu}
\icmlauthor{Tristan Guérand}{ntu}
\icmlauthor{Thomas Peyrin}{ntu}
\icmlauthor{Trevor Yap}{ntu}
\icmlauthor{Bryan Hooi}{nus}
\end{icmlauthorlist}

\icmlaffiliation{ntu}{Nanyang Technological University, Singapore}
\icmlaffiliation{nus}{National University of Singapore, Singapore}

\icmlcorrespondingauthor{Adrien Benamira}{adrien002@e.ntu.edu.sg}

\icmlkeywords{Machine Learning, ICML}

\vskip 0.3in
]



\printAffiliationsAndNotice{}  

\begin{abstract}

We propose $\mathcal{T}$ruth $\mathcal{T}$able net ($\mathcal{TT}$net), a novel Convolutional Neural Network (CNN) architecture that addresses, by design, the open challenges of interpretability, formal verification, and logic gate conversion. $\mathcal{TT}$net is built using CNNs' filters that are equivalent to tractable truth tables and that we call Learning Truth Table (LTT) blocks. The dual form of LTT blocks allows the truth tables to be easily trained with gradient descent and makes these CNNs easy to interpret, verify and infer. Specifically, $\mathcal{TT}$net is a deep CNN model that can be automatically represented, after post-training transformation, as a sum of Boolean decision trees, or as a sum of Disjunctive/Conjunctive Normal Form (DNF/CNF) formulas, or as a compact Boolean logic circuit. We demonstrate the effectiveness and scalability of $\mathcal{TT}$net on multiple datasets, showing comparable interpretability to decision trees, fast complete/sound formal verification, and scalable logic gate representation, all compared to state-of-the-art methods. We believe this work represents a step towards making CNNs more transparent and trustworthy for real-world critical applications.

\end{abstract}

\section{Introduction}\label{sec:intro}

Deep Convolutional Neural Networks (DCNNs) have been widely and successfully applied in various tasks in the field of machine learning. However, despite their achievements, there are still issues that need to be addressed. In this paper, we focus on three major challenges of DCNNs: 1) Global and exact interpretability, 2) Complete and sound formal verification, 3) Differentiable logic gates CNNs.

\vspace*{\subseclen}
\subsection{Related works}

\myparagraphtitle{Global and exact interpretability.} Interpretability is crucial for understanding the decision-making process of the trained model, particularly in security-sensitive applications~\cite{osoba2017intelligence}. Decision trees are widely used in scenarios requiring high interpretability~\cite{bertsimas2017optimal} as they produce all the rules, i.e. the global nature of interpretability, that permit the exact computation of the output model result, i.e. the exact nature of interpretability. In recent years, there has been a growing interest in developing DNNs that are as globally and exactly interpretable as rule-based models~\cite{wang2021scalable, yang2021learning}, including a NeurIPS 2021 spotlight paper~\cite{agarwal2020neural}. To date, there is no family of convolutional DNNs that are globally and exactly interpretable as decision trees by design, without the need for an explainer (which often offers local and inexact explanations~\cite{ribeiro2016should, ribeiro2018anchors}).

\myparagraph{Complete and sound formal verification.} Formal verification is a critical aspect in ensuring the soundness and completeness of neural network properties, particularly for safety-critical applications~\cite{driscoll2020system}. Conventional verifiers work with real-valued networks, but they face scalability challenges e.g. hundreds of seconds to verify an MNIST image property~\cite{muller2022prima} and provide no guarantees of correctness due to floating point errors~\cite{jia2021exploiting}. Therefore, developing an efficient DNNs verification process is an important topic, as evidenced by the $\alpha-\beta$-Crown paper~\cite{xu2020fast, wang2021beta} and the winner of the Verification Neural Network competition~\cite{bak2021second}. In this work, instead of designing a new general verification method for DCNNs, we propose the first architecture that can be fast and correctly verified with any SAT verification solver~\cite{roussel2009pseudo}. This research direction is attracting interest, as exemplified by~\cite{jia2020efficient}, who strives to show that Binary Neural Networks (BNNs~\cite{hubara2016binarized}) are faster to verify if a specific SAT solver is designed, e.g. less than a second to verify an MNIST image property.

\myparagraph{Differentiable logic gates CNNs.} Logic gates DNN representation is essential for deploying compact DNNs on resource-constrained devices~\cite{wang2019lutnet}. These logic gate networks consist of logic ``AND'' and ``OR'' gates, denoted $\wedge$ and $\vee$ respectively~\cite{darwiche2002knowledge}, which allow very fast execution. However, the challenge in learning logic gate networks is that they are typically non-differentiable, making them difficult to train with gradient descent~\cite{rumelhart1986learning}. To date, there is no family of differentiable convolutional DNNs logic gates, and as stated in~\cite{petersen2022deep}: ``Convolutional logic gate networks [...] are interesting and important directions for future research.''

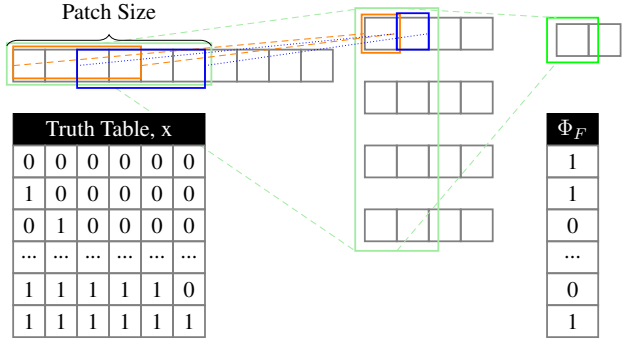
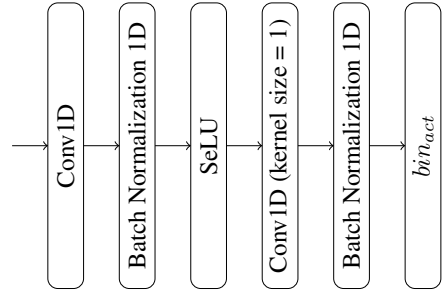
\begin{figure*}[]
    \centering
\begin{subfigure}[t]{0.48\textwidth}
\scalebox{0.85}{
\centering
\begin{tikzpicture}
\foreach \x in {0,...,9}
    \draw[gray, thick] (0 + \x*0.5,0.5) rectangle (0.5 + \x*0.5,1);
    
\foreach \y in {0,...,3}
    \foreach \x in {0,...,3}
        \draw[gray, thick] (5.5 + \x*0.5,1 - \y) rectangle (6+ \x*0.5,1.5- \y);

\foreach \x in {0,1}
    \draw[gray,thick] (8.5+\x*0.5,0.9) rectangle (9+\x*0.5,1.4); 

\draw[darkseagreen2, thick] (5.35 + 0*0.5,1-3.15) rectangle (6.65,1.65);

\draw[green, thick] (8.35+0*0.5,1.5) rectangle (9.15+0*0.5,0.8);
\draw[darkseagreen2, densely dashed] (5.5 + 1*0.5,1.65+0.0) -- (8.5+0*0.5,0.6+0.9);
\draw[darkseagreen2, densely dashed] (5.5 + 1*0.5,1-3.15) -- (8.5+0*0.5,-0.1+0.9);
\draw[darkseagreen2, densely dashed] (-0.1+ 3*0.5,0.6+0.5) -- (5.5,1.65);
\draw[darkseagreen2, densely dashed] (0.1 + 3*0.5,-0.1+0.5) -- (5.35+0*0.5,1-3.15);
\draw[darkseagreen2, thick] (-0.1,0.65-0.2) rectangle (0.1 + 6*0.5,1.1);

\draw [decorate,decoration={brace,amplitude=5pt,raise=1ex}]
  (-0.1,0.9) -- (0.1 + 6*0.5,0.9) node[midway,yshift= 2em]{Patch Size};
 
\draw[orange, thick] (0 + 0*0.5,0.55) rectangle (0.5 + 3*0.5,1.05);
\draw[orange, thick] (5.45 + 0*0.5,1 - 0.05) rectangle (6.05+ 0*0.5,1.5+ 0.05);
\draw[orange, densely dashed] (0,0.25+0.5) -- (5.5,1.25);
\draw[orange, densely dashed] (0+4*0.5,0.25+0.5) -- (5.5+0.5,1.25);

\draw[blue, thick] (1 + 0*0.5,0.5-0.1) rectangle (1.5 + 3*0.5,1.0);
\draw[blue, thick] (6 + 0*0.5,1 - 0) rectangle (6.5+ 0*0.5,1.5+0.07);
\draw[blue, densely dotted] (1,0.25+0.5) -- (6,1.25);
\draw[blue, densely dotted] (1+4*0.5,0.25+0.5) -- (6+0.5,1.25);

\draw[black, fill=black] (0, -0.5) rectangle (0 + 6*0.5,0) node[pos=0.5,text=white] {Truth Table, x};

\foreach \x in {0,...,5}
    \draw[gray, thick] (0 + \x*0.5,-0.5-1*0.5) rectangle (0.5 + \x*0.5,-0.-1*0.5) node[pos = 0.5, black] {0};
\foreach \x in {1,...,5}    
    \draw[gray, thick] (0 + \x*0.5,-0.5-2*0.5) rectangle (0.5 + \x*0.5,-0.-2*0.5) node[pos = 0.5, black] {0};
\draw[gray, thick] (0 + 0*0.5,-0.5-2*0.5) rectangle (0.5 + 0*0.5,-0.-2*0.5) node[pos = 0.5, black] {1};    
\draw[gray, thick] (0 + 0*0.5,-0.5-3*0.5) rectangle (0.5 + 0*0.5,-0.-3*0.5) node[pos = 0.5, black] {0};
\foreach \x in {2,...,5}    
    \draw[gray, thick] (0 + \x*0.5,-0.5-3*0.5) rectangle (0.5 + \x*0.5,-0.-3*0.5) node[pos = 0.5, black] {0};
\draw[gray, thick] (0 + 1*0.5,-0.5-3*0.5) rectangle (0.5 + 1*0.5,-0.-3*0.5) node[pos = 0.5, black] {1};    
\foreach \y in {4}
\foreach \x in {0,...,5}
    \draw[gray, thick] (0 + \x*0.5,-0.5-\y*0.5) rectangle (0.5 + \x*0.5,-0.-\y*0.5) node[pos = 0.5, black] {...};
\foreach \y in {5,6}
    \foreach \x in {0,...,4}
        \draw[gray, thick] (0 + \x*0.5,-0.5-\y*0.5) rectangle (0.5 + \x*0.5,-0.-\y*0.5) node[pos = 0.5, black] {1};
\draw[gray, thick] (0 + 5*0.5,-0.5-5*0.5) rectangle (0.5 + 5*0.5,-0.-5*0.5) node[pos = 0.5, black] {0};
\draw[gray, thick] (0 + 5*0.5,-0.5-6*0.5) rectangle (0.5 + 5*0.5,-0.-6*0.5) node[pos = 0.5, black] {1};

\draw[black, fill=black] (8.35, -0.5) rectangle (9.15,0.0) node[pos=0.5,text=white] {$\Phi_F$};
\draw[gray, thick] (8.35,-0.5-1*0.5) rectangle (9.15,-0-1*0.5) node[pos = 0.5, black] {1};
\draw[gray, thick] (8.35,-0.5-2*0.5) rectangle (9.15,-0.-2*0.5) node[pos = 0.5, black] {1};
\draw[gray, thick] (8.35,-0.5-3*0.5) rectangle (9.15,-0.-3*0.5) node[pos = 0.5, black] {0};
\draw[gray, thick] (8.35,-0.5-4*0.5) rectangle (9.15,-0.-4*0.5) node[pos = 0.5, black] {...};
\draw[gray, thick] (8.35,-0.5-5*0.5) rectangle (9.15,-0.-5*0.5) node[pos = 0.5, black] {0};
\draw[gray, thick] (8.35,-0.5-6*0.5) rectangle (9.15,-0.-6*0.5) node[pos = 0.5, black] {1};
\end{tikzpicture} }
    \caption{Converting $\Phi_F$ into a truth table. The above example has two layers: the first one has parameters (input channel, output channel, kernel size, stride) = $(1,4,4,2)$, while the second $(4,1,2,2)$. The patch size of $\Phi_F$ is $6$ (i.e. \textcolor{darkseagreen2}{green} box) since the output feature (i.e. \textcolor{green}{light green} box) requires $6$ input entries (\textcolor{orange}{orange} and \textcolor{blue}{blue} boxes).}
    \label{fig:TT_compute}
\end{subfigure}
\hspace{0.4cm}
\begin{subfigure}[t]{0.48\textwidth}
\scalebox{0.95}{
\hspace{1cm}
\begin{tikzpicture}
\foreach \x in {0,...,5} {
    \draw[rounded corners] (0+ \x, 0) rectangle (0.5+\x, 4);
    \draw[->] (0.5+ \x -1, 2) -- (1+\x-1, 2); }
\path (0, 0) -- node[rotate=90,anchor=center] {Conv1D}(0.5, 4);
\path (0+1, 0) -- node[rotate=90,anchor=center] {Batch Normalization 1D}(0.5 + 1, 4);
\path (0+2, 0) -- node[rotate=90,anchor=center] {SeLU}(0.5+2, 4);
\path (0+3, 0) -- node[rotate=90,anchor=center] {Conv1D (kernel size = 1)}(0.5+3, 4);
\path (0+4, 0) -- node[rotate=90,anchor=center] {Batch Normalization 1D}(0.5+4, 4);
\path (0+5, 0) -- node[rotate=90,anchor=center] {$bin_{act}$}(0.5+5, 4);
\end{tikzpicture}
}
    \caption{Architecture description of one family of LTT block in 1 dimension: Expanding AutoEncoder LTT (E-AE LTT). Such a computation is showcased in Figure~\ref{fig:TT_compute}. LTT block inputs are always binary.}
    \label{fig:E-AE_LTT_1D}
\end{subfigure}
\caption{A Learning Truth Table (LTT) block. The intermediate values and weights are floating points, input/output values are binary.  We hope that LLT block can bridge the gap between logic symbolic Boolean AI and deep learning.}
\end{figure*}

\vspace{-0.2cm}

\myparagraph{Overall strategy.} Among all the families of possible DCNN models, some are easier to manipulate than others. A perfect example is BNNs~\cite{hubara2016binarized}, which by binarizing the architecture, manage to find a friendly DCNN architecture that makes a lot of open problems easier~\cite{jia2020efficient}. In this paper, we propose to follow the same strategy. \textit{We address the three open challenges presented above by introducing the first DCNN family that is specifically designed to be (i) as interpretable as a sum of Boolean decision trees, (ii) amenable to fast complete/sound verification using any SAT solver, (iii) easily differentiable with standard machine learning framework and capable of compact logic gate CNN representation, finally (iv) scalable to large datasets.}

\vspace*{\subseclen}
\subsection{Our work}

\myparagraphtitle{Our contributions.} In this paper, we introduce an architecture, $\mathcal{TT}$net, that addresses these open questions by providing a CNN filter model that is transformable into a compact truth table representation by design. $\mathcal{TT}$net utilizes a novel building block called Learning Truth Table (LTT) block, which combines the dual form of tractable truth tables and CNNs, allowing for truth tables to be easily trained and optimized with standard machine learning libraries. The LTT blocks enable $\mathcal{TT}$net to be represented as a sum of truth tables, or a sum of DNF/CNF formulas, or sum of Boolean decision trees, or as a compact Boolean logic gate circuit, and the conversion is automatic after training. We hope that LLT block can bridge the gap between logic symbolic Boolean AI and Deep Learning~\cite{marques2022logic, shi2020tractable, yu2021learning, choi2017compiling}


To address global and exact interpretability, $\mathcal{TT}$net is designed to provide a clear and concise representation of the decision-making process of the network in the form of a sum of Boolean decisions trees~\footnote{Linear models, shallow decision trees and $\mathcal{TT}$net are interpretable only if the features they are trained on are interpretable.} \textbf{(claim 1A)}. We also describe an automatic process for human knowledge injection \textbf{(claim 1B)}, a unique feature that is essential for critical applications.

To address sound \& complete formal verification, $\mathcal{TT}$net is designed to support fast property verification using generic SAT solvers. This allows for the formal verification of the network's properties in dozens of milliseconds, which is about $10^3$ times faster than $\alpha-\beta$-Crown~\cite{xu2020fast,wang2021beta}, the state-of-the-art DNN verification solver and the winner of the VNN competition~\cite{bak2021second}. \textbf{(claim 2A)}. We also provide a comparison with BNNs verification~\cite{jia2020efficient, narodytska2019search}. We demonstrate that our method is sligthly more accurate in low noise cases and faster in all scenarios, whereas BNNs require a specific SAT solver, we do not. \textbf{(claim 2B)}.

To address CNN differentiable logic gate representation, $\mathcal{TT}$net is built upon CNN-based blocks, LTT blocks, that are transformable into logic gate circuits - as desired by~\cite{petersen2022deep}. Our results demonstrate more compact and accurate model on tabular datasets than~\cite{petersen2022deep}, while being as compact as the \cite{petersen2022deep} on MNIST, with a slightly lower accuracy and a shorter training time \textbf{(claim 3A)}. Additionally, we show that our approach is more scalable in terms of accuracy and training time, achieving 70.2\% on CIFAR-10 with less than an hour of training. It represents an 8\% improvement and 270 times faster training than~\cite{petersen2022deep} approach, at the cost of being less compact \textbf{(claim 3B)}.

Our \textbf{(claim 4)} is that we demonstrate $\mathcal{TT}$net can achieve ImageNet classification task as previous works such as XnorNet~\cite{rastegari2016xnor} and the BNN original paper~\cite{hubara2016binarized}.

\myparagraph{Outline.} In Section~\ref{sec:Background}, we provide a comprehensive overview of the relevant background in the field of Boolean logic and formal verification. In Section~\ref{sec:TTmodel}, we introduce the design principles and architecture of $\mathcal{TT}$net model. In Section~\ref{sec:Results}, we provide a thorough evaluation of the performance of our model on various datasets, demonstrating its effectiveness in addressing the three open challenges highlighted above. We also discuss the opportunities for further research and we conclude in Section~\ref{sec:Conclusion}. A glossary can be find in Appendix~\ref{glossary}.

\section{Background}\label{sec:Background}

\myparagraphtitle{Boolean logic.} A truth table is a mathematical representation of the output of a Boolean function for all possible input combinations. It can therefore be obtained when the corresponding Boolean function is tractable. Note that the Boolean function of a BNN is generally not tractable as the input size of the function is too large. A truth table can be expressed in Conjunctive Normal Form (CNF) or Disjunctive Normal Form (DNF)\footnote{A CNF is a conjunction of a set of clauses: $\Phi=(c_1 \land \cdots \land c_m)$, where each clause $c_j$ is a disjunction of some literals $c_j = l_{j1} \lor \cdots \lor l_{jr}$. A DNF is a disjunction of a set of clauses: $\Phi=(c_1 \lor \cdots \lor c_m)$, where each clause $c_j$ is a conjunction of some literals $c_j = l_{j1} \land \cdots \land l_{jr}$.}, both being standard forms to represent a Boolean function, especially in formal verification. Each DNF equation becomes a rule if it is evaluated to a space position: see \textbf{Example 1B} in Section~\ref{subsubsec:1D_1channel}. A Boolean circuit is a physical or mathematical representation of a Boolean function, made up of interconnected logic gates that implement the rules~\cite{arora2009computational, klir1997fuzzy}. 

\myparagraph{Formal verification.} Complete and sound property verification of SAT-convertible DNNs has been presented in \cite{narodytska2018verifying}, where a precondition is placed on the inputs, a property is placed on the outputs, and a CNF relation is defined between inputs and outputs. An application example of property verification is to check for the existence of an adversarial perturbation in a trained DNN. In this case, the precondition defines an $\epsilon$-ball of valid perturbations around the original image, and the property states that the classification should not change under these small perturbations. Thus, we distinguish the traditional ``natural accuracy'' from the ``verified accuracy'', the latter measuring the fraction of predictions that remain correct for all adversarial attacks within the perturbation constraints. We give formal definitions in Appendix~\ref{sec:ANN_Background}, examples in Appendix~\ref{ANN:example}.


\section{$\mathcal{T}$ruth $\mathcal{T}$able deep convolutional neural network}\label{sec:TTmodel}

In this section, we introduce $\mathcal{TT}$net. We also provide \href{https://youtu.be/loGlpVcy0AI}{a companion video} that illustrates our design principles. Appendix~\ref{sec:ANN_TTnet} offers complementary details on the architecture.

\vspace*{\subseclen}
\subsection{General}\label{subsubsec:TTmodel_general}
\vspace*{\subseclen}

We propose a DCNN architecture for security applications. We argue that while recent developments in DNN architecture have focused on improving performance~\cite{vaswani2017attention,brown2020language,dosovitskiy2020image}, the resulting models have become increasingly complex and difficult to verify, interpret and implement~\cite{hoefler2021sparsity}. To address this issue, we focus on CNNs, which are widely used in the field, and we try to transform them into tractable Boolean functions.

\myparagraph{CNN filter as a tractable Boolean function.} The conversion property of the floating CNN weights filter into a binary truth table is achieved by transforming the CNN filter function into a tractable Boolean function. To accomplish this, the complexity of the CNN filter function is reduced by: (A) decreasing the input size of the CNN filter (noted as $n$ in the rest of the paper), (B) using binary inputs and (C) using binary outputs. (A) is achieved by decreasing the number of connections between convolution layers and (B-C) by utilizing the Heaviside step function\footnote{ $bin_{act}(x) =\frac{1}{2} + \frac{sgn(x)}{2}$, and to train with this function without much loss in performance we adopted the Straight-Through Estimator (STE) proposed by~\cite{hubara2016binarized}.}, denoted as $bin_{act}$, to binarize the features. It is worth noting that our model differs from BNNs in that we preserve real-valued weights. The combination of (A-B-C) allows for a tractable Boolean function, as all input-output pairs can be computed in $2^n$ operations (for a not-too-large $n$). The resulting function is then stored as a truth table, an example of which is shown in Figure~\ref{fig:TT_compute}. Additionally, in Section~\ref{subsubsec:nonlin}, we introduce a condition (D): the CNN filter must be non-linear before the Heaviside step function to achieve satisfactory accuracy.

\myparagraph{Unique property of  $\mathcal{TT}$net: trainable truth table, logic circuit \& CNF/DNF.} The conversion of a floating-weighted CNN filter function into a tractable Boolean function enables its formulation as a truth table, referred to as the Learning Truth Table (LTT). The resulting neural network is named the $\mathcal{T}$ruth $\mathcal{T}$able Deep Convolutional Neural Network ( $\mathcal{TT}$net), as it utilizes LTTs as small truth tables in place of floating-weighted CNN filters. It is important to note that one LTT block can be expressed in multiple forms: as a neural network, a truth table, a Boolean function, and a logic circuit in CNF/DNF form. This means that the LTT has the unique property of being both a differentiable neural network that can be trained using standard optimization techniques and a tractable Boolean function that can be expressed as a truth table, CNF/DNF or a logic gate circuit. As a result, CNNs have the benefits of interpretability and compactness of Boolean functions; at the same time, the Boolean functions benefit from the differentiability and the training feature of CNNs. We also highlight that this transformation is automatic and occurs after training. 


Next, we build the non-linear LTT block architecture step by step for 1D-CNN, which can be generalized to 2D-CNN.

\vspace*{\subseclen}
\subsection{Construction of non-linear LTT block} \label{subsec:Construction_LTT}

\subsubsection{1D-convolution, one input channel} 
\label{subsubsec:1D_1channel}
We denote as $\Phi_F$ a 1D-convolution associated with the filter $F$ of kernel size $ker = n$, stride $s$, and no padding. Let the input feature with a single input channel $chn_{input}=1$ be represented as $v_0 \dots v_{N-1}$ where $N$ is the length of the input feature. We define $y_{i, F}$, the output of the function $\Phi_F$ at position $i$ and $y_{bin, i, F} = bin_{act}(y_{i, F}) = bin_{act}(\Phi_F(v_{i \times s}, v_{i \times s+1}, \dots, v_{i \times s + (n-1)}))$. If $v_{i\times s+ q}$ are binary values for all $q = 0, \dots, n-1$, we can express the 1D-convolutional layer $\Phi_F$ as a truth table by enumerating all $2^n$ possible input combinations. The truth table can then be converted into a simplified CNF/DNF formula using the Quine–McCluskey algorithm~\cite{blake1938} for interpretation. However, there is a computational limitation when using this algorithm because it is solving an NP-complete problem~\cite{1634621}. Therefore, we shall limit ourselves to $n \leq 16$ for this paper. In that case, as $\Phi_F$ fulfilled conditions (A-B-C), $\Phi_F$ is a linear LTT.

\myparagraph{Example 1A: from CNN filter to truth table to CNF/DNF.} We consider a 1D-convolution with one filter of kernel size $ker = 4$, a stride of size $2$ and no padding. Let the weights of the 1D-convolutional layer be $\bm{W_1} = (10, -1, 3, -5).$ We define the following CNF/DNF literals [$x_0$ \: $x_1$ \: $x_2$ \:$x_3$]. As the inputs and outputs are binary and the number of input entries for the CNN layer is $4$ (kernel size $= 4$ and number of input channel $chn_{input}=1$), we have $2^4=16$ possible entries: [$0$ \: $0$ \: $0$ \: $0$], [$0$ \: $0$ \: $0$ \: $1$], $\cdots$, [$1$ \: $1$ \: $1$ \: $1$]. For each input, we calculate the corresponding output resulting from the convolution of $\bm{W_1}$ with the $16$ possible literal entries: $[0, -5, 3, -2, -1, -5, 3, -2, 10, 5, 13, 8, 9, 4, 12, 7]$. After binarization with the Heaviside step function, we have $y_{binary} = [0, 0, 1, 0, 0, 0, 1, 0, 1, 1, 1, 1, 1, 1, 1, 1]$. So $\Phi_F$ is the function that maps $\{0,1\}^n$ to the corresponding $y_{binary}$: this is the truth table form and we do not need $W_1$ anymore. We then transform $\Phi_F$ into DNF/CNF by applying the Quine-McCluskey algorithm~\cite{blake1938} which gives $DNF= (x_2 \wedge ~x_3) \vee x_0 $, $CNF= ( x_2 \vee x_0) \wedge (~x_3 \vee x_0) $ (simplified DNF/CNF).

In the previous paragraph, we described the general procedure for transforming a CNN filter into LTT into a CNF/DNF logic gate expression. This expression is independent of the spatial position of the feature, commonly referred to as the patch\footnote{Patch is the region of the input that produces the feature, which is commonly referred to as the receptive field~\cite{araujo2019computing}.}. When we apply the LTT DNF expression to a specific spatial position on the input, we convert the DNF into a rule. To convert the general DNF form into a set of rules $\mathcal{R}$, we divide the input into patches and replace the DNF literals with the corresponding feature names. The number of rules for one filter corresponds to the number of patches. An example of this process is provided below.

\myparagraph{Example 1B: conversion of DNF expressions to rules.} To obtain the rules, we need to consider the padding and the stride of the LTT block. Consider the following 6-feature binary input: [Age $>34$, Male, Go to University, Married, Born in US, Born in France]. In our case, with a stride at 2 and no padding, we get 2 patches: [Age$>34$, Male, Go to University, Married] and [Go to University, Married, Born in the US, Born in France]. After the replacement of the literal by the corresponding feature, we get 2 rules: $\Rule_{1}^{\DNF} = (\text{Go to University} \land \overline{ \text{Married}}) \lor (\text{Age}>34) $ and $\Rule_{2}^{\DNF} = (\text{Born in the US} \land \overline{\text{Born in France}}) \lor \text{Go to University} $. We underline the logic redundancy in $\Rule_{2}^{\DNF}$: if someone is born in the US, he/she is necessarily not born in France. We solve this issue by injecting human knowledge into the truth table as we will see in Section~\ref{subsec:HK}.

\subsubsection{1D-convolution, multiple input channels} 

In general, a 1D-convolution takes in several channels as input (\textit{i.e.} $chn_{input}>1$) and therefore the number of input variables of $\Phi_F$ will rise substantially. For example, a 1D-CNN that takes $32$ input channels with a kernel size of $4$ yields an input of size $128$, which exceeds our limit of $n=16$. In order to overcome this, we group the channels using the group parameter \cite{dumoulin2016guide}. Grouped convolutions divide the input channels into $g$ groups, then apply separate convolutions within each group; this effectively decreases the number of inputs of $\Phi_F$ to each individual filter by a factor of $g$. In that case, we have $ n = ker \times chn_{input}/g$ where $chn_{input}$ is the number of input channels and $ker$ is the kernel size. 

\myparagraph{Example 2: the grouping parameter.} We consider a 1D-CNN layer with one filter of kernel size $ker = 5$, a stride of size $s = 2$ and no padding. Let the 1D-CNN weights be \scalebox{0.8}{$\bm{W_1} = \begin{bmatrix}  10 & -1 & 3 & -5 & 6 \\ 8 & -2 & 5 & -1 & 6 \\ -3 & 0 & 2 & 4 & 6 \\ -10 & 1 & -3 & 5 & 6 \end{bmatrix}$} for the $4$-channel input. We can observe that $\bm{W_1}$ requires $20 =5 \times 4$ binary inputs and leads to a truth table of size $2^{20}$, which is too large according to our criteria of $2^{16}$. Therefore, by defining a group $g=2$, then $W_1$ becomes $2$ filters matrices: $\bm{W_{1,1}} = \begin{bmatrix}  10 & -1 & 3 & -5 & 6 \\ 8 & -2 & 5 & -1 & 6\end{bmatrix}$ for filter $1$ which takes as input only the first two input channels and $\bm{W_{1,2}} = \begin{bmatrix}  -3 & 0 & 2 & 4 & 6 \\ -10 & 1 & -3 & 5 & 6 \end{bmatrix}$ for filter $2$ which takes as input only the last two input channels. Therefore, thanks to this grouping, each filter has an input size of $n = 10 = 5 \times 4/2$ which fits our criteria of $n \leq 16$.

\subsubsection{Non linear LTT for multi-input channels}\label{subsubsec:nonlin} 
In all of the above examples, the LTT $\Phi_F$ was a linear function before the step function. In general, if $\Phi_F$ is a 1D-convolution, the $\mathcal{TT}$net will not be able to learn complex tasks such as image classification~\cite{imagenet_cvpr09} (see Appendix~\ref{subsection:App_Ablation} that supports this claim). Inspired by the work~\cite{brendel2019approximating, dosovitskiy2020image}, we therefore decided to transform the linear function $\Phi_F$ into a non-linear one to increase the learning capacity of the neural network. The idea is to use a DNN architecture for $\Phi_F$ without increasing the input size (and so the patch of the function) of $\Phi_F$. The $\Phi_F$ input size is now defined as $n = pc/g$ where $p$ is the patch size instead of the kernel size $ker$. 

\myparagraph{Non-linear LTT block example: Expanding-AutoEncoder (E-AE LTT).} Among all the families of possible LTT blocks, we represent in Figure~\ref{fig:E-AE_LTT_1D} an Expanding-AutoEncoder LTT block (E-AE LTT): $\Phi_F$, with one input channel and $g = 1$ with a patch size of $6$. $F$ now refers to the $F^{th}$ output channel of the last layer for DNN $\Phi_F$. The E-AE LTT block is built upon a layer we call the amplification layer. The amplification layer works by simply adding a new convolution layer with kernel size $1$ after a convolution layer. Doing so will not increase the patch size. Overall, the LTT block comprises two 1D-convolution layers, denoted as Conv1D, with an amplification parameter $\tau$, which is the ratio between the number of channels of the first layer and the number of channels of the second layer. Each CNN layer in the LTT block is followed by a batch normalization~\cite{ioffe2015batch} and a non-linear activation function. In our case, the first is SeLU~\cite{klambauer2017selfnormalizing} and the second one is $bin_{act}$.  We can observe that the conditions (A-B-C-D) are now fulfilled: the input size is small ($n=6$), the input/output values are binary and the function is non-linear thanks to the SeLU function. Note that the LTT weights and the intermediate values are real.

\vspace*{\subseclen}
\subsection{Overall $\mathcal{TT}$net architecture} 
\label{subsubsec:Architecture}
\vspace*{\subseclen}

We integrated LTT blocks into $\mathcal{TT}$net, just as CNN filters are integrated into DCNNs: each LTT layer is composed of multiple LTT blocks and there are multiple LTT layers in total. Additionally, there is a pre-processing layer and a final layer. These two layers provide flexibility in adapting to different applications: scalability, interpretability, formal verification, and logic circuit design.

\myparagraph{Pre-processing layer.} To maintain the interpretability, verifiability and Boolean circuit properties of the network, we apply a 3-step pre-processing procedure to the inputs of DCNNs, which are typically floating-point values: (i) quantization of inputs \cite{jia2020efficient}; (ii) batch normalization \cite{narodytska2019search}; and (iii) step function. The details of the quantization step (i) can be found in Appendix~\ref{sec:App_1stlayer}. To improve scalability, we allow the addition of a first float-weighted CNN layer before batch normalization.

\myparagraph{Final layer.} To maintain the interpretability, verifiability and Boolean circuit properties of the network, we apply a final layer that is a single linear layer on 1-bit or 4-bit weight, which is equivalent to having 4 linear layers with 1-bit weight. For scalability, we also allowed the option of having this layer with floating weights.

\vspace*{\subseclen}
\subsection{Integration of human knowledge}\label{subsec:HK}
\vspace*{\subseclen}

One of the most valuable and unique features of $\mathcal{TT}$net is the integration of human knowledge through the automatic injection of ``Don’t Care Terms'' ($DCTs$) in the truth table as a post-training treatment. $DCTs$ represent the fact that for a particular input, the LTT block output can be equivalently 0 or 1 without affecting the overall performance of the DCNN. The Quine-McCluskey algorithm can then assign the optimal value to the $DCTs$ to reduce the DNF equation size. The injection of $DCTs$ is achieved automatically through human common sense constraint at post-training.

\myparagraph{Example 1C:} The above Example 1B illustrates the need for human knowledge injection: no one can be born in both France and US. This implies that the literals $x_2$ and $x_3$ must not be at 1 at the same time, uniquely for the second rule. Then, the DNF equation changes: we inject the $DCT$ inside the truth table as $y_{binary} =$ [0, 0, 1, $DCT$, 0, 0, 1, $DCT$, 1, 1, 1, $DCT$, 1, 1, 1, $DCT$].  This results in the new rule: $\Rule_{2, H.K.}^{\DNF} = \text{Born in the US}  \lor \text{Go to University}$. 

This automatic post-training operation significantly decreases the size of the rules while keeping the accuracy the same and avoiding nonsensical rules conditions.

\section{Results}
\label{sec:Results}

\vspace*{\subseclen}
\subsection{Experimental environment}
\label{subsec:Expe_settings}
\vspace*{\subseclen}

The project code 
was coded in Python with PyTorch library~\cite{paszke2019pytorch} for training, Numpy~\cite{van2011numpy} for testing. We used 4 Nvidia GeForce RTX 3090 GPUs and 8 cores Intel(R) Core(TM) i7-8650U CPU clocked at 1.90 GHz, 16 GB RAM.

\vspace*{\subseclen}
\subsection{Scalability: Claim 4}
\vspace*{\subseclen}

$\mathcal{TT}$net shows an accuracy of 41.1\% on ImageNet with truth tables of size $n=16$, a pre-processing layer of one CNN layer with floating weights, and a final layer with floating weights (as in \cite{hubara2016binarized, rastegari2016xnor}). These results are comparable to those achieved by the original BNN paper~\cite{hubara2016binarized} and XNOR-net~\cite{rastegari2016xnor}, see Table~\ref{table:Results_perfnaturelle_imagenets}. Our experimental results confirm that $\mathcal{TT}$net can achieve high accuracy on a large dataset. However, there is still room for improvement in terms of accuracy as BNNs now can achieve $\approx 70\% - 75\%$~\cite{liu2018bi} on ImageNet. Detailed information is given in Appendix~\ref{sec:Ann_natural}.

\begin{table}[htb!]
\vspace*{-0.2cm}
\caption{ \label{table:Results_perfnaturelle_imagenets} Comparison of top 1 and top 5 natural accuracy between $\mathcal{TT}$net, BNN and XNOR-Net on ImageNet.}
\resizebox{1\columnwidth}{!}{
\begin{tabular}{@{}lcccc@{}}
\toprule
                          \textbf{Accuracy} & \textbf{$\mathcal{TT}$net} ($n=16$) & \textbf{original BNN}   & \textbf{XNOR-Net} \\ \midrule

 top 1                                                             & 41.1 \%                                                          & 27.9 \%   & 44.2 \%            \\
                          top 5                                                             & 65.6 \%                                                         & 50.4 \%   & 69.2 \%         \\ \bottomrule
\end{tabular}}
\end{table}

\myparagraph{Opportunities.} $\mathcal{TT}$net presents several opportunities for future research. These include exploring its application on time series or graph-based datasets, developing better LTT architectures and training methods to achieve higher accuracy on ImageNet, and creating heuristics for converting truth tables to optimal CNF/DNF for large $n$. 

\vspace*{\subseclen}
\subsection{Global and exact interpretability: Claims 1A / 1B}
\vspace*{\subseclen}

In this section, we present the results of applying $\mathcal{TT}$net on two tabular datasets Adult and Breast Cancer~\cite{dua2017uci}. The accuracy and compactness performances of our model are reported in Table~\ref{tab:res_tab}, along with comparisons to Diff Logic Net~\cite{petersen2022deep}. Our results demonstrate that $\mathcal{TT}$net achieves higher accuracy with fewer parameters. Detailed information on the training conditions, architectures, and additional results can be found in Appendix~\ref{appendix:Global}. We provide experimental evidence to support our claims of interpretability on an Adult use case, which can be extended to other tabular datasets.

\begin{table}[htb!]
\vspace*{-0.2cm}
\centering
\caption{Accuracy (Acc.) performance of $\mathcal{TT}$net on two tabular datasets alongside the effect of the automatic post-training $DCTs$ incorporation with Human Knowledge (H.K.) in truth table on the number of gates (\# Gates idem as the complexity of the model). Comparison proposed with Diff Logic Net~\cite{petersen2022deep}, more detailed results are given in Appendix~\ref{subsec_appedix:Tabular}, Table~\ref{tab:tabular_appendix}.}
\label{tab:res_tab}
\resizebox{\columnwidth}{!}{
\begin{tabular}{@{}l|cc|cc|cc@{}}
\toprule
Models        & \multicolumn{2}{c|}{Diff Logic Net} & \multicolumn{2}{c|}{Ours without H.K.} & \multicolumn{2}{c}{Ours with H.K.} \\ \midrule
              & Acc.          & \# Gates         & Acc.     & \# Gates                     & Acc.            & \# Gates          \\ \midrule
Adult         & 84.8\%           & 1280             & 85.3\%   &  2156   & 85.3\%          & 1475             \\
Breast Cancer & 76.1\%           & 640             & 77.6\%   & 123                         & 77.6\%          & 71               \\
\bottomrule
\end{tabular}}
\end{table}

\begin{figure*}[htb!]
\vspace*{-0.2cm}
\centering
\hspace*{-0.75cm}
\includegraphics[scale=0.38]{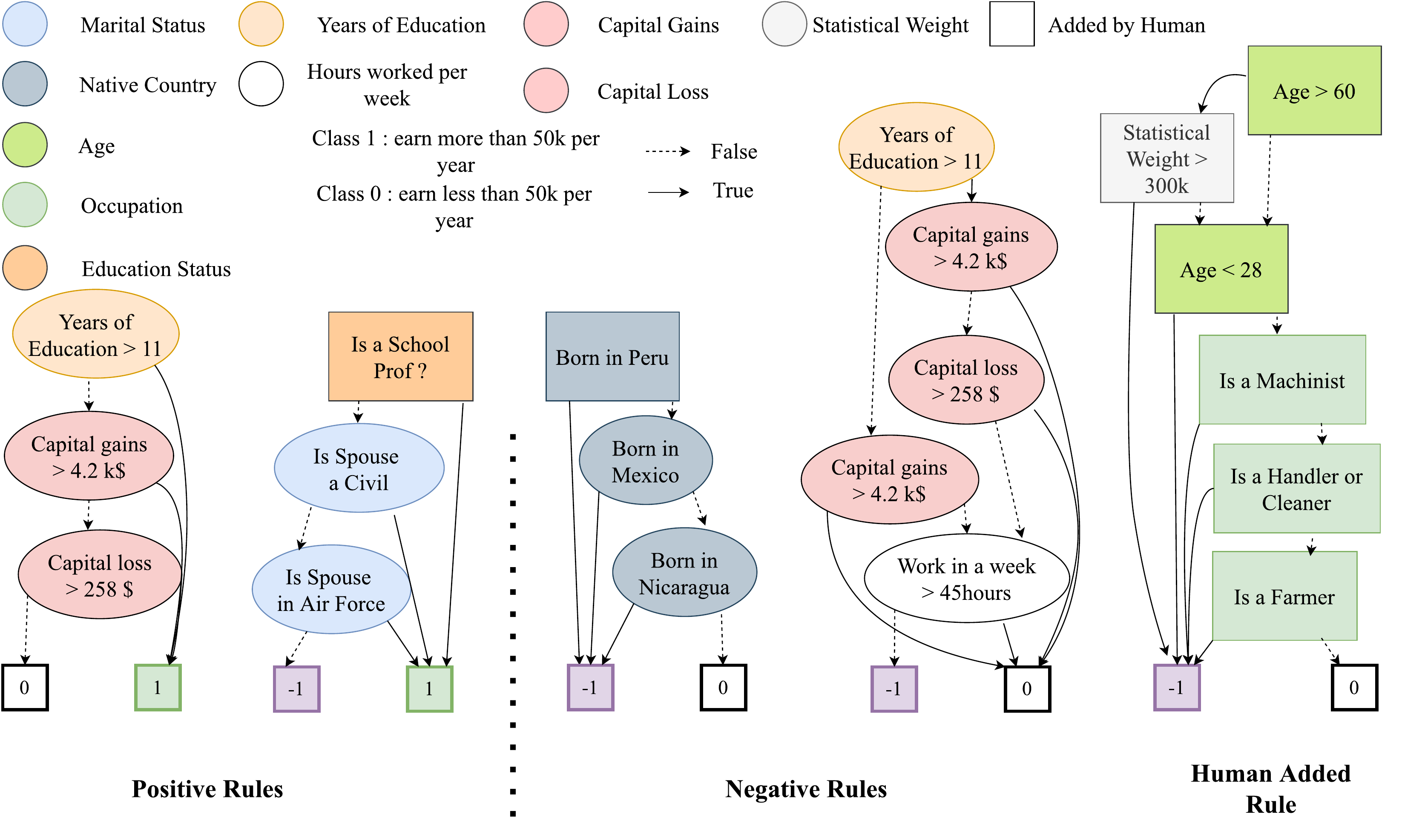}
\vspace*{-0.6cm}
\captionof{figure}{\label{fig:casestudy} DCNN $\mathcal{TT}$net model trained on Adult in the form of Boolean decision trees: the output of the DCNN and the output of these decision trees are the same, reaching 83.6\% accuracy. Added conditions are represented in rectangles. By modifying existing rules and incorporating the \textbf{Human Added Rule} we reach 84.6\% accuracy. On the same test set, Random Forest reaches 85.1\% accuracy and Decision Tree 84.4\% with depth 10. There is no contradiction to the rules: one person can not be born in both Mexico and Nicaragua.
}
\vspace*{-0.2cm}
\end{figure*}

\myparagraph{Adult use case: exact and global interpretability.} For global and exact interpretability, we first convert each LTT block of the DCNN $\mathcal{TT}$net model into truth tables, then into DNFs, then into a set of rules $\mathcal{R}$. Then we transform the rules in $\mathcal{R}$ into their equivalent Reduced Ordered Binary Decision Diagram (ROBDD) representation. This transformation is fast and automatic. One DCNN $\mathcal{TT}$net decision tree form can be observed in Figure~\ref{fig:casestudy}: the resulting decision mechanism is small and easily understandable.

In the Adult dataset, the goal is to predict whether an individual $I$ will earn more than \$50,000 per year in 1994. Given an individual's feature inputs $I$, the first rule of Figure~\ref{fig:casestudy} can be read as follows: if $I$ has completed more than 11 years of education, then the rule is satisfied. If not, then the rule is satisfied if $I$ earns more than \$4,200 in investments per month or loses more than \$228. If the rule is satisfied, $I$ earns one positive point. If $I$ has more positive points than negative points, the model predicts that $I$ will earn more than \$50,000 per year. 

\myparagraph{Mitigating contextual drift in $\mathcal{TT}$net DCNN through global and ixact interpretability.} The proposed process provides a clear method for obtaining a comprehensive and precise interpretation of the $\mathcal{TT}$net DCNN through the delivery of all Boolean decision trees. This results in a complete understanding of the model and the ability to calculate its predictions exactly. It is essential to recognize that machine learning models may not always generalize well to new data from different geographic locations or contexts, a phenomenon known as "contextual drift" or "concept drift"~\cite{gama2004learning}. The global and exact interpretation of the $\mathcal{TT}$net DCNN is vital in this regard, as it allows for human feedback on the model's rules and the potential for these rules to be influenced by contextual drift. For example, as depicted in Figure~\ref{fig:casestudy}, this accurate model trained on US data is highly biased towards the US and is likely to perform poorly if applied in South America due to rule number 3. This highlights once again the significance of having global and exact interpretability of deep neural networks, as emphasized by recent NIST Artificial Intelligence Risk Management Framework~\cite{ai2023artificial}.

\myparagraph{Discussion on automatic post-training human knowledge incorporation with $DCTs$.} Table~\ref{tab:res_tab} illustrates the impact of incorporating human knowledge through the use of automatic $DCTs$ in the truth table in terms of the number of gates and accuracy. As shown, incorporating $DCTs$ results in a significant decrease in the number of gates without any decrease of accuracy. For a concrete example, a rule without human knowledge is $\Rule = \text{Born in Nicaragua}  \lor \text{Born in Haiti}  \lor (\text{Born in Columbia} \land \text{Born in Jamaica} ) \lor (\overline{\text{Born in Jamaica}} \land (\text{Age} >46) )$, while a rule with automatic human knowledge injection via $DCTs$ becomes: $\Rule_{H.K.} = \text{Born in Nicaragua}  \lor \text{Born in Haiti} \lor (\overline{\text{Born in Jamaica}} \land (\text{Age} >46) )$. The condition $ (\text{Born in Columbia} \land \text{Born in Jamaica} )$ was automatically removed as it is impossible to satisfy. This experiment demonstrates the unique ability to incorporate human knowledge in a post-training process, which is not possible with other models~\cite{agarwal2020neural, wang2021scalable, yang2021learning}. 

Note that in Figure~\ref{fig:casestudy}, we also illustrate the ability of $\mathcal{TT}$net to incorporate hand-crafted human knowledge by allowing the modification of existing rules.


\myparagraph{Opportunities.} For future research, it would be interesting to apply $\mathcal{TT}$net to regression tasks, to investigate its potential for studying causality, and to propose an agnostic global explainer for any model based on $\mathcal{TT}$net.

\vspace*{\subseclen}
\subsection{Complete/sound formal verification: Claims 2A / 2B}
\vspace*{\subseclen}

In formal verification, two strategies are commonly applied: either using a specific solver to verify ReLU based-DNN like $\beta$-Crown~\cite{wang2021beta}, or using a specific architecture that allows a generic verification method. We compared both methods based on natural accuracy, verified accuracy for $l_{\infty}$-norm bounded input perturbations, and verification time on MNIST and CIFAR-10. Detailed information on the training conditions, architectures, and additional results can be found in Appendix~\ref{appendix:Sound}. It is important to note that our pre-processing layer and training configuration are the same as \cite{jia2020efficient}.

\begin{table}[htb!]
\vspace*{-0.2cm}
\centering
\caption{\label{tab:Crown_res}Comparison of verification strategies: usage of a general DCNN to verify with $\alpha$-$\beta$-Crown \cite{xu2020fast, wang2021beta} or using specific $\mathcal{TT}$net with a general SAT verification method. The comparison is based on the 7 benchmarks from the VNN competition, and the results are presented as an average, full results are available in Table~\ref{tab:Appendix_beta_comp} of Appendix~\ref{subsec:Appendix_beta_comp} }
\resizebox{\columnwidth}{!}{
\begin{tabular}{l|cc|cc}
\hline
         & \multicolumn{2}{c|}{\begin{tabular}[c]{@{}c@{}}General DCNN + $\alpha$-$\beta$-Crown \\ \cite{xu2020fast, wang2021beta} \end{tabular}} & \multicolumn{2}{c}{\begin{tabular}[c]{@{}c@{}}$\mathcal{TT}$net + General SAT \\  Verification Pipeline\end{tabular}} \\ \hline
         & Verif. time   (s)                           & Timeout  (\%)                            & Verif. time (s)                                       & Timeout (\%)                                      \\ \hline
MNIST    & 96                                             & 13                                   & 0.06 ($\times$1600)                                             & 0                                             \\
CIFAR-10 & 175                                            & 27                                   & 0.14 ($\times$1250)                                             & 0                                             \\ \hline
\end{tabular}}
\end{table}

In Table~\ref{tab:Crown_res}, we present a comparison of our proposed verification strategy, which utilizes the $\mathcal{TT}$net architecture and classical verification tools, against the state-of-the-art $\alpha$-$\beta$-Crown method, the winner of the VNN competition 2021. The comparison is based on the 7 benchmarks from the VNN competition, and the results are presented as an average. Our approach demonstrates a significant improvement in verification time, with an average speedup of 1250x for CIFAR-10 and 1600x for MNIST, at the same noise level. Additionally, a higher verified accuracy (+4\% and +7\%) was observed on the cifar\_10\_resnet benchmark (see Table~\ref{tab:Appendix_beta_comp} in Appendix~\ref{subsec:Appendix_beta_comp}). Furthermore, our approach did not encounter any timeouts, whereas $\alpha$-$\beta$-Crown had an average of more than 10\% timeouts.

It should be noted that our strategy cannot be directly compared in the VNN competition as the competition focuses on novel DNN verification algorithms/pipelines, whereas we propose a new DNN family ($\mathcal{TT}$net) that can be easily verified using classical verification tools. However, the results presented demonstrate the competitiveness of our approach. One can use our strategy to verify the robustness property on CIFAR-10, 1K images, standard DCNN with $\alpha$-$\beta$-Crown which takes 2 days, whereas it takes 14sec if one chooses to verify $\mathcal{TT}$net.

\myparagraph{Comparison with BNN.} We compared our work with the state-of-the-art of exact verification for BNNs \cite{jia2020efficient, narodytska2019search}. As shown in Table~\ref{table:Results_TTDCNN}, our verified accuracy is competitive to the one of BNN in all cases. In low noise case, we slightly outperform BNN~\cite{jia2020efficient} in terms of verified accuracy and verification time. For high noise, we offer a trade-off: a slightly better verification time for a slightly lower verified accuracy. In addition, we experienced a better resolution time than BNNs while using a general SAT solver, namely MiniCard \cite{liffiton2012cardinality}. We highlight here that in~\cite{jia2020efficient} the SAT solver is custom-made and specific to their problem, while in our case we use a general one. We tested 9 SAT solvers \cite{imms-sat18} on MNIST high noise: the best one is MiniCard with 8e-4s, the worst is MapleCM with 33e-4s ; whereas MiniSat takes 0.242s for BNN in~\cite{jia2020efficient}.

\begin{table}[!htb]
\vspace*{-0.2cm}
\centering
\caption{\label{table:Results_TTDCNN} Application of $\mathcal{TT}$net to complete adversarial robustness verification for low and high noise bounded by $l_{\infty}$. We tabulate results of verified accuracy, natural accuracy and mean verification time on MNIST and CIFAR-10 datasets in comparison to state-of-the-art SAT methods. The best verified accuracy and verification time are displayed in bold.}
\begin{minipage}[t]{\linewidth}
\resizebox{1\columnwidth}{!}{
\begin{tabular}{@{}lclccccccc@{}}
\toprule
\textbf{Dataset}                      & &  \textbf{Complete}                 &                      & \multicolumn{2}{c}{\textbf{Accuracy}} &                      & \textbf{Verification}  & \textbf{Timeout} \\ 
 (noise) &               &    \textbf{Method}     &                      & Verified     & Natural     & & \textbf{time} (s) &  \\ \midrule
\multirow{3}{*}{\begin{tabular}[l]{@{}l@{}}MNIST\\  ($\epsilon_{test}$ = 0.1)\end{tabular}}     &        & $\mathcal{TT}$net              &                      & \textbf{95.12\%}        & 98.33\%     &                      & \textbf{0.012} &                                                    0  \\ 
                                                                                                   &                                   &   \cite{jia2020efficient}  &                      & 91.68\%        & 97.46\%         &                      & 0.1115                                                     & 0                    \\ 
                                                                                                   &                                   &   \cite{narodytska2019search} $^*$        &                      & 20.0\%        & 96.0 \%     &                      & 5 & 0                    \\  \midrule
\multirow{2}{*}{\begin{tabular}[l]{@{}l@{}}MNIST\\($\epsilon_{test}$ = 0.3)\end{tabular}}     &        & $\mathcal{TT}$net              &                      & 66.24\%       & 97.43 \%    &                      & \textbf{0.065} & 0                    \\ 
                                                                                                   &                                   &  \cite{jia2020efficient}          &                      & \textbf{77.59\%}        & 96.36\%     &                      & 0.1179                                                  & 0                 \\  \midrule \midrule
\multirow{2}{*}{\begin{tabular}[l]{@{}l@{}}CIFAR-10\\  ($\epsilon_{test}$ = 2/255)\end{tabular}} &        & $\mathcal{TT}$net              &                      & \textbf{32.32\%}        & 49.23\%     &                      & \textbf{0.06 }                                                   & 0                   \\ 
                                                                                                   &                                   &  \cite{jia2020efficient}          &                      & 30.49\%        & 47.35\%     &                      & \t0.1750                                                  &                0       \\  \midrule
\multirow{2}{*}{\begin{tabular}[l]{@{}l@{}}CIFAR-10 \\ ($\epsilon_{test}$ = 8/255)\end{tabular}} &      & $\mathcal{TT}$net              &                      & 21.08\%        & 31.13\%     &                      & \textbf{0.04} & 0                   \\ 
                                                                                                   &                                   & \cite{jia2020efficient}          & \multicolumn{1}{c}{} & \textbf{22.55\%  }      & 35.00\%     &                      & 0.1781                                                 & 0                      \\  \bottomrule
\end{tabular}}
\smallskip
\parbox[t]{\textwidth}{\tiny
    \textit{$^*$~results given on the first 1K images of the test set. Moreover, the authors only authorize a maximum of 20 pixels to switch.}
}
\end{minipage}%
\vspace*{-0.2cm}
\end{table}

\myparagraph{Discussion on formal verification strategies.} In this section, we are presenting a comparison of three different approaches for the formal verification of neural networks. The first approach, $\beta$-Crown~\cite{xu2020fast, wang2021beta}, is a general method that can be applied to any ReLU-based DNN but does not provide any guarantee of correctness. The second approach is specific to BNNs and provides a guarantee of correctness but requires a well-crafted specific SAT solver. Our proposed $\mathcal{TT}$net architecture constitutes a third approach in that it can be verified using any SAT solver, while providing a guarantee of correctness. In terms of verification time, our approach is significantly faster than $\alpha-\beta$-Crown and slightly faster than BNN, except for low noise CIFAR-10 case. On the other hand, it should be noted that our method may show a lower verified accuracy, particularly in high-noise scenarios.


\myparagraph{Opportunities.} One potential opportunity for future research is to propose a dedicated robustness training for increasing $\mathcal{TT}$net verified accuracy, especially for high noise.

\vspace*{\subseclen}
\subsection{Differentiable logic gates CNNs: Claims 3A / 3B}
\vspace*{\subseclen}

We have experimental evidence to support our claims of providing easy-to-train, scalable, and compact (compact for tabular and MNIST) logic gate CNNs. The results are presented in Table~\ref{tab:res_tab} for tabular datasets (Adult and Breast Cancer) and Table~\ref{tab:compactiy_image} for image datasets (MNIST and CIFAR-10) along with Diff Logic Net~\cite{petersen2022deep}. More information on training conditions and architectures can be found in Appendix~\ref{appendix:Differentiable}.

\begin{table}[htb!]
\vspace*{-0.2cm}
\centering
\caption{\label{tab:compactiy_image} Comparison of $\mathcal{TT}$net with state-of-the-art Differentiable Logic Gate model~\cite{petersen2022deep} on MNIST and CIFAR-10 Datasets for two models. Acc. stands for Accuracy, OPs for binary OPerations, \# Gates for number of gates.}
\resizebox{\columnwidth}{!}{
\begin{tabular}{ll|ccc|ccc}
\hline
\multicolumn{2}{l|}{Datasets}                                                      & \multicolumn{3}{c|}{MNSIT}                                                                                                     & \multicolumn{3}{c}{CIFAR-10}                                                                                                   \\ \hline
\multicolumn{2}{l|}{}                                                              & Acc.    & \begin{tabular}[c]{@{}c@{}}\# Gates \\ = OPs\end{tabular} & \begin{tabular}[c]{@{}c@{}}Training \\ Time\end{tabular} & Acc.     & \begin{tabular}[c]{@{}c@{}}\# Gates \\ = OPs\end{tabular} & \begin{tabular}[c]{@{}c@{}}Training \\ Time\end{tabular} \\ \hline
\multirow{2}{*}{\begin{tabular}[c]{@{}l@{}}Diff Logic Net\\\cite{petersen2022deep} \end{tabular}} & Small & 97.69\% & 48K                                                       & 1.8h                                                     & 51.27\% & 48K                                                       & 1.3h                                                     \\
                                                                           & Big   & 98.47\% & 384K                                                      & 5.3h                                                     & 62.14\% & 5.12M                                                     & 90.3h                                                    \\ \hline
\multirow{2}{*}{\begin{tabular}[c]{@{}l@{}}$\mathcal{TT}$net\\ \end{tabular}}      & Small & 97.23\% & 46K                                                       & 10min                                                    & 51.76\% & 12M                                                       & 20min                                                    \\
                                                                           & Big   & 98.02\% & 360K                                                      & 10min                                                    & 70.20\% & 6015M                                                     & 20min                                                    \\ \hline
\end{tabular}}
\vspace*{-0.2cm}
\end{table}

\myparagraph{Convolutional logic gate networks.} As highlighted in recent research, convolutional logic gate networks are a promising direction for future study~\cite{petersen2022deep}. Our work successfully demonstrates that $\mathcal{TT}$net is a Boolean circuit, providing evidence of its potential as a compact and interpretable neural network architecture. Finally, in convolutional logic gate networks as well as in logic gate circuit networks, we do not need activation functions as they are intrinsically non-linear. To show the easy-to-train character of our model, Table~\ref{tab:compactiy_image} reports the training times: $\mathcal{TT}$net can be trained in less than 20 minutes and \cite{petersen2022deep} in between 1.8 hour and 90 hours: our training method is much efficient (see next Section for a hardware discussion). We recall that all the experiments were coded in Pytorch~\cite{paszke2019pytorch}, a standard ML framework.

\myparagraph{Compactness on tabular and MNIST.} Table~\ref{tab:res_tab} data illustrate that on the tabular dataset, $\mathcal{TT}$net outperforms Diff Logic Net~\cite{petersen2022deep} in both accuracy and number of parameters. On the MNIST dataset, in Table~\ref{tab:compactiy_image}, our model has slightly fewer parameters and slightly lower accuracy than Diff Logic Net~\cite{petersen2022deep}. We highlight that we did not count the number of gates needed to perform the final binary layer as in~\cite{petersen2022deep}. 

\myparagraph{Scalability on CIFAR-10.}  We define scalability in terms of accuracy, training time, and number of Boolean gates. The results in Table~\ref{tab:compactiy_image} show that while Diff Logic Net ~\cite{petersen2022deep} achieves a maximum accuracy of 62\%, $\mathcal{TT}$net achieves an accuracy of 70.2\%. This highlights the superior scalability of $\mathcal{TT}$net in terms of accuracy. However, we note that $\mathcal{TT}$net does not scale as well as Diff Logic for small numbers of parameters. However, this limitation is offset by the fact that $\mathcal{TT}$net is much more trainable-friendly Logic gate architecture than Diff Logic: $\mathcal{TT}$net can reach 70\% accuracy in less than an hour of training, compared to Diff Logic which takes 90 hours to reach a lower accuracy (-8\%) with an overall Diff Logic Net architecture which is more than 100 times smaller. This training time difference is consistent across the datasets and the model size. Please note that \cite{petersen2022deep} trained on one NVIDIA A6000 GPU we trained on one NVIDIA GeForce RTX 3090 GPU\footnote{According to this benchmark - both GPUs are equivalent in terms of images training time: \href{https://lambdalabs.com/blog/nvidia-rtx-a6000-vs-rtx-3090-benchmarks}{https://lambdalabs.com/blog/nvidia-rtx-a6000-vs-rtx-3090-benchmarks}}. Overall, the high accuracy achieved by $\mathcal{TT}$net on the CIFAR-10 dataset with effective training highlights its potential as a differentiable CNN logic gate architecture.


\myparagraph{Opportunities.} Our $\mathcal{TT}$net model has demonstrated promising results, but there is still room for improvement. Specifically, we aim to reduce the complexity (\# Gates) of our model when scaling to CIFAR-10 and explore the use of CNNs Boolean circuits for larger datasets like ImageNet.

\section{Conclusion}
\label{sec:Conclusion}

This work is a step towards more transparent DCNNs. While there is still room for improvement, we hope that it will inspire further exploration into the use of truth tables as a tool for applying DCNNs to critical applications.

\newpage

\bibliography{example_paper}

\begin{thebibliography}{117}
\providecommand{\natexlab}[1]{#1}
\providecommand{\url}[1]{\texttt{#1}}
\expandafter\ifx\csname urlstyle\endcsname\relax
  \providecommand{\doi}[1]{doi: #1}\else
  \providecommand{\doi}{doi: \begingroup \urlstyle{rm}\Url}\fi

\bibitem[Ab{\'\i}o et~al.(2011)Ab{\'\i}o, Nieuwenhuis, Oliveras, and
  Rodr{\'\i}guez-Carbonell]{abio2011bdds}
Ab{\'\i}o, I., Nieuwenhuis, R., Oliveras, A., and Rodr{\'\i}guez-Carbonell, E.
\newblock Bdds for pseudo-boolean constraints--revisited.
\newblock In \emph{International Conference on Theory and Applications of
  Satisfiability Testing}, pp.\  61--75. Springer, 2011.

\bibitem[Agarwal et~al.(2020)Agarwal, Frosst, Zhang, Caruana, and
  Hinton]{agarwal2020neural}
Agarwal, R., Frosst, N., Zhang, X., Caruana, R., and Hinton, G.~E.
\newblock Neural additive models: Interpretable machine learning with neural
  nets.
\newblock \emph{arXiv preprint arXiv:2004.13912}, 2020.

\bibitem[AI(2023)]{ai2023artificial}
AI, N.
\newblock Artificial intelligence risk management framework (ai rmf 1.0).
\newblock 2023.

\bibitem[Akers(1978)]{akers1978binary}
Akers, S.~B.
\newblock Binary decision diagrams.
\newblock \emph{IEEE Transactions on computers}, 27\penalty0 (06):\penalty0
  509--516, 1978.

\bibitem[Andriushchenko \& Hein(2019)Andriushchenko and
  Hein]{andriushchenko2019provably}
Andriushchenko, M. and Hein, M.
\newblock Provably robust boosted decision stumps and trees against adversarial
  attacks.
\newblock \emph{Advances in Neural Information Processing Systems}, 32, 2019.

\bibitem[Angelino et~al.(2017)Angelino, Larus-Stone, Alabi, Seltzer, and
  Rudin]{angelino2017learning}
Angelino, E., Larus-Stone, N., Alabi, D., Seltzer, M., and Rudin, C.
\newblock Learning certifiably optimal rule lists for categorical data.
\newblock \emph{arXiv preprint arXiv:1704.01701}, 2017.

\bibitem[Aqajari et~al.(2021)Aqajari, Naeini, Mehrabadi, Labbaf, Dutt, and
  Rahmani]{aqajari2021pyeda}
Aqajari, S. A.~H., Naeini, E.~K., Mehrabadi, M.~A., Labbaf, S., Dutt, N., and
  Rahmani, A.~M.
\newblock pyeda: An open-source python toolkit for pre-processing and feature
  extraction of electrodermal activity.
\newblock \emph{Procedia Computer Science}, 184:\penalty0 99--106, 2021.

\bibitem[Araujo et~al.(2019)Araujo, Norris, and Sim]{araujo2019computing}
Araujo, A., Norris, W., and Sim, J.
\newblock Computing receptive fields of convolutional neural networks.
\newblock \emph{Distill}, 2019.
\newblock \doi{10.23915/distill.00021}.
\newblock https://distill.pub/2019/computing-receptive-fields.

\bibitem[Arora \& Barak(2009)Arora and Barak]{arora2009computational}
Arora, S. and Barak, B.
\newblock \emph{Computational complexity: a modern approach}.
\newblock Cambridge University Press, 2009.

\bibitem[Bak et~al.(2021)Bak, Liu, and Johnson]{bak2021second}
Bak, S., Liu, C., and Johnson, T.
\newblock The second international verification of neural networks competition
  (vnn-comp 2021): Summary and results.
\newblock \emph{arXiv preprint arXiv:2109.00498}, 2021.

\bibitem[Baluta et~al.(2019)Baluta, Shen, Shinde, Meel, and
  Saxena]{baluta2019quantitative}
Baluta, T., Shen, S., Shinde, S., Meel, K.~S., and Saxena, P.
\newblock Quantitative verification of neural networks and its security
  applications.
\newblock In \emph{Proceedings of the 2019 ACM SIGSAC Conference on Computer
  and Communications Security}, pp.\  1249--1264, 2019.

\bibitem[Bertsimas \& Dunn(2017)Bertsimas and Dunn]{bertsimas2017optimal}
Bertsimas, D. and Dunn, J.
\newblock Optimal classification trees.
\newblock \emph{Machine Learning}, 106:\penalty0 1039--1082, 2017.

\bibitem[Bessiere et~al.(2009)Bessiere, Hebrard, and
  O’Sullivan]{bessiere2009minimising}
Bessiere, C., Hebrard, E., and O’Sullivan, B.
\newblock Minimising decision tree size as combinatorial optimisation.
\newblock In \emph{International Conference on Principles and Practice of
  Constraint Programming}, pp.\  173--187. Springer, 2009.

\bibitem[Biere et~al.(2009)Biere, Heule, and van Maaren]{biere2009handbook}
Biere, A., Heule, M., and van Maaren, H.
\newblock \emph{Handbook of satisfiability}, volume 185.
\newblock IOS press, 2009.

\bibitem[Blake(1938)]{blake1938}
Blake, A.
\newblock {Corrections to Canonical expressions in Boolean algebra}.
\newblock \emph{Journal of Symbolic Logic}, 3\penalty0 (2):\penalty0 112–113,
  1938.
\newblock \doi{10.2307/2267595}.

\bibitem[Breiman(2001)]{breiman2001random}
Breiman, L.
\newblock Random forests.
\newblock \emph{Machine learning}, 45\penalty0 (1):\penalty0 5--32, 2001.

\bibitem[Brendel \& Bethge(2019)Brendel and Bethge]{brendel2019approximating}
Brendel, W. and Bethge, M.
\newblock Approximating cnns with bag-of-local-features models works
  surprisingly well on imagenet.
\newblock \emph{arXiv preprint arXiv:1904.00760}, 2019.

\bibitem[Brown et~al.(2020)Brown, Mann, Ryder, Subbiah, Kaplan, Dhariwal,
  Neelakantan, Shyam, Sastry, Askell, et~al.]{brown2020language}
Brown, T.~B., Mann, B., Ryder, N., Subbiah, M., Kaplan, J., Dhariwal, P.,
  Neelakantan, A., Shyam, P., Sastry, G., Askell, A., et~al.
\newblock Language models are few-shot learners.
\newblock \emph{arXiv preprint arXiv:2005.14165}, 2020.

\bibitem[Bryant(1986)]{bryant1986graph}
Bryant, R.~E.
\newblock Graph-based algorithms for boolean function manipulation.
\newblock \emph{Computers, IEEE Transactions on}, 100\penalty0 (8):\penalty0
  677--691, 1986.

\bibitem[Carlini et~al.(2017)Carlini, Katz, Barrett, and
  Dill]{carlini2017provably}
Carlini, N., Katz, G., Barrett, C., and Dill, D.~L.
\newblock Provably minimally-distorted adversarial examples.
\newblock \emph{arXiv preprint arXiv:1709.10207}, 2017.

\bibitem[Carlini et~al.(2019)Carlini, Athalye, Papernot, Brendel, Rauber,
  Tsipras, Goodfellow, Madry, and Kurakin]{carlini2019evaluating}
Carlini, N., Athalye, A., Papernot, N., Brendel, W., Rauber, J., Tsipras, D.,
  Goodfellow, I., Madry, A., and Kurakin, A.
\newblock On evaluating adversarial robustness.
\newblock \emph{arXiv preprint arXiv:1902.06705}, 2019.

\bibitem[Cheng et~al.(2017)Cheng, N{\"u}hrenberg, and Ruess]{cheng2017maximum}
Cheng, C.-H., N{\"u}hrenberg, G., and Ruess, H.
\newblock Maximum resilience of artificial neural networks.
\newblock In \emph{International Symposium on Automated Technology for
  Verification and Analysis}, pp.\  251--268. Springer, 2017.

\bibitem[Cheng et~al.(2018)Cheng, N{\"u}hrenberg, Huang, and
  Ruess]{cheng2018verification}
Cheng, C.-H., N{\"u}hrenberg, G., Huang, C.-H., and Ruess, H.
\newblock Verification of binarized neural networks via inter-neuron factoring.
\newblock In \emph{Working Conference on Verified Software: Theories, Tools,
  and Experiments}, pp.\  279--290. Springer, 2018.

\bibitem[Choi et~al.(2017)Choi, Shi, Shih, and Darwiche]{choi2017compiling}
Choi, A., Shi, W., Shih, A., and Darwiche, A.
\newblock Compiling neural networks into tractable boolean circuits.
\newblock \emph{intelligence}, 2017.

\bibitem[Cohen(1995)]{cohen1995fast}
Cohen, W.~W.
\newblock Fast effective rule induction.
\newblock In \emph{Machine learning proceedings 1995}, pp.\  115--123.
  Elsevier, 1995.

\bibitem[Cohen \& Singer(1999)Cohen and Singer]{cohen1999simple}
Cohen, W.~W. and Singer, Y.
\newblock A simple, fast, and effective rule learner.
\newblock \emph{AAAI/IAAI}, 99\penalty0 (335-342):\penalty0 3, 1999.

\bibitem[Croce et~al.(2021)Croce, Andriushchenko, Sehwag, Debenedetti,
  Flammarion, Chiang, Mittal, and Hein]{croce2021robustbench}
Croce, F., Andriushchenko, M., Sehwag, V., Debenedetti, E., Flammarion, N.,
  Chiang, M., Mittal, P., and Hein, M.
\newblock Robustbench: a standardized adversarial robustness benchmark.
\newblock In \emph{Thirty-fifth Conference on Neural Information Processing
  Systems Datasets and Benchmarks Track}, 2021.
\newblock URL \url{https://openreview.net/forum?id=SSKZPJCt7B}.

\bibitem[Darwiche \& Marquis(2002)Darwiche and Marquis]{darwiche2002knowledge}
Darwiche, A. and Marquis, P.
\newblock A knowledge compilation map.
\newblock \emph{Journal of Artificial Intelligence Research}, 17:\penalty0
  229--264, 2002.

\bibitem[Dash et~al.(2018)Dash, Gunluk, and Wei]{dash2018boolean}
Dash, S., Gunluk, O., and Wei, D.
\newblock Boolean decision rules via column generation.
\newblock \emph{Advances in neural information processing systems}, 31, 2018.

\bibitem[Deng et~al.(2009)Deng, Dong, Socher, Li, Li, and
  Fei-Fei]{imagenet_cvpr09}
Deng, J., Dong, W., Socher, R., Li, L.-J., Li, K., and Fei-Fei, L.
\newblock {ImageNet: A Large-Scale Hierarchical Image Database}.
\newblock In \emph{CVPR09}, 2009.

\bibitem[Dosovitskiy et~al.(2020)Dosovitskiy, Beyer, Kolesnikov, Weissenborn,
  Zhai, Unterthiner, Dehghani, Minderer, Heigold, Gelly,
  et~al.]{dosovitskiy2020image}
Dosovitskiy, A., Beyer, L., Kolesnikov, A., Weissenborn, D., Zhai, X.,
  Unterthiner, T., Dehghani, M., Minderer, M., Heigold, G., Gelly, S., et~al.
\newblock An image is worth 16x16 words: Transformers for image recognition at
  scale.
\newblock \emph{arXiv preprint arXiv:2010.11929}, 2020.

\bibitem[Driscoll(2020)]{driscoll2020system}
Driscoll, M.
\newblock System and method for adapting a neural network model on a hardware
  platform, July~2 2020.
\newblock US Patent App. 16/728,884.

\bibitem[Dua \& Graff(2017)Dua and Graff]{Dua:2019}
Dua, D. and Graff, C.
\newblock {UCI} machine learning repository, 2017.
\newblock URL \url{http://archive.ics.uci.edu/ml}.

\bibitem[Dua et~al.(2017)Dua, Graff, et~al.]{dua2017uci}
Dua, D., Graff, C., et~al.
\newblock Uci machine learning repository.
\newblock 2017.

\bibitem[Dumoulin \& Visin(2016)Dumoulin and Visin]{dumoulin2016guide}
Dumoulin, V. and Visin, F.
\newblock A guide to convolution arithmetic for deep learning.
\newblock \emph{arXiv preprint arXiv:1603.07285}, 2016.

\bibitem[Dvijotham et~al.(2018)Dvijotham, Gowal, Stanforth, Arandjelovic,
  O'Donoghue, Uesato, and Kohli]{dvijotham2018training}
Dvijotham, K., Gowal, S., Stanforth, R., Arandjelovic, R., O'Donoghue, B.,
  Uesato, J., and Kohli, P.
\newblock Training verified learners with learned verifiers.
\newblock \emph{arXiv preprint arXiv:1805.10265}, 2018.

\bibitem[E{\'e}n \& S{\"o}rensson(2006)E{\'e}n and
  S{\"o}rensson]{een2006translating}
E{\'e}n, N. and S{\"o}rensson, N.
\newblock Translating pseudo-boolean constraints into sat.
\newblock \emph{Journal on Satisfiability, Boolean Modeling and Computation},
  2\penalty0 (1-4):\penalty0 1--26, 2006.

\bibitem[Ehlers(2017)]{ehlers2017formal}
Ehlers, R.
\newblock Formal verification of piece-wise linear feed-forward neural
  networks.
\newblock In \emph{International Symposium on Automated Technology for
  Verification and Analysis}, pp.\  269--286. Springer, 2017.

\bibitem[Evans et~al.(2021)Evans, Bo{\v{s}}njak, Buesing, Ellis, Pfau, Kohli,
  and Sergot]{evans2021making}
Evans, R., Bo{\v{s}}njak, M., Buesing, L., Ellis, K., Pfau, D., Kohli, P., and
  Sergot, M.
\newblock Making sense of raw input.
\newblock \emph{Artificial Intelligence}, 299:\penalty0 103521, 2021.

\bibitem[Freitas(2014)]{freitas2014comprehensible}
Freitas, A.~A.
\newblock Comprehensible classification models: a position paper.
\newblock \emph{ACM SIGKDD explorations newsletter}, 15\penalty0 (1):\penalty0
  1--10, 2014.

\bibitem[Friedman et~al.(1997)Friedman, Geiger, and
  Goldszmidt]{friedman1997bayesian}
Friedman, N., Geiger, D., and Goldszmidt, M.
\newblock Bayesian network classifiers.
\newblock \emph{Machine learning}, 29\penalty0 (2):\penalty0 131--163, 1997.

\bibitem[Fryer et~al.(2021)Fryer, Str{\"u}mke, and Nguyen]{fryer2021shapley}
Fryer, D., Str{\"u}mke, I., and Nguyen, H.
\newblock Shapley values for feature selection: the good, the bad, and the
  axioms.
\newblock \emph{IEEE Access}, 9:\penalty0 144352--144360, 2021.

\bibitem[Gama et~al.(2004)Gama, Medas, Castillo, and
  Rodrigues]{gama2004learning}
Gama, J., Medas, P., Castillo, G., and Rodrigues, P.
\newblock Learning with drift detection.
\newblock In \emph{Advances in Artificial Intelligence--SBIA 2004: 17th
  Brazilian Symposium on Artificial Intelligence, Sao Luis, Maranhao, Brazil,
  September 29-Ocotber 1, 2004. Proceedings 17}, pp.\  286--295. Springer,
  2004.

\bibitem[Gottemukkula(2020)]{gottemukkula2020polynomial}
Gottemukkula, V.
\newblock Polynomial activation functions.
\newblock 2020.

\bibitem[He et~al.(2020)He, Ma, and Wang]{he2020extract}
He, C., Ma, M., and Wang, P.
\newblock Extract interpretability-accuracy balanced rules from artificial
  neural networks: A review.
\newblock \emph{Neurocomputing}, 387:\penalty0 346--358, 2020.

\bibitem[Hoefler et~al.(2021)Hoefler, Alistarh, Ben-Nun, Dryden, and
  Peste]{hoefler2021sparsity}
Hoefler, T., Alistarh, D., Ben-Nun, T., Dryden, N., and Peste, A.
\newblock Sparsity in deep learning: Pruning and growth for efficient inference
  and training in neural networks.
\newblock \emph{J. Mach. Learn. Res.}, 22\penalty0 (241):\penalty0 1--124,
  2021.

\bibitem[H{\"o}lldobler et~al.(2012)H{\"o}lldobler, Manthey, and
  Steinke]{holldobler2012compact}
H{\"o}lldobler, S., Manthey, N., and Steinke, P.
\newblock A compact encoding of pseudo-boolean constraints into sat.
\newblock In \emph{Annual Conference on Artificial Intelligence}, pp.\
  107--118. Springer, 2012.

\bibitem[Huan et~al.(2022)Huan, Kaidi, Shiqi, and Cho-Jui]{tutorialAAAI}
Huan, Z., Kaidi, X., Shiqi, W., and Cho-Jui, H.
\newblock Aaai 2022: 'tutorial on neural network verification: Theory and
  practice'.
\newblock \url{https://neural-network-verification.com/}, 2022.

\bibitem[Hubara et~al.(2016)Hubara, Courbariaux, Soudry, El-Yaniv, and
  Bengio]{hubara2016binarized}
Hubara, I., Courbariaux, M., Soudry, D., El-Yaniv, R., and Bengio, Y.
\newblock Binarized neural networks.
\newblock \emph{Advances in neural information processing systems}, 29, 2016.

\bibitem[Ignatiev et~al.(2018)Ignatiev, Morgado, and
  Marques{-}Silva]{imms-sat18}
Ignatiev, A., Morgado, A., and Marques{-}Silva, J.
\newblock {PySAT:} {A} {Python} toolkit for prototyping with {SAT} oracles.
\newblock In \emph{SAT}, pp.\  428--437, 2018.
\newblock \doi{10.1007/978-3-319-94144-8_26}.
\newblock URL \url{https://doi.org/10.1007/978-3-319-94144-8_26}.

\bibitem[Ioffe \& Szegedy(2015)Ioffe and Szegedy]{ioffe2015batch}
Ioffe, S. and Szegedy, C.
\newblock {Batch normalization: Accelerating deep network training by reducing
  internal covariate shift}.
\newblock In \emph{{International conference on machine learning}}, pp.\
  448--456. PMLR, 2015.

\bibitem[Jia \& Rinard(2020)Jia and Rinard]{jia2020efficient}
Jia, K. and Rinard, M.
\newblock Efficient exact verification of binarized neural networks.
\newblock \emph{arXiv preprint arXiv:2005.03597}, 2020.

\bibitem[Jia \& Rinard(2021{\natexlab{a}})Jia and Rinard]{jia2021exploiting}
Jia, K. and Rinard, M.
\newblock Exploiting verified neural networks via floating point numerical
  error.
\newblock In \emph{International Static Analysis Symposium}, pp.\  191--205.
  Springer, 2021{\natexlab{a}}.

\bibitem[Jia \& Rinard(2021{\natexlab{b}})Jia and Rinard]{jia2021verifying}
Jia, K. and Rinard, M.
\newblock Verifying low-dimensional input neural networks via input
  quantization.
\newblock In \emph{International Static Analysis Symposium}, pp.\  206--214.
  Springer, 2021{\natexlab{b}}.

\bibitem[Katz et~al.(2017)Katz, Barrett, Dill, Julian, and
  Kochenderfer]{katz2017reluplex}
Katz, G., Barrett, C., Dill, D.~L., Julian, K., and Kochenderfer, M.~J.
\newblock Reluplex: An efficient smt solver for verifying deep neural networks.
\newblock In \emph{International Conference on Computer Aided Verification},
  pp.\  97--117. Springer, 2017.

\bibitem[Kingma \& Ba(2014)Kingma and Ba]{kingma2014adam}
Kingma, D.~P. and Ba, J.
\newblock Adam: A method for stochastic optimization.
\newblock \emph{arXiv preprint arXiv:1412.6980}, 2014.

\bibitem[Klambauer et~al.(2017)Klambauer, Unterthiner, Mayr, and
  Hochreiter]{klambauer2017selfnormalizing}
Klambauer, G., Unterthiner, T., Mayr, A., and Hochreiter, S.
\newblock {Self-Normalizing Neural Networks}, 2017.

\bibitem[Klir et~al.(1997)Klir, St.~Clair, and Yuan]{klir1997fuzzy}
Klir, G.~J., St.~Clair, U., and Yuan, B.
\newblock \emph{Fuzzy set theory: foundations and applications}.
\newblock Prentice-Hall, Inc., 1997.

\bibitem[Kurtz \& Bah(2021)Kurtz and Bah]{kurtz2021efficient}
Kurtz, J. and Bah, B.
\newblock Efficient and robust mixed-integer optimization methods for training
  binarized deep neural networks.
\newblock \emph{arXiv preprint arXiv:2110.11382}, 2021.

\bibitem[Lakkaraju et~al.(2016)Lakkaraju, Bach, and
  Leskovec]{lakkaraju2016interpretable}
Lakkaraju, H., Bach, S.~H., and Leskovec, J.
\newblock Interpretable decision sets: A joint framework for description and
  prediction.
\newblock In \emph{Proceedings of the 22nd ACM SIGKDD international conference
  on knowledge discovery and data mining}, pp.\  1675--1684, 2016.

\bibitem[Lee(1959)]{lee1959representation}
Lee, C.-Y.
\newblock Representation of switching circuits by binary-decision programs.
\newblock \emph{The Bell System Technical Journal}, 38\penalty0 (4):\penalty0
  985--999, 1959.

\bibitem[Liffiton \& Maglalang(2012)Liffiton and
  Maglalang]{liffiton2012cardinality}
Liffiton, M.~H. and Maglalang, J.~C.
\newblock A cardinality solver: More expressive constraints for free.
\newblock In \emph{International Conference on Theory and Applications of
  Satisfiability Testing}, pp.\  485--486. Springer, 2012.

\bibitem[Liu et~al.(2018)Liu, Wu, Luo, Yang, Liu, and Cheng]{liu2018bi}
Liu, Z., Wu, B., Luo, W., Yang, X., Liu, W., and Cheng, K.-T.
\newblock Bi-real net: Enhancing the performance of 1-bit cnns with improved
  representational capability and advanced training algorithm.
\newblock In \emph{Proceedings of the European conference on computer vision
  (ECCV)}, pp.\  722--737, 2018.

\bibitem[Lomuscio \& Maganti(2017)Lomuscio and Maganti]{lomuscio2017approach}
Lomuscio, A. and Maganti, L.
\newblock An approach to reachability analysis for feed-forward relu neural
  networks.
\newblock \emph{arXiv preprint arXiv:1706.07351}, 2017.

\bibitem[Lomuscio et~al.(2017)Lomuscio, Qu, and Raimondi]{lomuscio2017mcmas}
Lomuscio, A., Qu, H., and Raimondi, F.
\newblock Mcmas: an open-source model checker for the verification of
  multi-agent systems.
\newblock \emph{International Journal on Software Tools for Technology
  Transfer}, 19\penalty0 (1):\penalty0 9--30, 2017.

\bibitem[Loshchilov \& Hutter(2016)Loshchilov and Hutter]{loshchilov2016sgdr}
Loshchilov, I. and Hutter, F.
\newblock Sgdr: Stochastic gradient descent with warm restarts.
\newblock \emph{arXiv preprint arXiv:1608.03983}, 2016.

\bibitem[Madry et~al.(2017)Madry, Makelov, Schmidt, Tsipras, and
  Vladu]{madry2017towards}
Madry, A., Makelov, A., Schmidt, L., Tsipras, D., and Vladu, A.
\newblock Towards deep learning models resistant to adversarial attacks.
\newblock \emph{arXiv preprint arXiv:1706.06083}, 2017.

\bibitem[Manthey et~al.(2014)Manthey, Philipp, and Steinke]{manthey2014more}
Manthey, N., Philipp, T., and Steinke, P.
\newblock A more compact translation of pseudo-boolean constraints into cnf
  such that generalized arc consistency is maintained.
\newblock In \emph{Joint German/Austrian Conference on Artificial Intelligence
  (K{\"u}nstliche Intelligenz)}, pp.\  123--134. Springer, 2014.

\bibitem[Marques-Silva(2022)]{marques2022logic}
Marques-Silva, J.
\newblock Logic-based explainability in machine learning.
\newblock \emph{arXiv preprint arXiv:2211.00541}, 2022.

\bibitem[Meurer et~al.(2017)Meurer, Smith, Paprocki, \v{C}ert\'{i}k, Kirpichev,
  Rocklin, Kumar, Ivanov, Moore, Singh, Rathnayake, Vig, Granger, Muller,
  Bonazzi, Gupta, Vats, Johansson, Pedregosa, Curry, Terrel, Rou\v{c}ka, Saboo,
  Fernando, Kulal, Cimrman, and Scopatz]{10.7717/peerj-cs.103}
Meurer, A., Smith, C.~P., Paprocki, M., \v{C}ert\'{i}k, O., Kirpichev, S.~B.,
  Rocklin, M., Kumar, A., Ivanov, S., Moore, J.~K., Singh, S., Rathnayake, T.,
  Vig, S., Granger, B.~E., Muller, R.~P., Bonazzi, F., Gupta, H., Vats, S.,
  Johansson, F., Pedregosa, F., Curry, M.~J., Terrel, A.~R., Rou\v{c}ka, v.,
  Saboo, A., Fernando, I., Kulal, S., Cimrman, R., and Scopatz, A.
\newblock Sympy: symbolic computing in python.
\newblock \emph{PeerJ Computer Science}, 3:\penalty0 e103, January 2017.
\newblock ISSN 2376-5992.
\newblock \doi{10.7717/peerj-cs.103}.
\newblock URL \url{https://doi.org/10.7717/peerj-cs.103}.

\bibitem[Mirman et~al.(2018)Mirman, Gehr, and Vechev]{mirman2018differentiable}
Mirman, M., Gehr, T., and Vechev, M.
\newblock Differentiable abstract interpretation for provably robust neural
  networks.
\newblock In \emph{International Conference on Machine Learning}, pp.\
  3578--3586. PMLR, 2018.

\bibitem[Molnar(2020)]{molnar2020interpretable}
Molnar, C.
\newblock \emph{Interpretable machine learning}.
\newblock Lulu. com, 2020.

\bibitem[M{\"u}ller et~al.(2022)M{\"u}ller, Makarchuk, Singh, P{\"u}schel, and
  Vechev]{muller2022prima}
M{\"u}ller, M.~N., Makarchuk, G., Singh, G., P{\"u}schel, M., and Vechev, M.
\newblock Prima: general and precise neural network certification via scalable
  convex hull approximations.
\newblock \emph{Proceedings of the ACM on Programming Languages}, 6\penalty0
  (POPL):\penalty0 1--33, 2022.

\bibitem[Narodytska et~al.(2018)Narodytska, Kasiviswanathan, Ryzhyk, Sagiv, and
  Walsh]{narodytska2018verifying}
Narodytska, N., Kasiviswanathan, S., Ryzhyk, L., Sagiv, M., and Walsh, T.
\newblock Verifying properties of binarized deep neural networks.
\newblock In \emph{Proceedings of the AAAI Conference on Artificial
  Intelligence}, volume~32, 2018.

\bibitem[Narodytska et~al.(2019{\natexlab{a}})Narodytska, Shrotri, Meel,
  Ignatiev, and Marques-Silva]{narodytska2019assessing}
Narodytska, N., Shrotri, A., Meel, K.~S., Ignatiev, A., and Marques-Silva, J.
\newblock Assessing heuristic machine learning explanations with model
  counting.
\newblock In \emph{International Conference on Theory and Applications of
  Satisfiability Testing}, pp.\  267--278. Springer, 2019{\natexlab{a}}.

\bibitem[Narodytska et~al.(2019{\natexlab{b}})Narodytska, Zhang, Gupta, and
  Walsh]{narodytska2019search}
Narodytska, N., Zhang, H., Gupta, A., and Walsh, T.
\newblock In search for a sat-friendly binarized neural network architecture.
\newblock In \emph{International Conference on Learning Representations},
  2019{\natexlab{b}}.

\bibitem[Osoba \& Welser~IV(2017)Osoba and Welser~IV]{osoba2017intelligence}
Osoba, O.~A. and Welser~IV, W.
\newblock \emph{An intelligence in our image: The risks of bias and errors in
  artificial intelligence}.
\newblock Rand Corporation, 2017.

\bibitem[Paszke et~al.(2019)Paszke, Gross, Massa, Lerer, Bradbury, Chanan,
  Killeen, Lin, Gimelshein, Antiga, et~al.]{paszke2019pytorch}
Paszke, A., Gross, S., Massa, F., Lerer, A., Bradbury, J., Chanan, G., Killeen,
  T., Lin, Z., Gimelshein, N., Antiga, L., et~al.
\newblock Pytorch: An imperative style, high-performance deep learning library.
\newblock \emph{Advances in neural information processing systems},
  32:\penalty0 8026--8037, 2019.

\bibitem[Petersen et~al.(2022)Petersen, Borgelt, Kuehne, and
  Deussen]{petersen2022deep}
Petersen, F., Borgelt, C., Kuehne, H., and Deussen, O.
\newblock Deep differentiable logic gate networks.
\newblock \emph{arXiv preprint arXiv:2210.08277}, 2022.

\bibitem[Quine(1952)]{quine1952problem}
Quine, W.~V.
\newblock The problem of simplifying truth functions.
\newblock \emph{The American mathematical monthly}, 59\penalty0 (8):\penalty0
  521--531, 1952.

\bibitem[Quinlan(1986)]{quinlan1986induction}
Quinlan, J.~R.
\newblock Induction of decision trees.
\newblock \emph{Machine learning}, 1\penalty0 (1):\penalty0 81--106, 1986.

\bibitem[Quinlan(1987)]{quinlan1987simplifying}
Quinlan, J.~R.
\newblock Simplifying decision trees.
\newblock \emph{International journal of man-machine studies}, 27\penalty0
  (3):\penalty0 221--234, 1987.

\bibitem[Quinlan(2014)]{quinlan2014c4}
Quinlan, J.~R.
\newblock \emph{C4. 5: programs for machine learning}.
\newblock Elsevier, 2014.

\bibitem[Raghunathan et~al.(2018)Raghunathan, Steinhardt, and
  Liang]{raghunathan2018certified}
Raghunathan, A., Steinhardt, J., and Liang, P.
\newblock Certified defenses against adversarial examples.
\newblock \emph{arXiv preprint arXiv:1801.09344}, 2018.

\bibitem[Rastegari et~al.(2016)Rastegari, Ordonez, Redmon, and
  Farhadi]{rastegari2016xnor}
Rastegari, M., Ordonez, V., Redmon, J., and Farhadi, A.
\newblock Xnor-net: Imagenet classification using binary convolutional neural
  networks.
\newblock In \emph{European conference on computer vision}, pp.\  525--542.
  Springer, 2016.

\bibitem[Regulation(2016)]{regulation2016regulation}
Regulation, G. D.~P.
\newblock Regulation eu 2016/679 of the european parliament and of the council
  of 27 april 2016.
\newblock \emph{Official Journal of the European Union. Available at:
  http://ec. europa.
  eu/justice/data-protection/reform/files/regulation\_oj\_en. pdf (accessed 20
  September 2017)}, 2016.

\bibitem[Ribeiro et~al.(2016)Ribeiro, Singh, and Guestrin]{ribeiro2016should}
Ribeiro, M.~T., Singh, S., and Guestrin, C.
\newblock " why should i trust you?" explaining the predictions of any
  classifier.
\newblock In \emph{Proceedings of the 22nd ACM SIGKDD international conference
  on knowledge discovery and data mining}, pp.\  1135--1144, 2016.

\bibitem[Ribeiro et~al.(2018)Ribeiro, Singh, and Guestrin]{ribeiro2018anchors}
Ribeiro, M.~T., Singh, S., and Guestrin, C.
\newblock Anchors: High-precision model-agnostic explanations.
\newblock In \emph{Proceedings of the AAAI conference on artificial
  intelligence}, volume~32, 2018.

\bibitem[Rivest(1987)]{rivest1987learning}
Rivest, R.~L.
\newblock Learning decision lists.
\newblock \emph{Machine learning}, 2\penalty0 (3):\penalty0 229--246, 1987.

\bibitem[Roussel \& Manquinho(2009)Roussel and Manquinho]{roussel2009pseudo}
Roussel, O. and Manquinho, V.
\newblock Pseudo-boolean and cardinality constraints.
\newblock In \emph{Handbook of satisfiability}, pp.\  695--733. IOS Press,
  2009.

\bibitem[Rumelhart et~al.(1986)Rumelhart, Hinton, and
  Williams]{rumelhart1986learning}
Rumelhart, D.~E., Hinton, G.~E., and Williams, R.~J.
\newblock Learning representations by back-propagating errors.
\newblock \emph{nature}, 323\penalty0 (6088):\penalty0 533--536, 1986.

\bibitem[Sahoo et~al.(2020)Sahoo, Venugopalan, Li, Singh, and
  Riley]{sahoo2020scaling}
Sahoo, S.~S., Venugopalan, S., Li, L., Singh, R., and Riley, P.
\newblock Scaling symbolic methods using gradients for neural model
  explanation.
\newblock \emph{arXiv preprint arXiv:2006.16322}, 2020.

\bibitem[Sandler et~al.(2018)Sandler, Howard, Zhu, Zhmoginov, and
  Chen]{sandler2018mobilenetv2}
Sandler, M., Howard, A., Zhu, M., Zhmoginov, A., and Chen, L.-C.
\newblock Mobilenetv2: Inverted residuals and linear bottlenecks.
\newblock In \emph{Proceedings of the IEEE conference on computer vision and
  pattern recognition}, pp.\  4510--4520, 2018.

\bibitem[Shi et~al.(2020)Shi, Shih, Darwiche, and Choi]{shi2020tractable}
Shi, W., Shih, A., Darwiche, A., and Choi, A.
\newblock On tractable representations of binary neural networks.
\newblock \emph{arXiv preprint arXiv:2004.02082}, 2020.

\bibitem[Shih et~al.(2019)Shih, Darwiche, and Choi]{shih2019verifying}
Shih, A., Darwiche, A., and Choi, A.
\newblock Verifying binarized neural networks by local automaton learning.
\newblock In \emph{AAAI Spring Symposium on Verification of Neural Networks
  (VNN)}, 2019.

\bibitem[Tjeng et~al.(2019)Tjeng, Xiao, and Tedrake]{tjeng2017evaluating}
Tjeng, V., Xiao, K., and Tedrake, R.
\newblock Evaluating robustness of neural networks with mixed integer
  programming.
\newblock \emph{ICLR}, 2019.

\bibitem[Umans et~al.(2006)Umans, Villa, and Sangiovanni-Vincentelli]{1634621}
Umans, C., Villa, T., and Sangiovanni-Vincentelli, A.
\newblock {Complexity of two-level logic minimization}.
\newblock \emph{IEEE Transactions on Computer-Aided Design of Integrated
  Circuits and Systems}, 25\penalty0 (7):\penalty0 1230--1246, 2006.
\newblock \doi{10.1109/TCAD.2005.855944}.

\bibitem[Van Der~Walt et~al.(2011)Van Der~Walt, Colbert, and
  Varoquaux]{van2011numpy}
Van Der~Walt, S., Colbert, S.~C., and Varoquaux, G.
\newblock The numpy array: a structure for efficient numerical computation.
\newblock \emph{Computing in science \& engineering}, 13\penalty0 (2):\penalty0
  22--30, 2011.

\bibitem[Vaswani et~al.(2017)Vaswani, Shazeer, Parmar, Uszkoreit, Jones, Gomez,
  Kaiser, and Polosukhin]{vaswani2017attention}
Vaswani, A., Shazeer, N., Parmar, N., Uszkoreit, J., Jones, L., Gomez, A.~N.,
  Kaiser, {\L}., and Polosukhin, I.
\newblock Attention is all you need.
\newblock In \emph{Advances in neural information processing systems}, pp.\
  5998--6008, 2017.

\bibitem[Wang et~al.(2019)Wang, Davis, Cheung, and
  Constantinides]{wang2019lutnet}
Wang, E., Davis, J.~J., Cheung, P.~Y., and Constantinides, G.~A.
\newblock Lutnet: Rethinking inference in fpga soft logic.
\newblock In \emph{2019 IEEE 27th Annual International Symposium on
  Field-Programmable Custom Computing Machines (FCCM)}, pp.\  26--34. IEEE,
  2019.

\bibitem[Wang et~al.(2021{\natexlab{a}})Wang, Zhang, Xu, Lin, Jana, Hsieh, and
  Kolter]{wang2021beta}
Wang, S., Zhang, H., Xu, K., Lin, X., Jana, S., Hsieh, C.-J., and Kolter, J.~Z.
\newblock Beta-crown: Efficient bound propagation with per-neuron split
  constraints for complete and incomplete neural network verification.
\newblock \emph{arXiv preprint arXiv:2103.06624}, 2021{\natexlab{a}}.

\bibitem[Wang et~al.(2021{\natexlab{b}})Wang, Zhang, Liu, and
  Wang]{wang2021scalable}
Wang, Z., Zhang, W., Liu, N., and Wang, J.
\newblock Scalable rule-based representation learning for interpretable
  classification.
\newblock \emph{Advances in Neural Information Processing Systems}, 34,
  2021{\natexlab{b}}.

\bibitem[Wei et~al.(2019)Wei, Dash, Gao, and Gunluk]{wei2019generalized}
Wei, D., Dash, S., Gao, T., and Gunluk, O.
\newblock Generalized linear rule models.
\newblock In \emph{International Conference on Machine Learning}, pp.\
  6687--6696. PMLR, 2019.

\bibitem[Wong \& Kolter(2018)Wong and Kolter]{wong2018provable}
Wong, E. and Kolter, Z.
\newblock Provable defenses against adversarial examples via the convex outer
  adversarial polytope.
\newblock In \emph{International Conference on Machine Learning}, pp.\
  5286--5295. PMLR, 2018.

\bibitem[Wong et~al.(2018)Wong, Schmidt, Metzen, and Kolter]{wong2018scaling}
Wong, E., Schmidt, F.~R., Metzen, J.~H., and Kolter, J.~Z.
\newblock Scaling provable adversarial defenses.
\newblock \emph{arXiv preprint arXiv:1805.12514}, 2018.

\bibitem[Wu et~al.(2022)Wu, Barrett, Sharif, Narodytska, and
  Singh]{wu2022scalable}
Wu, H., Barrett, C., Sharif, M., Narodytska, N., and Singh, G.
\newblock Scalable verification of gnn-based job schedulers.
\newblock \emph{arXiv preprint arXiv:2203.03153}, 2022.

\bibitem[Xiao et~al.(2019)Xiao, Tjeng, Shafiullah, and Madry]{xiao2018training}
Xiao, K.~Y., Tjeng, V., Shafiullah, N. M.~M., and Madry, A.
\newblock Training for faster adversarial robustness verification via inducing
  relu stability.
\newblock In \emph{International Conference on Learning Representations}, 2019.

\bibitem[Xu et~al.(2020{\natexlab{a}})Xu, Shi, Zhang, Wang, Chang, Huang,
  Kailkhura, Lin, and Hsieh]{NEURIPS2020_0cbc5671}
Xu, K., Shi, Z., Zhang, H., Wang, Y., Chang, K.-W., Huang, M., Kailkhura, B.,
  Lin, X., and Hsieh, C.-J.
\newblock Automatic perturbation analysis for scalable certified robustness and
  beyond.
\newblock In Larochelle, H., Ranzato, M., Hadsell, R., Balcan, M., and Lin, H.
  (eds.), \emph{Advances in Neural Information Processing Systems}, volume~33,
  pp.\  1129--1141. Curran Associates, Inc., 2020{\natexlab{a}}.
\newblock URL
  \url{https://proceedings.neurips.cc/paper/2020/file/0cbc5671ae26f67871cb914d81ef8fc1-Paper.pdf}.

\bibitem[Xu et~al.(2020{\natexlab{b}})Xu, Zhang, Wang, Wang, Jana, Lin, and
  Hsieh]{xu2020fast}
Xu, K., Zhang, H., Wang, S., Wang, Y., Jana, S., Lin, X., and Hsieh, C.-J.
\newblock Fast and complete: Enabling complete neural network verification with
  rapid and massively parallel incomplete verifiers.
\newblock \emph{arXiv preprint arXiv:2011.13824}, 2020{\natexlab{b}}.

\bibitem[Yang et~al.(2021)Yang, He, Yang, Du, Yang, Yang, and
  Sun]{yang2021learning}
Yang, F., He, K., Yang, L., Du, H., Yang, J., Yang, B., and Sun, L.
\newblock Learning interpretable decision rule sets: A submodular optimization
  approach.
\newblock \emph{Advances in Neural Information Processing Systems}, 34, 2021.

\bibitem[Yu et~al.(2021)Yu, Ignatiev, Stuckey, and Le~Bodic]{yu2021learning}
Yu, J., Ignatiev, A., Stuckey, P.~J., and Le~Bodic, P.
\newblock Learning optimal decision sets and lists with sat.
\newblock \emph{Journal of Artificial Intelligence Research}, 72:\penalty0
  1251--1279, 2021.

\bibitem[Zhang et~al.(2021{\natexlab{a}})Zhang, Cai, Lu, He, and
  Wang]{zhang2021towards}
Zhang, B., Cai, T., Lu, Z., He, D., and Wang, L.
\newblock Towards certifying l-infinity robustness using neural networks with
  l-inf-dist neurons.
\newblock In \emph{International Conference on Machine Learning}, pp.\
  12368--12379. PMLR, 2021{\natexlab{a}}.

\bibitem[Zhang et~al.(2019)Zhang, Chen, Xiao, Gowal, Stanforth, Li, Boning, and
  Hsieh]{zhang2019towards}
Zhang, H., Chen, H., Xiao, C., Gowal, S., Stanforth, R., Li, B., Boning, D.,
  and Hsieh, C.-J.
\newblock Towards stable and efficient training of verifiably robust neural
  networks.
\newblock \emph{arXiv preprint arXiv:1906.06316}, 2019.

\bibitem[Zhang et~al.(2021{\natexlab{b}})Zhang, Ti{\v{n}}o, Leonardis, and
  Tang]{zhang2021survey}
Zhang, Y., Ti{\v{n}}o, P., Leonardis, A., and Tang, K.
\newblock A survey on neural network interpretability.
\newblock \emph{IEEE Transactions on Emerging Topics in Computational
  Intelligence}, 2021{\natexlab{b}}.

\bibitem[Zhang et~al.(2021{\natexlab{c}})Zhang, Zhao, Chen, Song, and
  Chen]{zhang2021bdd4bnn}
Zhang, Y., Zhao, Z., Chen, G., Song, F., and Chen, T.
\newblock Bdd4bnn: a bdd-based quantitative analysis framework for binarized
  neural networks.
\newblock In \emph{International Conference on Computer Aided Verification},
  pp.\  175--200. Springer, 2021{\natexlab{c}}.

\bibitem[Zhou et~al.(2016)Zhou, Wu, Ni, Zhou, Wen, and Zou]{zhou2016dorefa}
Zhou, S., Wu, Y., Ni, Z., Zhou, X., Wen, H., and Zou, Y.
\newblock Dorefa-net: Training low bitwidth convolutional neural networks with
  low bitwidth gradients.
\newblock \emph{arXiv preprint arXiv:1606.06160}, 2016.

\bibitem[Zwitter \& Soklic(1988)Zwitter and Soklic]{zwitter1988uci}
Zwitter, M. and Soklic, M.
\newblock Uci machine learning repository: Breast cancer data set, 1988.

\end{thebibliography}
\bibliographystyle{icml2023}

\appendix
\onecolumn
\section*{Appendix}

\section{Background \& Related Work}
\label{sec:ANN_Background}

\subsection{Definitions and notations} 
\label{sec:ANN_Definitions}

First, we introduce the definitions of a DNN verifiably robust, soundness and completeness verification methods, and the verified accuracy. 

We define a dataset $\mathcal{D}$ and the subsets $\mathcal{D}_1,\mathcal{D}_2$ of $\mathcal{D}$ such that $\mathcal{D}_1 \cup \mathcal{D}_2 = \mathcal{D}$ and $ \mathcal{D}_1 \cap \mathcal{D}_2 = \emptyset$. We define $F_{\theta}$ as the DNN with parameters $\theta$ and $\mathcal{P}$ as the property. We consider that for all inputs from $\mathcal{D}_1$ we have that $\mathcal{P}$ holds, whereas for all inputs from $\mathcal{D}_2$, $\mathcal{P}$ does not hold. In our case, we only consider the verifiably robust DNN as the property $\mathcal{P}$, which is defined as follows.

\smallskip

\myparagraph{Definition: Verifiably robust property.} \textit{Given an input $x$ and its label $y$ from the dataset $\mathcal{D}$, a norm $p$, a DNN $F_{\theta}$ is \textbf{verifiably robust} on input $x$ at the given noise level $\epsilon >0$ if for $||x-x'||_p < \epsilon$ one has $F_{\theta}(x)= F_{\theta}(x') = y$ for all $x'$.}

\smallskip

\myparagraph{Definition: Sound, complete, verifiable, certification, robustness methods and accuracies.}
\textit{Given a dataset $\mathcal{D}$ and a DNN $F_{\theta}$, the \textbf{natural accuracy} is the ratio of data $(x,y) \in \mathcal{D}$ such that $F_{\theta}(x)=y$. For a method $\mathcal{M}$ and a property $\mathcal{P}$, we define various type of methods as follows: }

\begin{itemize}
    \item \textit{if for all inputs in $\mathcal{D}_2$, $\mathcal{M}$ states that $\mathcal{P}$ does not hold then $\mathcal{M}$ is \textbf{sound}.}
    \item \textit{if for all inputs in $\mathcal{D}_1$, $\mathcal{M}$ states that $\mathcal{P}$ does hold then $\mathcal{M}$ is \textbf{complete}.}
    \item \textit{if there exists an input in $\mathcal{D}_2$ such that $\mathcal{M}$ states that $\mathcal{P}$ holds or $\mathcal{M}$ never halts, then $\mathcal{M}$ is \textbf{unsound}.}
    \item  \textit{if there exist an input in $\mathcal{D}_1$ such that $\mathcal{M}$ states that $\mathcal{P}$ does not hold or $\mathcal{M}$ never halts, $\mathcal{M}$ is \textbf{incomplete}.}
    \item \textit{\textbf{a verifiable method $\mathcal{M}_V$} is a sound and complete method. The \textbf{verified accuracy} is the ratio of data $(x,y) \in \mathcal{D}$ such that $F_{\theta}(x)=y$ and $\mathcal{M}_V$ states that $\mathcal{P}$ holds. }
    \item \textit{\textbf{a certification method $\mathcal{M}_C$} is a sound and incomplete method. The \textbf{certification accuracy} is the ratio of data $(x,y) \in \mathcal{D}$ such that $F_{\theta}(x)=y$ and $\mathcal{M}_C$ states that $\mathcal{P}$ holds. }
    \item \textit{\textbf{a robust method $\mathcal{M}_R$} is an unsound method. The \textbf{robust accuracy} is the ratio of data $(x,y) \in \mathcal{D}$ such that $F_{\theta}(x)=y$ and $\mathcal{M}_R$ states that $\mathcal{P}$ holds. }
\end{itemize}

We also define the method $\mathcal{M}$ of verification of the property $\mathcal{P}$ on $F_{\theta}$ for the dataset $\mathcal{D}$. $\mathcal{M}$ can use an exact solver (like a SAT solver), a probabilistic solver, or no solver at all (as in the case of PGD attacks~\cite{madry2017towards}). $\mathcal{M}$ has three different possible outcomes: $\mathcal{P}$ holds, $\mathcal{P}$ does not hold, or $\mathcal{M}$ never halts. When using solvers and if the method never halts, the solvers will output a timeout. 

Verifiable methods can be mainly divided into SMT-based methods \cite{katz2017reluplex}, Mixed-Integer Linear Programming MILP-based methods \cite{xiao2018training}, or SAT-based methods. There are few works and the gap between natural and verified accuracy remains important. Certification methods try to fill this gap by proposing a trade-off between security and performance. In contrary, the literature concerning robustness methods is large: we refer to \cite{croce2021robustbench,carlini2019evaluating} that compares and benchmarks these methods. 

In this work, $F_{\theta}$ is the $\mathcal{TT}$net model, $\mathcal{D}$ is either MNIST or CIFAR-10 and we only consider the infinity norm ($p = \infty$). Furthermore, the noise level $\epsilon$ is $\{0.1,0.3\}$ for MNIST and $\{2/255, 8/255\}$ for CIFAR-10. Our method $\mathcal{M}_V$ is verified as it is sound and complete. We use the exact and generic SAT solver MiniCard \cite{liffiton2012cardinality}, in contrast to a probabilistic one as in \cite{zhang2021bdd4bnn} or a handcrafted one as in \cite{jia2020efficient}).

\subsection{Background on Boolean SATisfiability} The SAT problem \cite{biere2009handbook} is that of deciding whether there exists a variable assignment to satisfy a given Boolean expression $ \Phi $. We can consider a Boolean expression in a Conjunctive Normal Form (CNF) or in a Disjunctive Normal Form (DNF). They are both defined over a set of Boolean variables $(x_1, \cdots ,x_n)$. A literal $l_i$ is defined as a variable $x_i$ or its complement $\overline{x_i}$. A CNF is a conjunction of a set of clauses: $\Phi=(c_1 \land \cdots \land c_m)$, where each clause $c_j$ is a disjunction of some literals $c_j = l_{j1} \lor \cdots \lor l_{jr}$. A DNF is a disjunction of a set of clauses: $\Phi=(c_1 \lor \cdots \lor c_m)$, where each clause $c_j$ is a conjunction of some literals $c_j = l_{j1} \land \cdots \land l_{jr}$. A pseudo-Boolean constraint has the form: $\textstyle \sum_{p=1}^N a_p l_p \circ b$, where $a_p \in \mathbb{Z}$, $b \in \mathbb{Z}$ and $\circ \in \{\leq,=,\geq\}$, which can be mapped to a SAT formula \cite{roussel2009pseudo}. However, such a conversion generally leads to a tremendous number of clauses and literals compared to the number of variables in the original pseudo-Boolean form, making it impractical to comprehend. For an illustration of the latter, see Appendix~\ref{ExampleEncoding} with examples of inequality conversion into SAT formulas according to different methodologies.

\subsection{Background on SAT encoding of neural networks}
The sole published method converting a DNN into a SAT formula is limited to BNNs \cite{narodytska2018verifying,cheng2018verification} and involves recomposing a block formed of a Two-Dimensional Convolutional Neural Network (2D-CNN) layer, a batch normalization layer and a step function into an inequality in order to apply the pseudo-Boolean constraint \cite{roussel2009pseudo}. This approach has been further refined using a different training method and a specific SAT solver, resulting in a significantly reduced verification time \cite{jia2020efficient}. Although the proposed inequality rewriting is elegant, the corresponding SAT formula still contains a large number of clauses and literals compared to the number of variables in the pseudo-Boolean constraint. This prevents the tractability of those SAT/BNNs formulas. In this work, we will only focus on comparison with DNN models, and not other machine learning family such as \cite{andriushchenko2019provably}. The main feature of DNNs is that there are more parameters in the model than samples in the dataset. This is not the case with decision trees. Their learning capacity is very different and for example, we are able to learn on ImageNET, in contrary to them. 

\subsection{Background on sound, complete verification and robustness}
\label{Completeverification} We provide in Appendix~\ref{sec:ANN_Definitions} the definition of completeness, soundness, etc. Complete and sound property verification of SAT-convertible DNNs has been presented in \cite{narodytska2018verifying} as follows: given a precondition $prec$ on the inputs $x$, a property $prop$ on the outputs $o$ and a SAT relations given by a DNN between inputs/outputs denoted as $DNN(x,o)$, we check whether the following statement is valid: $prec(x) \land DNN(x,o) \implies prop(o) $. In order to seek a counter-example to this property, we look for a satisfying assignment of $prec(x) \land DNN(x,o) \land \overline{prop(o)}$. An application example of property verification is to check for the existence of an adversarial perturbation in a trained DNN. In this case, $prec$ defines an $\epsilon$-ball of valid perturbations around the original image and $prop$ states that the classification should not change under these small perturbations. Therefore, we distinguish the traditional "natural accuracy" from the "verified accuracy", the latter measuring the fraction of the predictions which remains correct for all adversarial attacks within the perturbation constraints. 

\subsection{Background on 2-dimensional convolutional neural networks}
\label{sec:Background_P:CNN}

A Boolean function has the form $\{0,1\}^n \rightarrow \{0,1\}$ and its corresponding truth table lists the outputs for all $2^n$ possible inputs combinations (easily set up when $n$ is not too large). Within our method, we consider the 2D-CNN as a Boolean function $\Phi_{(f,s,p,k,c)}$ which, for a given filter $f$, takes $n = k^2 c/g$ inputs at position $(i,j) $ with $k$ the kernel size, $c$ the number of input channels, $s$ the stride, $p$ the padding and $g$ the group parameter \cite{dumoulin2016guide}. The outputs can be written as $y_f ^{(i,j)} = \Phi_f(x_1^{(i,j)}, \cdots, x_{n}^{(i,j)})$. If we now consider a multi-layer network, a similar truth table can be constructed, except for the kernel size $k$ that needs to be replaced by a \emph{patch function}, sometimes also referred to as the size of a receptive field \cite{araujo2019computing}. 
We denote the vector obtained after the flatten operation and before the final classifier layer as the vector of features $V$.

\subsection{Background on global exact interpretability}

Lack of explainability and difficulty in integrating human knowledge are well-known concerns for Deep Neural Networks (DNNs) and more generally for all ensemble ML models due to their large complexity~\cite{zhang2021survey, he2020extract,ribeiro2016should, ribeiro2018anchors, molnar2020interpretable, fryer2021shapley}. Therefore, the global and exact interpretability of these systems is the subject of intense research efforts, especially for safety-critical applications~\cite{driscoll2020system, regulation2016regulation, osoba2017intelligence}. 

In practice, rule-based models~\cite{freitas2014comprehensible}, like tree-based models~\cite{quinlan1986induction,quinlan1987simplifying, bessiere2009minimising}, can easily provide global explanations, i.e. independent of the input, and exact, i.e. giving the same output as the model. However, despite their intrinsic interpretable nature, their performances are lower than those of other less interpretable family models including DNN or ensemble ML models such as Random Forest~\cite{breiman2001random}.

Rule-based models produce predicates for the decision that are expressed in DNF. For example, in the Adult Census dataset \cite{Dua:2019} one rule of the model for determining whether an individual would be reaching yearly earnings exceeding 50K\$/year might be: \vspace{-0.2cm}
$$((\text{Age}>34) \land \text{Maried}) \lor (\text{Male} \land (\text{Capital Loss} < \text{1k/year})) \lor ((\text{Age}<34) \land \text{Go to University})$$
Despite the large number of works published on DNN interpretability~\cite{zhang2021survey, he2020extract}, few can provide as strong interpretations as those from rule-based models. Rules are particularly suited for tabular data that contain mixed-type features and exhibit complex high-order feature interactions. The quality assessment of a model is usually based on three criteria: a) the performance, b) the number of rules and c) the size of each of the rules~\cite{freitas2014comprehensible}.

\subsection{Related work on formal verification}

As application, we study Deep Convolutional Neural Networks (DCNNs) from the standpoint of their complete and sound formal verification: in other words, knowing a certain property, we want to confirm whether this property holds or not for a specific DNN. DNNs complete and sound verification property methods are mainly based either on Satisfiability Modulo Theory (SMT) \cite{katz2017reluplex} or Mixed-Integer Programming (MIP) \cite{xiao2018training} which are not yet scalable to real-valued DNNs. Some recent publications \cite{jia2020efficient,narodytska2019search,narodytska2018verifying} approached the problem of complete verification from the well-known Boolean SATisfiability (SAT) \cite{biere2009handbook} point of view where BNNs \cite{hubara2016binarized} are first converted into SAT formulas and then formally verified using SAT. This pipeline is computationally efficient \cite{jia2020efficient}, enables security verification \cite{baluta2019quantitative} and more generally can answer a large range of questions including how many adversarial attacks exist for a given BNN, image and noise level \cite{narodytska2019assessing}. However, to date, only BNNs can be transformed into a SAT formula and, as they were not designed for this application, their corresponding SAT conversion method intrinsically leads to formulas with a large number of variables and clauses, impeding formal verification scalability. In \cite{jia2020efficient}, the authors developed their SAT solver to achieve verification scaling. 

A well-known example of property to verify on image classification datasets is the robustness to adversarial attacks. The performance of formal DNN robustness verification methods is evaluated against two main characteristics: verified accuracy (i.e. the ratio of images that are correctly predicted and that do not have any adversarial attacks) and verification time (i.e. the duration to verify that one correctly predicted image in the test set does/does not have an adversarial attack). The verification time is the sum of the problem construction time and its resolution time. The latter is intimately related to the quality of the solver and the complexity of the DNN encoding: a less efficient encoding and the solver will lead to a longer time to verify an image. Most robustness improvements in the literature have been in the form of new trainings \cite{carlini2017provably,wong2018provable,wong2018scaling,dvijotham2018training,raghunathan2018certified,mirman2018differentiable}, new testing method to increase robustness \cite{carlini2017provably,sahoo2020scaling,dvijotham2018training}, new certification method to guarantee robustness properties \cite{sahoo2020scaling,wong2018provable,wong2018scaling,dvijotham2018training} or new formal verification methods \cite{ehlers2017formal,lomuscio2017approach,cheng2017maximum,mirman2018differentiable,shih2019verifying,wu2022scalable,lomuscio2017mcmas}. Table~\ref{table:SOA} presents a comparison of the functionalities of current state-of-the-art sound and complete verification methods. For a more detailed state-of-the-art, we refer to \cite{tutorialAAAI}.

\begin{table}[htb!]
\centering
\caption{Comparison of state-of-the-art formal verification methods according to four criteria: the method type (sound and complete), the method applicability scope (can the method extend to all properties and all DNNs), the solver used and the method scalability. }
  \label{table:SOA}
  \centering
  \renewcommand\arraystretch{1.}
  \resizebox{1.\columnwidth}{!}{
\begin{tabular}{|l|c|cc|c|cc|}
\toprule
\multicolumn{1}{|c|}{\multirow{3}{*}{\textbf{Method}}} & \textbf{Type}                                & \multicolumn{2}{c|}{\textbf{Applicability scope }}                      & \textbf{Solver Used}                & \multicolumn{2}{c|}{\textbf{Scalability}}           \\ \cmidrule(l){2-7} 
                        & \multirow{2}{*}{Sound \quad Complete} & All &  All  & \multirow{2}{*}{SAT/MIP} & MNIST               & CIFAR-10            \\ \cmidrule(l){6-7} 
                        &                                     &              Properties  &  DNNs              &                            & Low / High        & Low / High        \\ \midrule
\cite{jia2020efficient}                       & $\surd$ \quad\quad\quad $\surd$                 & $\surd$ & only BNN                       & specific SAT                        & $\surd$ \quad  $\surd$ & $\surd$  \quad $\surd$ \\
\cite{narodytska2019assessing}                          & $\surd$ \quad\quad\quad $\surd$                 & $\surd$ & only BNN                       & specific SAT                        & \hspace*{-0.42cm} only $l_p$ \quad {\hspace*{-0.4cm}  $\times$} & $\times$  \quad $\times$ \\
\cite{xiao2018training}                         & $\surd$ \quad\quad\quad $\surd$                 & $\surd$ & $\surd$                       & MIP                        & $\surd$ \quad $\surd$ & $\surd$ \quad $\surd$ \\
\cite{tjeng2017evaluating}                         & $\surd$ \quad\quad\quad $\surd$                 & $\surd$ & $\surd$                       & MIP                        & $\surd$ \quad $\surd$ & $\surd$ \quad $\surd$ \\
\cite{muller2022prima}                         & $\surd$ \quad\quad\quad $\times$                 & $\times$ & $\surd$                       & MIP                        & $\surd$ \quad $\surd$ & $\surd$ \quad $\surd$ \\
\cite{wang2021beta}                         & $\surd$ \quad\quad\quad $\surd$                 & $\surd$ & $\surd$                       & MIP                        & $\surd$ \quad $\surd$ & $\surd$ \quad $\surd$ \\
\cite{NEURIPS2020_0cbc5671}                         & $\surd$ \quad\quad\quad $\times$                 & $\times$ & $\surd$                       & -                        & $\times$ \quad $\times$ & $\times$ \quad $\times$ \\
\cite{zhang2021towards}                         & $\surd$ \quad\quad\quad $\times$                 & $\times$ & $\surd$                       & -                        & $\times$ \quad $\surd$ & $\times$ \quad $\surd$ \\
\cite{zhang2019towards}                         & $\surd$ \quad\quad\quad $\times$                 & $\times$ & $\surd$                       & -                        & $\surd$ \quad $\surd$ & $\surd$ \quad $\surd$ \\
\cite{andriushchenko2019provably}                         & $\surd$ \quad\quad\quad $\surd$                 & $\times$ & only dec. tree                       & -                        & $\times$ \quad $\surd$ & $\times$ \quad $\surd$ \\
\cite{kurtz2021efficient}                         & $\surd$ \quad\quad\quad $\surd$                 & $\times$ & only BNN                         & MIP                        & $\times$ \quad $\times$ & $\times$ \quad $\times$ \\
\cite{jia2021verifying}                         & $\surd$ \quad\quad\quad $\surd$                 & $\surd$ & $\surd$                       & MIP                        & $\times$ \quad $\times$ & $\times$ \quad $\times$ \\
\cite{zhang2021bdd4bnn}                         & $\surd$ \quad\quad\quad $\surd$                 & $\surd$ & only BNN                         & MIP$^{*}$                        & $\surd$ \quad $\surd$ & $\times$ \quad $\times$ \\
\midrule
Ours                       & $\surd$ \quad\quad\quad $\surd$                 & $\surd$ & only $\mathcal{TT}$net                       & general SAT                        & $\surd$ \quad $\surd$ & $\surd$ \quad $\surd$ \\
\bottomrule
\end{tabular}}
\smallskip
\parbox[t]{\textwidth}{\tiny
\textit{$^*$Probabilistic solver.}
}s
\end{table}

\subsection{Related Work on global exact interpretability} Along decision trees \cite{quinlan1986induction, quinlan1987simplifying, bessiere2009minimising}, rules lists \cite{rivest1987learning, angelino2017learning, dash2018boolean} or bayesian networks \cite{friedman1997bayesian, freitas2014comprehensible}, rules sets \cite{lakkaraju2016interpretable, cohen1995fast, cohen1999simple, quinlan2014c4, wei2019generalized} are part of the rule-based models family. In rules sets, each rule encodes a different class and all have the same weight. Therefore, our work may be classified in the rules sets field. Interestingly, in \cite{lakkaraju2016interpretable}, the authors show that rules sets are ultimately more interpretable than rules lists as they are easier to infer. Yet, to the best of our knowledge, none of the previous rules sets models managed to propose rules for more complex tasks than tabular datasets classification, like image datasets classification, for example~\cite{yang2021learning, wang2021scalable}. Moreover, most of them handle only binary classification.

\subsection{Examples of pseudo-Boolean constraint encoding}
\label{ExampleEncoding}

Table \ref{table:SATconv} presents a few examples of pseudo-Boolean constraint encoding.

\begin{table*}[htb!]
\caption{\label{table:SATconv} Examples of inequality conversion into SAT formulas according to different methodologies. The first row presents one inequality example containing 3 natural variables $(x_1, x_2, x_3)$. The corresponding output SAT encodings are given for 5 different published methods \cite{abio2011bdds,holldobler2012compact,een2006translating,manthey2014more}. The SAT equation outputs include many literals and clauses, adding complexity to the problem. Moreover, there is no straightforward relationship between the SAT literals $l_i$ and the variables $x_i$, or between the clauses and the inequality coefficients.\strut}
\resizebox{\columnwidth}{!}{
\begin{tabular}{@{}|l|c@{}|}
\toprule
\textbf{SAT encoding and source} & \textbf{Inequality to convert into SAT Formulas: $x_1 - 2 x_2 + 3 x_3 \leq 3 $} \\ \midrule
\textbf{Encoding 1} \cite{abio2011bdds}                           &                              $(l_4) \land (\overline{l_1} \lor l_2 \lor \overline{l_5}) \land (l_5 \lor \overline{l_3} \lor \overline{l_6}) \land (l_6)$ \\ \midrule
\textbf{Encoding 2} \cite{holldobler2012compact}                           &                    

\begin{tabular}[c]{@{}c@{}}

 $(\overline{l_4} \lor l_9) \land (\overline{l_5} \lor l_{10}) \land (\overline{l_6} \lor l_{11})$ \\ $ \land (\overline{l_7} \lor l_{12}) \land (\overline{l_8} \lor l_{13}) \land (\overline{l_9} \lor l_{14}) \land (\overline{l_{10}} \lor l_{15}) \land (\overline{l_{11}} \lor l_{16}) $ \\ $\land (\overline{l_{12}} \lor l_{17}) \land (\overline{l_{13}} \lor l_{18}) \land$ \\ $ (\overline{l_3} \lor l_{4}) \land (\overline{l_3} \lor l_{5}) \land (\overline{l_3} \lor l_{6}) \land$ \\ $ (l_{2} \lor l_{9}) \land (l_{2} \lor l_{10}) \land (\overline{l_1} \lor 14) \land (\overline{l_4} \lor l_{2} \lor l_{11}) \land $ \\ $(\overline{l_5} \lor l_{2} \lor l_{12}) \land (\overline{l_6} \lor l_{2} \lor l_{13}) \land (\overline{l_9} \lor \overline{l_1} \lor l_{15}) \land$ \\ $ (\overline{l_{10}} \lor \overline{l_1} \lor l_{16}) \land (\overline{l_11} \lor \overline{l_1} \lor l_{17}) \land (\overline{l_{12}} \lor \overline{l_1} \lor l_{18}) \land (\overline{l_7}) \land$ \\ $ (\overline{l_8}) \land (\overline{l_7} \lor l_{2}) \land (\overline{l_{13}} \lor \overline{l_1})$ 
 \end{tabular} \\ \midrule
\textbf{Encoding 3} \cite{een2006translating}                        &                              $( l_{5} \lor \overline{l_3} \lor  l_{2}) \land ( l_{7} \lor \overline{l_3} \lor \overline{l_1}) \land (\overline{l_8} \lor \overline{l_3}) \land (\overline{l_8} \lor  l_{2}) \land ( l_{6} \lor  l_{8} \lor \overline{l_7}) \land ( l_{4} \lor \overline{l_6} \lor \overline{l_5}) \land (\overline{l_4})$ \\ \midrule
\textbf{Encoding 4} \cite{een2006translating}                                    &                              
\begin{tabular}[c]{@{}c@{}}

$(\overline{l_3} \lor \overline{l_1} \lor \overline{l_4}) \land (l_{3} \lor l_{1} \lor \overline{l_4}) \land (\overline{l_3} \lor l_{1} \lor l_{4}) \land (l_{3} \lor \overline{l_1} \lor l_{4}) \land (l_{3} \lor \overline{l_5}) \land $ \\ $ (l_{1} \lor \overline{l_5}) \land (\overline{l_3} \lor \overline{l_1} \lor l_{5}) \land (l_{3} \lor \overline{l_2} \lor l_{5} \lor \overline{l_6}) \land (l_{3} \lor l_{2} \lor \overline{l_5} \lor \overline{l_6}) \land $ \\ $(\overline{l_3} \lor \overline{l_2} \lor \overline{l_5} \lor \overline{l_6}) \land (\overline{l_3} \lor l_{2} \lor l_{5} \lor \overline{l_6}) \land  (\overline{l_3} \lor l_{2} \lor \overline{l_5} \lor l_{6}) $ \\ $\land (\overline{l_3} \lor \overline{l_2} \lor l_{5} \lor l_{6}) \land (l_{3} \lor l_{2} \lor l_{5} \lor l_{6}) \land (l_{3} \lor \overline{l_2} \lor \overline{l_5} \lor l_{6}) \land (\overline{l_2} \lor l_{5} \lor \overline{l_7}) \land (l_{3} \lor l_{5} \lor \overline{l_7}) \land (l_{3} \lor \overline{l_2} \lor \overline{l_7}) $ \\ $\land (l_2 \lor \overline{l_5} \lor l_7) \land (\overline{l_3} \lor \overline{l_5} \lor l_7) \land (\overline{l_3} \lor l_2 \lor l_7) \land (\overline{l_7} \lor \overline{l_6} \lor l_3) \land $ \\ $ (\overline{l_7} \lor \overline{l_6} \lor \overline{l_2}) \land (\overline{l_7} \lor \overline{l_6} \lor l_5) \land (l_7 \lor l_6 \lor \overline{l_3}) \land (l_7 \lor l_6 \lor l_2) \land (l_7 \lor l_6 \lor \overline{l_5}) \land (\overline{l_7} \lor \overline{l_6})
$

\end{tabular} 

 \\\midrule
\textbf{Encoding 5} \cite{manthey2014more}                       &                             

\begin{tabular}[c]{@{}c@{}}
 $(l_4) \land (\overline{l_3} \lor l_5) \land (\overline{l_1} \lor l_5) \land (\overline{l_3} \lor \overline{l_1} \lor l_6) \land (\overline{l_3} \lor l_7) \land (l_2 \lor l_7) \land (\overline{l_3} \lor l_2 \lor l_8) \land (\overline{l_7} \lor l_9) \land $ \\ $ (\overline{l_8} \lor l_{10}) \land (\overline{l_6} \lor l_9) \land (\overline{l_7} \lor \overline{l_6} \lor l_{10}) \land (\overline{l_8} \lor \overline{l_6} \lor l_{11}) \land (\overline{l_11})$ 

\end{tabular}  
 \\ \bottomrule
\end{tabular}}
\end{table*}

\section{$\mathcal{TT}$net complementary details}
\label{sec:ANN_TTnet}

\subsection{General architecture}

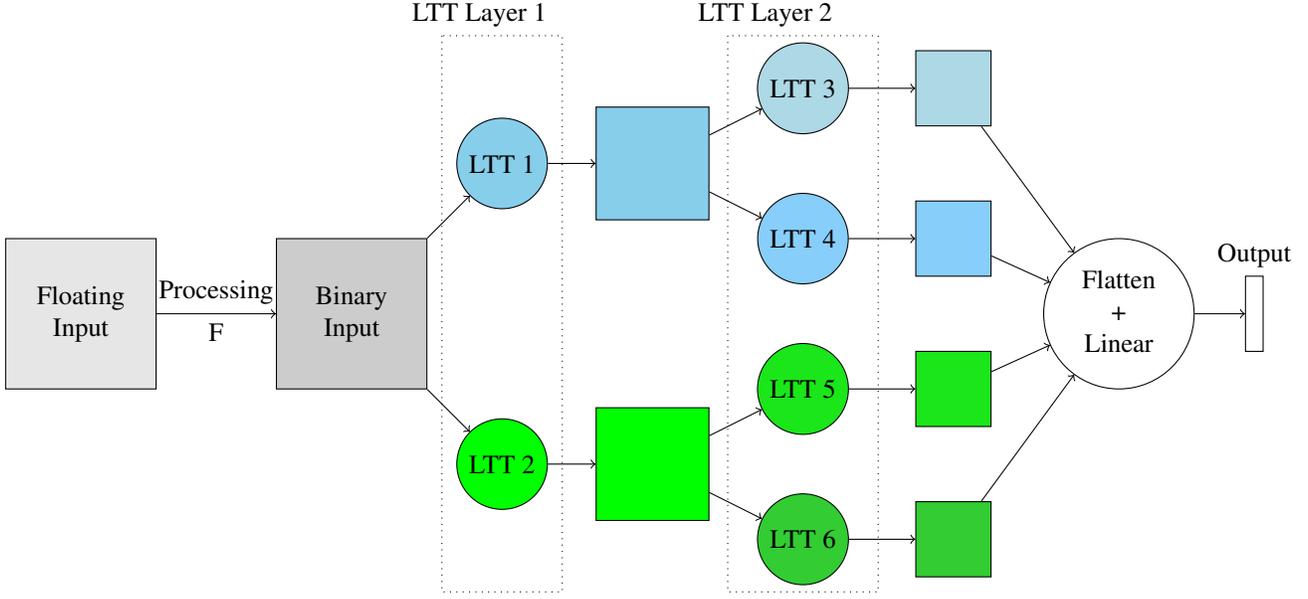
\begin{figure*}[htb!]
    \centering
\begin{tikzpicture}

\definecolor{red1}{rgb}{1,0,0}
\definecolor{red2}{rgb}{0.9,0.1,0.1}
\definecolor{red3}{rgb}{0.8,0.2,0.2}
\definecolor{blue1}{HTML}{87CEEB}
\definecolor{blue2}{HTML}{ADD8E6}
\definecolor{blue3}{HTML}{87CEFA}
\definecolor{green1}{rgb}{0,1,0}
\definecolor{green2}{rgb}{0.1,0.9,0.1}
\definecolor{green3}{rgb}{0.2,0.8,0.2}

\definecolor{gray1}{gray}{0.9}
\definecolor{gray2}{gray}{0.8}

\draw[dotted] (1,3.7) rectangle (3,-3.7);
\draw[dotted] (-2.8,3.7) rectangle (-1.2,-3.7);
\node[above] at (1.5,3.7) {LTT Layer 2};
\node[above] at (-2.3,3.7) {LTT Layer 1};

\node[draw,rectangle, minimum width=2cm,minimum height=2cm,align=center,fill=gray1] (square) at (-7.6,0) {\begin{tabular}{@{}c@{}}
        Floating \\ Input
    \end{tabular}};
\node[draw,rectangle, minimum width=2cm,minimum height=2cm,align=center, fill=gray2] (square2) at (-4,0) {\begin{tabular}{@{}c@{}}
        Binary \\ Input
    \end{tabular}};
\node[draw,circle,minimum width=1cm,align=center,fill=blue1] (circle1) at (-2,2) {LTT 1};
\node[draw,circle,minimum width=1cm,align=center,fill=green1] (circle3) at (-2,-2) {LTT 2};
\node[draw,minimum width=1.5cm,minimum height=1.5cm,align=center,fill=blue1] (square3) at (0,2) {};
\node[draw,minimum width=1.5cm,minimum height=1.5cm,align=center,fill=green1] (square5) at (0,-2) {};

\node[draw,circle,minimum width=1cm,align=center,fill=blue2] (circle4) at (2,3) {LTT 3};
\node[draw,circle,minimum width=1cm,align=center,fill=blue3] (circle5) at (2,1) {LTT 4};
\node[draw,circle,minimum width=1cm,align=center,fill=green2] (circle8) at (2,-1) {LTT 5};
\node[draw,circle,minimum width=1cm,align=center,fill=green3] (circle9) at (2,-3) {LTT 6};

\node[draw,minimum width=1cm, minimum height=1cm,align=center,fill=blue2] (square6) at (4,3) {};
\node[draw,minimum width=1cm,minimum height=1cm,align=center,fill=blue3] (square7) at (4,1) {};
\node[draw,minimum width=1cm,minimum height=1cm,align=center,fill=green2] (square10) at (4,-1) {};
\node[draw,minimum width=1cm,minimum height=1cm,align=center,fill=green3] (square11) at (4,-3) {};

\node[draw,circle, minimum width=1cm,minimum height=2cm,align=center] (circle10) at (6.2,0) {\begin{tabular}{@{}c@{}}
        Flatten \\ + \\ Linear
    \end{tabular}};

\node[draw,minimum width=0.05cm,minimum height=1cm,align=center] (square12) at (8,0) {};
\node[above] at (8,0.5) {Output};

\draw[->] (square) -- node [above] {Processing} node [below] {F} (square2);

\draw[->] (square2) -- (circle1);
\draw[->] (square2) -- (circle3);

\draw[->] (circle1) -- (square3);
\draw[->] (circle3) -- (square5);

\draw[->] (square3) -- (circle4);
\draw[->] (square3) -- (circle5);
\draw[->] (square5) -- (circle8);
\draw[->] (square5) -- (circle9);

\draw[->] (circle4) -- (square6);
\draw[->] (circle5) -- (square7);
\draw[->] (circle8) -- (square10);
\draw[->] (circle9) -- (square11);

\draw[->] (square6) -- (circle10);
\draw[->] (square7) -- (circle10);
\draw[->] (square10) -- (circle10);
\draw[->] (square11) -- (circle10);

\draw[->] (circle10) -- (square12);

\end{tikzpicture}
    \caption{\label{fig:archi} General architecture of the  $\mathcal{TT}$net model with a one-channel input. Layer 0 is a pre-processing layer that allows image binarization. Then follow two layers of 2D Learning Truth Table (LTT) blocks with group $g=1$: two blocks in the first layer, four in the second. It should be noted that the LTT block of layer 2 does not take as input all the filters of layer 1, as it is usually the case: it only takes the filter of their groups. Finally, the last linear layer performs the classification. Depending on the application the processing $F$ and the final Linear Layer can change. }
\end{figure*}

Figure~\ref{fig:archi} gives the general architecture for a 2D-$\mathcal{TT}$net. In this Figure~\ref{fig:archi}, we can observe that  $\mathcal{TT}$net is similar to a Neural Additive Model (NAM)~\cite{agarwal2020neural}: like  $\mathcal{TT}$net, NAM is a sum of non-linear functions with small input size: 1 for NAM and $n\leq16$ for  $\mathcal{TT}$net. The main difference is that NAM is built for plotting only one continuous variable function, while we are built for multiple discrete variable functions.

\subsection{First layer: pre-processing layer}
\label{sec:App_1stlayer}

Before applying the batch normalisation and the step function, we quantify the inputs as $x^{q} = \lfloor \frac{x}{q} \rfloor \cdot s$ where $x$ is the real-valued input, $x^{q}$ is the quantized input to be fed into the $\mathcal{TT}$net and $s$ is the quantization step size which can be set to $s = 1/255$ to emulate 8-bit fixed-point values, or $2\epsilon$ for adversarial training with a $l_{\infty}$ disturbance limit of $\epsilon$. \cite{jia2020efficient}

\subsection{Last layer: linear classifier}
\label{sec:Lastlayer}

Depending on the application, the last layer can be floating point weights (scalability experiment), 1-bit weights or 4-bit weights. Next paragraphs describes 1) a possible natural accuracy optimisation and the transformation of the linear layer into CNF, especially for Formal verification.

\myparagraph{Last layer.}  The last linear layer is composed of a linear layer and a batch normalisation.  The weights of the linear layer can be natural instead of binary but this leads to a large increase in the size of the SAT formulas.

\myparagraph{Natural features.} We may also increase the amount of information held by the vector of features $V$ by accepting natural values. Coming back to our Example 1 for Section \ref{subsec:Construction_LTT}, we can see the output $y = \begin{bmatrix} 0,-5, 3, -2, 1,-4, 4, -1, 10,5, 13, 8, 11,6, 14, 9\end{bmatrix}$ as $ y = -5 \times \begin{bmatrix} 0,1, 0, 0, 0,0, 0, 0, 0,0, 0, 0, 0,0, 0, 0\end{bmatrix} + 3 \times \begin{bmatrix} 0,0, 1, 0, 0,0, 0, 0, 0,0, 0, 0, 0,0, 0, 0 \end{bmatrix} + \cdots + 9 \times \begin{bmatrix} 0,0, 0, 0, 0,0, 0, 0, 0,0, 0, 0, 0,0, 0, 1 \end{bmatrix} $. Therefore, each of these coefficients can have an associated SAT expression of its own.

For the sake of natural accuracy performance, as in \cite{evans2021making}, it is advisable to encode features and weights on 8 or 4 bits. However, the results in this paper are given for binary features and weights.

As we trained the $\mathcal{TT}$net with a final linear layer and then a batch normalisation, we encoded the last layer for the label $i$ as follows:

$$y_i = \sum_{k = 1}^{|V|} w_{k, i} V_k +b_i $$ and 

\begin{align}
y^{BN}_i  &=  \texttt{BatchNorm}(y_i) = \gamma \cdot \frac{y_i - \mathbb{E}(y_i)}{ \sqrt{Var[y_i] + \epsilon}} + \beta \nonumber 
          = \frac{ \sum_{k = 1}^{|V|} \gamma w_{k, i} V_k + \gamma b_i - \mathbb{E}(y_i)}{ \sqrt{Var[x] + \epsilon}} + \beta \nonumber
\end{align}

\noindent with $\epsilon = 1e-5$ and $\cdot$ denotes element-wise multiplication. We can rewrite the equality as:

\begin{align}
y^{BN}_i &= \sum_{k = 1}^{|V|} \tilde{w}_{k, i} V_k + \tilde{b_i} \nonumber \\
\tilde{w}_{k, i} &=  \lfloor \frac{ \gamma w_{k, i}}{ \sqrt{Var[y_i] + \epsilon}}  \rfloor \nonumber \\
\tilde{b_i}      &= \frac{ \gamma b_i - \mathbb{E}(y_i)}{ \sqrt{Var[y_i] + \epsilon}} + \beta \nonumber
\end{align}





To facilitate the CNF conversion, we also restrict the variance statistics and the
scale parameter $\gamma$ in $\texttt{BatchNorm}()$ of the last layer to be scalars computed on the whole feature map rather than per-channel statistics. Before rounding, we multiply $\tilde{w}_{k, i}$ and $ \tilde{b_i}$ by 100 in order to keep some details contained in the floating points. This inequality is encoded with the project \cite{imms-sat18}. The overall project is based on Sympy \cite{10.7717/peerj-cs.103}.

\subsection{Computation LTT with group $g=2$ in 1-dimension}

Computation of a LTT block 1D with group $g= 2$ is given in Table

\begin{figure*}[htb!]
    \centering
\begin{subfigure}[t]{0.45\textwidth}
\begin{tikzpicture}
\foreach \y in {0,...,3} 
    \foreach \x in {0,...,3}
        \draw[gray, thick] (0 + \x*0.5,0 - \y) rectangle (0.5 + \x*0.5,0.5 - \y);
    
\foreach \y in {0,...,5}
    \foreach \x in {1,...,2}
        \draw[gray, thick] (2.75 + \x*0.5,1 - \y) rectangle (3.25+ \x*0.5,1.5- \y);
\foreach \y in {0,1}
    \foreach \x in {0,...,2}
        \draw[gray,thick] (5+\x*0.5,0 - \y) rectangle (5.5+\x*0.5,0.5 - \y); 
\draw[babyblueeyes, thick] (0+ 0*0.5,0-1) rectangle (0.5+ 1*0.5,0.5);
\draw[babypink2, thick] (0+ 0*0.5,0-3) rectangle (0.5+ 1*0.5,0.5-2);

\foreach \y in {0,...,2}
    \draw[black, thick, fill=babyblueeyes] (2.75 + 0*0.5,1 - \y) rectangle (3.25+ 0*0.5,1.5- \y);
\foreach \y in {3,...,5}
    \draw[black, thick, fill=babypink2] (2.75 + 0*0.5,1 - \y) rectangle (3.25+ 0*0.5,1.5- \y);

\draw[black, thick, fill=babyblueeyes] (5+0*0.5,0 - 0) rectangle (5.5+0*0.5,0.5 - 0);
\draw[black, thick, fill=babypink2] (5+0*0.5,0 - 1) rectangle (5.5+0*0.5,0.5 - 1);

\foreach \y in {0,...,2}
    \draw[black, densely dotted] (0.5+ 1*0.5,0.5 - 0.75) -- (2.75 + 0*0.5,1 - \y + 0.25);

\foreach \y in {3,...,5}
    \draw[black, densely dotted] (0.5+ 1*0.5,0.5-2 - 0.75) -- (2.75 + 0*0.5,1 - \y + 0.25);
    
\foreach \y in {0,...,2}
    \draw[black, densely dotted] (3.25+ 0*0.5,1 - \y + 0.25) -- (5+0*0.5,0 - 0+0.25);

\foreach \y in {3,...,5}
    \draw[black, densely dotted] (3.25+ 0*0.5,1 - \y + 0.25) -- (5+0*0.5,0 - 1+0.25);

\end{tikzpicture}
    \caption{E-AE LTT internal working for 1 dimension: first 1D-CNN layer of kernel size $2$, stride $1$ and group $g = 2$. The amplification layer (second Conv1D layer) has an amplification parameter, $\tau = 3$ of kernel size $1$ and stride $1$.The intermediate values are real and the input/output values are binary.}
    \label{fig: Amplification Layer}
\end{subfigure}
\hspace{0.8cm}
\begin{subfigure}[t]{0.45\textwidth}
\begin{tikzpicture}
\foreach \x in {0,...,5} {
    \draw[rounded corners] (0+ \x, 0) rectangle (0.5+\x, 4);
    \draw[->] (0.5+ \x -1, 2) -- (1+\x-1, 2); }
\path (0, 0) -- node[rotate=90,anchor=center] {Conv1D}(0.5, 4);
\path (0+1, 0) -- node[rotate=90,anchor=center] {Batch Normalization 1D}(0.5 + 1, 4);
\path (0+2, 0) -- node[rotate=90,anchor=center] {SeLU}(0.5+2, 4);
\path (0+3, 0) -- node[rotate=90,anchor=center] {Conv1D (kernel size = 1)}(0.5+3, 4);
\path (0+4, 0) -- node[rotate=90,anchor=center] {Batch Normalization 1D}(0.5+4, 4);
\path (0+5, 0) -- node[rotate=90,anchor=center] {$bin_{act}$}(0.5+5, 4);
\end{tikzpicture}
    \caption{LTT overview of a Expanding AutoEncoder LTT in 1 dimension: the Conv1D with kernel size = 1 is the amplification layer. The intermediate values are real and the input/output values are binary.}
    \label{fig:LTT}
\end{subfigure}
\caption{Learning Truth Table Block (LTT)}
\label{fig: inner and outer Learning Truth Table Block}
\end{figure*}
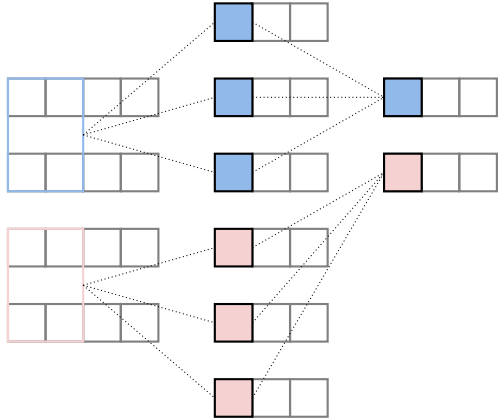
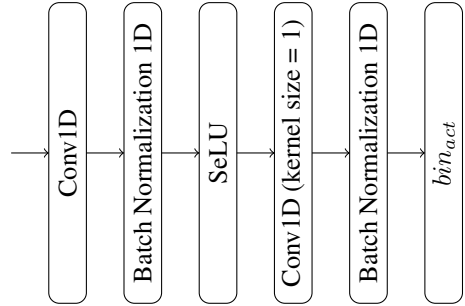

\subsection{Code comparison}

Figure~\ref{fig:cnn_numpy} gives a code comparison between a Pytorch~\cite{paszke2019pytorch} computation of CNN form of $\mathcal{TT}$net and its equivalent Numpy~\cite{van2011numpy} truth table form. Both gives the same result if the truth tale is created from the CNN.

\begin{figure*}
\centering
\begin{tikzpicture}
\node[draw, minimum width=0.45\textwidth,minimum height=2.5cm] (cnn) {
    \begin{tabular}{l}
    \texttt{\# PyTorch CNN computation}\\
    \texttt{import torch}\\
    \texttt{import torch.nn as nn}\\
    \texttt{conv1 = nn.Conv2d(1, 32, 3)}\\
    \texttt{image\_bin \# Shape (1, 1, 28, 28))}\\
    \texttt{output = conv1(image\_bin)}\\
    \texttt{output\_bin = act\_bin(output)}\\
    \end{tabular};
};
\node[draw, minimum width=0.4\textwidth,minimum height=2.5cm, right=of cnn] (numpy) {
    {\begin{tabular}{l}

\texttt{ def BitsToInt(bits):}\\
\texttt{           m,n = bits.shape}\\
\texttt{     a = 2**np.arange(n)[::-1]}\\
\texttt{return bits @ a}\\
\texttt{}\\
    
    \texttt{ \# Numpy LTT Layer computation}\\
    \texttt{import torch.nn.Unfold as U}\\
    \texttt{import numpy as np}\\
    \texttt{u = U(kernel\_size=3)}\\
    \texttt{image\_bin \# Shape (1, 1, 28, 28))}\\
    \texttt{TT\_l = np.array(32, 2**9)}\\
    \texttt{image\_unfold = u(image\_bin).reshape(9,-1)}\\
    \texttt{H0 = np.sqrt(image\_unfold.shape[-1])}\\
    \texttt{image\_unfoldint = BitsToInt(image\_unfold)}\\
    \texttt{res = TT\_l[:, image\_unfoldint]}\\
    \texttt{output = res.reshape(1,32,H0, H0)}
    \end{tabular}};
};
\end{tikzpicture}
\caption{Computation of DCNN standard and LTT layer with numpy.}
\label{fig:cnn_numpy}
\end{figure*}

\subsection{Automatic training \& post-training optimizations} 
\label{susubsec:Optimizations}

\subsubsection{\textit{Don't Care Terms} ($DCTs$) injection via Dual Step (DS) activation function}

We transform the classic step function by adding a noise zone in the middle as follows: Given a threshold $T \in \mathbb{R}$, we define the noisy binary activation function $h : \mathbb{R} \rightarrow \mathbb{R}$ for an input $x$ as:

\begin{align*}
DS_{act, T}(x) =
\begin{dcases*}
1  & if  x $\in \left[\frac{T}{2}, \infty\right)$, \\[1ex]
r  & if  x $\in \left[-\frac{T}{2}, \frac{T}{2}\right]$, \\[1ex]
0  & if  x $\in \left(-\infty, -\frac{T}{2}\right]$ \\[1ex]
\end{dcases*}
\end{align*}

where $r \in \{0, 1\}$ is a random binary integer that follows a Bernoulli distribution with probability $1/2$.

The goal of the activation function is to propose a way to integrate the "don't care" term directly in the learning phase. By doing so, we also introduce the "don't care" for the LTT block: in fact, instead of assigning the value 0 or 1 in the truth table, if the output falls in the "noisy part" - then it means for that input we don't care. It leads to a drastic simplification of the CNF/DNF equation.

\subsubsection{LTT correlation:  reducing the number of LTT}

\vspace*{-0.2cm}
To reduce the number of LTT obtained with  $\mathcal{TT}$net, we introduce a new metric called Truth Table Correlation ($TTC$). This metric is based on the idea that two different LTT blocks may learn similar Truth Tables as they are all completely decoupled from each other. Thus, to prevent filter/truth table redundancy, we introduce a correlation measure between two LTT block rules. It is defined as follows: 
\begin{equation*}
TTC(y_1, y_2) = \left\{
    \begin{array}{ll}
        \frac{HW(y_1, \overline{y_2})}{|y_1|} - 1 & \mbox{if } abs(\frac{HW(y_1, \overline{y_2})}{|y_1|} - 1) > \frac{HW(y_1, \overline{y_2})}{|y_1|} \\
        \frac{HW(y_1, y_2)}{|y_1|} & \mbox{otherwise.}
    \end{array}
\right.
\end{equation*}
where $y_1$ and $y_2$ are the outputs of the LTT blocks, $\overline{y_2}$ is the negation of $y_2$,  $|y_1|$ represents the number of elements in $y_1$, $HW$ is the Hamming distance function (the Hamming distance between two equal-length strings of symbols is the number of positions at which the corresponding symbols are not equal). $TTC$ metric varies from -1 to 1. For $TTC=-1$, the LTT blocks are exactly opposite while they are the same if $TTC=1$. We systematically filter redundant rules above the threshold correlation of $ \pm 0.9$. If the correlation is positive, we delete one of the two filters and give the same value to the second filter. If the correlation is negative, we delete one of the two filters and give the opposite value to the second filter.

\subsubsection{Polynomial activation function} 

We propose to use the order 2 polynomial approximation of the ReLu function as proposed in~\cite{gottemukkula2020polynomial}, to raise the accuracy without losing the property of interpretability, Boolean circuit conversion, and formal verification. for that, we simply use two binary layers for classification, separated by the order 2 polynomial activation function. As the feature and the weights are binary, we can transform this non-linear block into 1 linear block and increase the size of the feature size $F$ by order $F^2$ using AND gates. 

\subsubsection{Scaling global $\mathcal{TT}$net architecture}

\begin{figure*}[htb!]
 \center
    \includegraphics[width=0.7\textwidth]{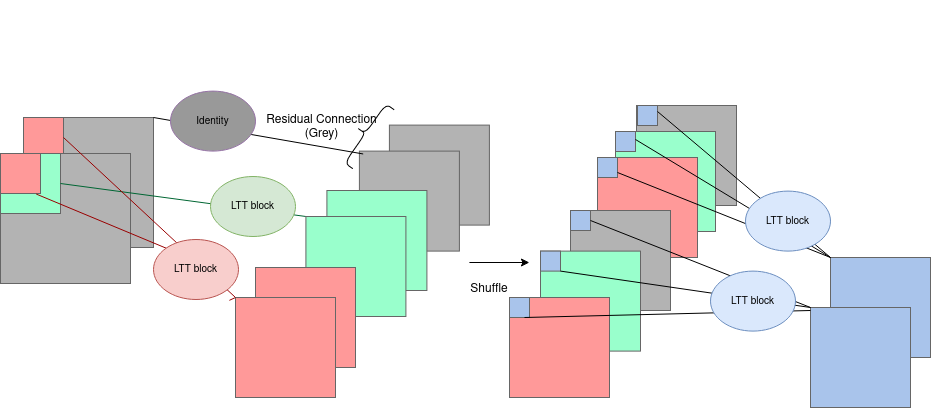}
  \caption{Multi-head CNN layers, with green layers as output from shallow-big LTT block, red as output from the deep-small LTT block and grey from the residual connection. }
  \label{fig: Multi-head CNN layers}
\end{figure*}

\myparagraph{Multi-head CNN layers.} First, we add multi-head CNN layers into $\mathcal{TT}$net. Figure \ref{fig: Multi-head CNN layers} shows the structure of the multi-head CNN layers used. With the aim to increase the learning capacity of $\mathcal{TT}$net while still maintaining a small look-up table, it makes perfect sense that we want some LTT blocks to see deeply between channels but have small kernel size (called deep-small LTT blocks), but also want some LTT blocks that have larger kernel size but shallow between channels (called shallow-big LTT block). 

\myparagraph{Residual connection.} We also stabilise the training by concatenating the input. In the case of stride $2$, we apply an average pooling and reuse the Dual Step activation function.

\myparagraph{Shuffle operation and second stage LTT block.} Finally, we shuffle the channels in order to set side by side the channels from the deep-small LTT block, followed by the shallow-small LTT block, and then the residual connection. Then we add a second LTT block that takes the three channels as input and outputs a value.

\subsubsection{Permutation strategy: improvement of the accuracy for tabular dataset} The $\mathcal{TT}$net decoupled nature is at the same time its strength and its weakness. It means that one feature of $V$, the final vector before the final binary linear regression, can only see a subpart of all the inputs - whereas in MLP one feature of $V$ can see all the inputs. To overcome this natural drawback of $\mathcal{TT}$net, we present a straightforward and efficient strategy that randomly tries multiple pairs of features. More precisely, the updated dataset $\mathcal{D}_f$ is composed of the original dataset $\mathcal{D}$, concatenated will $P$ permutated versions of its features: $\mathcal{D}_f = \mathcal{D} || Perm_1(\mathcal{D}) || \ldots || Perm_P(\mathcal{D})$, where $||$ denotes the concatenation operation and $Perm_i$ are randomly chosen permutations on $F$ elements.

\subsubsection{ROBDD rule format for tabular dataset only} 
\label{subsubsec:ROBDD}

At this step, a rule is in DNF format. However, we prefer to have the rules in a decision tree format. Therefore, we transform our DNF into its equivalent ROBDD graph. A Binary Decision Diagram (BDD) is a directed acyclic graph used to represent a Boolean function. They were originally introduced by~\cite{lee1959representation, akers1978binary}. In~\cite{bryant1986graph}, Randal Bryant introduced the Reduced Ordered BDD (ROBDD), which is a canonical form given an identical ordering of input variables, equivalent Boolean functions will always reduce to the same ROBDD. This is a desirable property for determining formal equivalence and very useful for visualizing rules. We perform the transformation with PyEda library~\cite{aqajari2021pyeda}. Graphs, given in Figure~\ref{fig:casestudy}, are small and easy to understand.

\subsection{Ablation study: amplification layer study for natural accuracy}
\label{subsection:App_Ablation}

From a practical standpoint, using eight filters in layer one and one filter in layer two drastically improves the learning capacity of the $\mathcal{TT}$net and therefore the natural accuracy as well. This observation is consistent with other published studies \cite{sandler2018mobilenetv2}. We computed the natural accuracy for two models (model A(6,6) and model B(7,7)), for three types of amplifications (no amplification, $\times$4, $\times$8) and high noise training (see Table~\ref{table:models} in Appendix~\ref{sec:Aann_Modeldescrption} for more details).  Results are presented in Table~\ref{table:amplification1}. We confirm that in all cases, adding an amplification layer significantly improves the resulting natural accuracy.

\begin{table*}[htb!]
\centering
\captionof{table}{\label{table:amplification1} Effect of amplification layer on $\mathcal{TT}$net natural accuracy. Natural accuracy was examined for three different types of amplification configuration, for CIFAR-10 and MNIST, in the case of training with high noise.}
\resizebox{0.75\columnwidth}{!}{
\begin{tabular}{@{}lcccc@{}}
\toprule
                  &           &  & \multicolumn{2}{c}{\textbf{Natural accuracy}} \\
          \textbf{Dataset}               &      \textbf{Noise Training}                   &              \textbf{Amplification} & Model A (6,6)    & Model B (7,7)   \\ \midrule
\multirow{3}{*}{MNIST}   
                         & \multirow{3}{*}{0.4}    & None            & 82.63\%        & 92.42\%     \\
                         &                         & $\times$4         & 85.67\%        & 94.19\%     \\
                         &                         & $\times$8        & 87.59\%        & 95.77\%     \\ \midrule
                         \midrule
\multirow{3}{*}{CIFAR-10} &  \multirow{3}{*}{16/255} & None            & 22.40\%        &   28.05\%    \\
                         &                         & $\times$4         & 27.86\%        &   37.29\%  \\
                         &                         & $\times$8        & 32.19\%        &   47.12\%   \\ \bottomrule
\end{tabular}}
\end{table*}




\section{ImageNet scalability- complementary details}
\label{sec:Ann_natural}

\myparagraphtitle{Training details.} 

We employed the DoReFa-Net method~\cite{zhou2016dorefa} for CIFAR-10, a technique for training convolutional neural networks with low-bitwidth activations. Finally, to overcome the limitations of the $\mathcal{TT}$net grouping method, we extended the training to 500 epochs, resulting in a more accurate model. We train the networks using the SGD optimizer with a minibatch size of 128. Learning rate is 0.001, the loss is cross entropy. We use SGD momentum is 0.9 and the weight decay is 5e-4. Finally the learning rate scheduler cosine annealing \cite{loshchilov2016sgdr} with a maximum number of iterations 200. We use random crop function and horizontal flip as data augmentation techniques for training. 

\myparagraph{Architecture details.} There are 3 layers of multihead-layer with $n=16$, the last 2 with stride 2, there are $64 \times l$ filters with $l$ the layer number.

\section{Global exact interpretability - complementary details}
\label{appendix:Global}       
\subsection{Model architecture and training conditions - general}
\label{appendix:reproductibility}

\myparagraphtitle{Training method.}  We built our training method on top of \cite{jia2020efficient} project and we refer to their notations for this Section. We trained the networks using the Adam optimizer \cite{kingma2014adam} for 10 epochs with a minibatch size of 128. The mean and variance statistics of batch normalization layers are recomputed on the whole training set after training finishes.

\myparagraph{Weights initialization.} 
Weights for the final connected layers are initialized from a Gaussian distribution with standard deviation 0.01 and the mask weights in BinMask are enforced to be positive by taking the absolute value during initialization.

\myparagraph{Other hyperparameters.} 
\label{sec:Aann_Modeldescrption4}
 We apply a weight decay of $1e-7$
on the binarized mask weight of BinMask.

\myparagraph{Architecture.} Each dataset has the same input layers and output layer, with of course the size adapted to the dataset for the output layer. The first layer is a Batch Normalization layer with an $\epsilon$ of $10^{-5}$ and a momemtum of $0.1$. It is followed by the LTT block, which will be detailled for each dataset. Finally, the binary linear regression is computed, with input and output features sizes detailled below.

\subsection{Model architecture and training conditions - dataset specific}
\label{appendix:reproductibility2}

\myparagraphtitle{Adult.} We trained the Adult dataset in 10 epochs. The first layer is the Batch Normalization layer as described above. It is followed by a LTT block. The first convolution has 10 filters. The convolution is done with a stride of 5, a kernel size of 5 and no padding. It is followed by a Batch Normalization layer with the same parameters as the input one. The following convolution has 10 filters and a kernel size and a stride of 1. A last Batch Normalization finishes with the same parameter as above the LTT Block. The final linear regression takes 200 features as input and 2 as outputs. The learning rate is 0.005.

\myparagraph{Breast-Cancer.} We trained the Breat-Cancer dataset in 10 epochs. The first layer is the Batch Normalization layer as described above. It is followed by a LTT block. The first convolution has 10 filters. The convolution is done with a stride of 4, a kernel size of 3 and no padding. It is followed by a Batch Normalization layer with the same parameters as the input one. The following convolution has 10 filters and a kernel size and a stride of 1. A last Batch Normalization finishes with the same parameter as above the LTT Block. The learning rate is 0.005.

\myparagraph{Compas.} We trained the Compas dataset in 60 epochs. The first layer is the Batch Normalization layer as described above. It is followed by a  LTT block. The first convolution has 20 filters (amplification of 20). The convolution is done with a stride of 1, a kernel size of 6 and no padding. It is followed by a Batch Normalization layer with the same parameters as the input one. The following convolution has 5 filters and a kernel size and a stride of 1. A last Batch Normalization finishes with the same parameter as above the LTT Block. The final linear regression takes 60 features as input and 2 as outputs. The learning rate is 0.0005. Only the best model in terms of testing accuracy was kept.

\myparagraph{Cancer.} We trained the Cancer dataset in 10 epochs. The first layer is the Batch Normalization layer as described above. It is followed by a LTT block. The first convolution has 10 filters (amplification of 10). The convolution is done with a stride of 1, a kernel size of 6 and no padding. It is followed by a Batch Normalization layer with the same parameters as the input one. The following convolution has 10 filters and a kernel size and a stride of 1. A last Batch Normalization finishes with the same parameter as above the LTT Block. The final linear regression takes 80 features as input and 2 as outputs. The learning rate is 0.005.

\myparagraph{Diabetes.} We trained the Diabetes dataset in 10 epochs. The first layer is the Batch Normalization layer as described above. It is followed by a LTT block. The first convolution has 10 filters (amplification of 10). The convolution is done with a stride of 1, a kernel size of 6 and no padding. It is followed by a Batch Normalization layer with the same parameters as the input one. The following convolution has 10 filters and a kernel size and a stride of 1. A last Batch Normalization finishes with the same parameter as above the LTT Block. The final linear regression takes 295 features as input and 3 as outputs. The learning rate is 0.0005.

\subsection{Dataset details}
\label{appendix:datasets}

All datasets have been split 5 times in a 80-20 train-test split for k-fold testing.

\myparagraph{Adult.} The Adult dataset contains 48,842 individuals with 18 binary features and a label indicating whether the income is greater than 50K\$ USD or not.

\myparagraph{Breast-Cancer.} More details in \cite{zwitter1988uci}.

\myparagraph{Compas.} The Compas dataset consists of 6,172 individuals with 10 binary features and a label that takes a value of 1 if the individual does not re-offend and 0 otherwise.

\myparagraph{Cancer.} Cancer dataset contains 569 data points with 30 numerical features. The goal is to predict if a tumor is cancerous or not. We encoded each integer value into a one-hot vector, resulting in a total of 81 binary features.

\myparagraph{Diabetes.} Regarding the Diabetes dataset, it contains 100000 data points of patients with 50 features, both categorical and numerical. We kept 43 features, 5 numerical and the other are categorical which resulted in 291 binary features and 5 numerical features. We will predict one of the three labels for hospital readmission.

\subsection{Complementary results ablation study}
\label{appendix:complement_ablation}

We present additional results for the ablation study in Table~\ref{table:res_Ablation}.

\myparagraph{Permutation strategy influence.} Except for Compas, the algorithm is valuable for $P \geq 5$. Indeed, the accuracy increases by $ 1.2\%, 0.7\%, 0.4\%$ for the datasets Adult, Cancer, and Diabetes respectively when using the permutation strategy. On the other hand, it drastically amplifies the complexity, with a 10 to 40-fold increase, depending on the dataset. We also note that the permutation strategy has barely any impact on the average number of conditions per rule. This is expected as this metric is heavily depending on the kernel size of $\mathcal{TT}$net. 

\myparagraph{Rules optimizations influence.} Complexity decreases drastically from $\mathcal{R}$ to $\mathcal{R}_{opt}$: between $\times 1.4$ and $\times 2.8$ for Adult, between $\times 1.9$ and $\times 2.3$ for Compas, between $\times 1.8$ and $\times 5.3$ for Cancer and between $\times 1.4$ and $\times 5.4$ for Diabetes. On the other hand, except for Compas, the optimizations lead to a slight decrease in measured accuracy, the largest observed decreases being $-0.5\%$, $-0.7\%$, and $-0.9\%$ for the datasets Adult, Cancer, and Diabetes respectively. In addition, we can observe that the size of each rule decreases: this is mainly due to the $DCTs$ injection. Quine-McCluskey algorithm~\cite{quine1952problem} outputs smaller rules with $DCTs$ usage. Finally, in the case $P=10$, many rules are filtered, even with the high threshold $\pm 0.9$.

\begin{table}[!htb]
\centering
\captionof{table}{\label{table:res_Ablation} Ablation study results for $\mathcal{TT}$net on the four datasets. All results are computed for 5 models with 5 different k-fold for learning. No permutation column refers to normal dataset. Complexity is defined as the total number of conditions for all rules.}
\resizebox{1\columnwidth}{!}{
\begin{tabular}{ll|cc|cc|cc}
\hline
\multirow{2}{*}{\textbf{Dataset}} & \textbf{Permutations }                & \multicolumn{2}{c|}{\textbf{No Permutation}}                   & \multicolumn{2}{c|}{\textbf{$P=5$}}                 & \multicolumn{2}{c}{\textbf{$P=10$}}                \\ 
 &  \textbf{Rule Set used}                  & $\mathcal{R}$               & $\mathcal{R}_{opt}$             & $\mathcal{R}$               & $\mathcal{R}_{opt}$                         & $\mathcal{R}$               & $\mathcal{R}_{opt}$             \\ \hline 

  \multirow{5}{*}{Adult} & Accuracy (\%)               & 83.4 $\pm$ 0.6 & 83.2 $\pm$ 0.5 & 84.6 $\pm$ 0.3 & 84.2 $\pm$ 0.3 & 84.4 $\pm$ 0.5 & 83.9 $\pm$ 0.3 \\
& F1-score (\%)               & 61.1 $\pm$ 1.3 & 61.1 $\pm$ 1.4 & 65.9 $\pm$ 1.4 & 66.1 $\pm$ 1.6 & 66.9 $\pm$ 0.6 & 64.0 $\pm$ 5.7 \\
 & Number of rules               & 7.0 $\pm$ 2.2 & 6.0 $\pm$ 2.2 & 137.0 $\pm$ 6.8 & 130.0 $\pm$ 9.8 & 310.0 $\pm$ 27.8 & 260.0 $\pm$ 23.0 \\
& Complexity              & 47.0 $\pm$ 15.6 & 24.0 $\pm$ 8.5 & 909.0 $\pm$ 212.2 & 673.0 $\pm$ 144.5 & 2156.0 $\pm$ 374.5 & 1404.0 $\pm$ 181.7 \\
 & Avgerage conditions/rule               & 5.6 $\pm$ 1.2 & 3.3 $\pm$ 1.5 & 6.7 $\pm$ 1.4 & 5.1 $\pm$ 0.9 & 7.8 $\pm$ 0.9 & 5.8 $\pm$ 0.4 \\
 & Statistical Parity               & 99.0 $\pm$ 0.6 & 82.4 $\pm$ 1.1 & 85.1 $\pm$ 7.1 & 96.3 $\pm$ 2.9 & 93.8 $\pm$ 2.5 & nan $\pm$ nan \\
 & Odds Equalized               & 0.0 $\pm$ 0.0 & 0.0 $\pm$ 0.0 & -0.1 $\pm$ 0.0 & -0.1 $\pm$ 0.1 & -0.1 $\pm$ 0.0 & -0.1 $\pm$ 0.1 \\ 
 & Extraction time (s.)              & 7 $\pm$ 2 & 8 $\pm$ 1 & 9 $\pm$ 1 & 10 $\pm$ 2 & 17 $\pm$ 2 & 17 $\pm$ 4 \\
 & Train Time (s.)                             & \multicolumn{2}{c|}{53 $\pm$ 2}                & \multicolumn{2}{c|}{56 $\pm$ 3}                & \multicolumn{2}{c}{58 $\pm$ 1}                 \\\hline
 
 \hline

  \multirow{5}{*}{Compas} & Accuracy (\%)               & 66.4 $\pm$ 1.3 & 66.4 $\pm$ 1.3 & 65.5 $\pm$ 1.2 & 65.5 $\pm$ 1.2 & 66.1 $\pm$ 1.6 & 66.1 $\pm$ 1.6 \\
& F1-score (\%)               & 72.2 $\pm$ 1.7 & 72.2 $\pm$ 1.7 & 71.3 $\pm$ 0.7 & 71.3 $\pm$ 0.7 & 70.8 $\pm$ 2.1 & 70.8 $\pm$ 2.1 \\
 & Number of rules               & 13.0 $\pm$ 1.8 & 13.0 $\pm$ 1.8 & 219.0 $\pm$ 9.9 & 219.0 $\pm$ 9.9 & 511.0 $\pm$ 9.5 & 511.0 $\pm$ 9.5 \\
& Complexity              & 343.0 $\pm$ 41.0 & 155.0 $\pm$ 21.8 & 6118.0 $\pm$ 812.3 & 2670.0 $\pm$ 468.4 & 15541.0 $\pm$ 1276.7 & 7974.0 $\pm$ 918.0 \\
 & Avgerage conditions/rule               & 27.1 $\pm$ 3.2 & 10.3 $\pm$ 1.3 & 27.9 $\pm$ 2.4 & 12.2 $\pm$ 1.6 & 30.4 $\pm$ 2.4 & 15.6 $\pm$ 1.8 \\
 & Statistical Parity               & 81.4 $\pm$ 3.0 & 81.4 $\pm$ 3.0 & 100.0 $\pm$ 0.0 & 100.0 $\pm$ 0.0 & 100.0 $\pm$ 0.0 & 100.0 $\pm$ 0.0 \\
 & Odds Equalized               & 0.0 $\pm$ 0.0 & 0.0 $\pm$ 0.0 & 0.0 $\pm$ 0.1 & 0.0 $\pm$ 0.1 & 0.1 $\pm$ 0.1 & 0.1 $\pm$ 0.1 \\
  & Extraction time (s.)              & 7 $\pm$ 1 & 7 $\pm$ 1 & 12 $\pm$ 1 & 12 $\pm$ 1 & 22 $\pm$ 1 & 21 $\pm$ 1 \\
 & Train Time                              & \multicolumn{2}{c|}{117 $\pm$ 7}                & \multicolumn{2}{c|}{118 $\pm$ 3}                & \multicolumn{2}{c}{117 $\pm$ 8}                 \\\hline

  \multirow{5}{*}{Cancer} & Accuracy (\%)               & 96.4 $\pm$ 1.0 & 95.7 $\pm$ 0.9 & 95.7 $\pm$ 0.9 & 95.7 $\pm$ 0.7 & 97.1 $\pm$ 0.6 & 97.1 $\pm$ 0.7 \\
 & F1-score (\%)               & 94.9 $\pm$ 1.6 & 94.3 $\pm$ 1.6 & 94.6 $\pm$ 1.4 & 93.8 $\pm$ 1.0 & 96.0 $\pm$ 1.0 & 96.0 $\pm$ 1.2 \\
 & Number of rules               & 60.0 $\pm$ 3.6 & 49.0 $\pm$ 6.8 & 288.0 $\pm$ 8.7 & 231.0 $\pm$ 21.5 & 615.0 $\pm$ 22.0 & 371.0 $\pm$ 61.8 \\
& Complexity              & 1098.0 $\pm$ 239.5 & 205.0 $\pm$ 33.7 & 5222.0 $\pm$ 167.4 & 2924.0 $\pm$ 328.5 & 10467.0 $\pm$ 1902.1 & 4665.0 $\pm$ 1027.4 \\
 & Avgerage conditions/rule               & 19.4 $\pm$ 2.8 & 4.2 $\pm$ 0.3 & 17.6 $\pm$ 0.6 & 13.3 $\pm$ 0.8 & 17.2 $\pm$ 2.5 & 12.1 $\pm$ 1.1 
\\ 
 & Extraction time (s.)             & 9 $\pm$ 1 & 9 $\pm$ 1 & 18 $\pm$ 1 & 18 $\pm$ 1 & 31 $\pm$ 5 & 30 $\pm$ 5 \\
 & Train Time (s.)                             & \multicolumn{2}{c|}{19 $\pm$ 1}                & \multicolumn{2}{c|}{20 $\pm$ 2}                & \multicolumn{2}{c}{20 $\pm$ 2}                 \\\hline

  \multirow{5}{*}{Diabetes} & Accuracy (\%)               & 56.7 $\pm$ 0.2 & 56.7 $\pm$ 0.2 & 57.0 $\pm$ 0.5 & 57.0 $\pm$ 0.5 & 57.4 $\pm$ 0.8 & 56.5 $\pm$ 0.9 \\
 & F1-score (\%)               & 56.7 $\pm$ 0.2 & 56.7 $\pm$ 0.2 & 57.0 $\pm$ 0.5 & 57.0 $\pm$ 0.5 & 57.4 $\pm$ 0.8 & 56.5 $\pm$ 0.9 \\
 & Number of rules               & 35.0 $\pm$ 5.9 & 33.0 $\pm$ 5.2 & 406.0 $\pm$ 15.7 & 406.0 $\pm$ 15.7 & 1059.0 $\pm$ 45.5 & 1043.0 $\pm$ 46.2 \\
& Complexity              & 568.0 $\pm$ 131.9 & 106.0 $\pm$ 27.6 & 7834.0 $\pm$ 1143.0 & 5561.0 $\pm$ 761.1 & 21644.0 $\pm$ 2802.2 & 15475.0 $\pm$ 2224.5 \\
 & Avgerage conditions/rule               & 16.2 $\pm$ 1.3 & 3.6 $\pm$ 0.4 & 19.4 $\pm$ 2.5 & 13.8 $\pm$ 1.7 & 20.8 $\pm$ 2.8 & 14.3 $\pm$ 1.7 \\
 & Extraction time (s.)            & 19 $\pm$ 3 & 18 $\pm$ 2 & 35 $\pm$ 6 & 34 $\pm$ 5 & 68 $\pm$ 4 & 67 $\pm$ 5 \\
 & Train Time (s.)                              & \multicolumn{2}{c|}{110 $\pm$ 4}                & \multicolumn{2}{c|}{126 $\pm$ 8}                & \multicolumn{2}{c}{149 $\pm$ 4}                 \\\hline

\end{tabular}}
\end{table}


\subsection{Results tabular dataset} 
\label{subsec_appedix:Tabular}

Table~\ref{tab:tabular_appendix} shows the results of various models on three tabular datasets: Adult breast Cancer. The comparisons include linear models such as Decision Tree and Logistic Regression, and non-linear models such as Diff Logic Net, RRL, Decision Rule Sets and the proposed $\mathcal{TT}$net model. The evaluation criteria include accuracy and the number of gates in the model. The results show that $\mathcal{TT}$net has comparable accuracy to other models while having a smaller number of gates. We also want to hightligths that RRL need more than a hour to train on adult wheras we need few seconds. Finally, any of these models can scale to Cifar-10 with 70\% accuracy as us. We also highlight that BNNs is not interpretable by design and therefore can not be considedred as an exact and global interpretable model. We deicde too put LogicNet ithat comparison, despitdes the fact that not interpration study was made on the model, as we think as the model is suitable for interpreatblility even if it was not made.

\begin{table*}[htb!]
\centering
\caption{\label{tab:tabular_appendix}Comparison of accuracy (acc.) and number of gates (\# Gates) for different models on tabular datasets. Our results is given with and without Human Knowledge (H.K.) injection. }
\resizebox{\columnwidth}{!}{
\begin{tabular}{@{}l|cccc|ccccccccclcccc@{}}
\toprule
              & \multicolumn{4}{c|}{Linear Models}                                                                                                                                & \multicolumn{14}{c}{Non Linear Model}                                                                                                                                                                                                                                                                                                                                                                                                                                                                                                                   \\ \midrule
              & \multicolumn{2}{c}{\begin{tabular}[c]{@{}c@{}}Decision \\ Tree\end{tabular}} & \multicolumn{2}{c|}{\begin{tabular}[c]{@{}c@{}}Logistic\\ Regression\end{tabular}} & \multicolumn{2}{c}{\begin{tabular}[c]{@{}c@{}}Diff Logic Net\\  ~\cite{petersen2022deep} \end{tabular}} & \multicolumn{2}{c}{RRL \cite{wang2021scalable}} & \multicolumn{2}{c}{\begin{tabular}[c]{@{}c@{}}Decision \\ Rule Sets \cite{yang2021learning}\end{tabular}} & \multicolumn{2}{c}{\begin{tabular}[c]{@{}c@{}}Ours (Big)\\ with H.K.\end{tabular}} & \multicolumn{2}{c}{\begin{tabular}[c]{@{}c@{}}Ours (Big)\\ without H.K.\end{tabular}} & \multicolumn{2}{c}{\begin{tabular}[c]{@{}c@{}}Ours (small)\\ with H.K.\end{tabular}} & \multicolumn{2}{c}{\begin{tabular}[c]{@{}c@{}}Ours (smal)\\ without H.K.\end{tabular}} \\ \midrule
              & Acc.                                 & \#Param                               & Acc.                                    & \#Param                                  & Acc.                                 & \#Gates                               & Acc.       & \#Gates    & Acc.                                    & \#Gates                                 & Acc.                                    & \#Gates                                  & Acc.                            & \multicolumn{1}{c}{\#Gates}                         & Acc.                                      & \#Gates                                  & Acc.                                       & \#Gates                                   \\ \midrule
Adult         & 79.5\%                               & 50                                    & 84.8\%                                  & 234                                      & 84.8\%                               & 1280                                  & 85.7\%     & 5 10**6    & 84.4\%                                  & 83                                      & 85.3\%                                  & 1475                                     & 85.3\%                          & 2156                                                & 84.8\%                                    & 471                                      & 84.8\%                                     & 563                                       \\
Breast Cancer & 71.9\%                               & 100                                   & 72.9\%                                  & 104                                      & 76.1\%                               & 640                                   & -          & -          & -                                       & -                                       & -                                       & -                                        & -                               & -                                                   & 77.6 \%                                   & 71                                       & 77.6 \%                                    & 123                                       \\
Compas        & -                                    & -                                     & -                                       & -                                        & -                                    & -                                     & -          & -          & 66.5\%                                  & 42                                      & -                                       & -                                        & -                               & -                                                   & 66.4\%                                    & 155                                      & 66.4\%                                     & 343                                       \\ \bottomrule
\end{tabular}}
\end{table*}

\subsection{Complementary results on adult dataset use-case}
\label{app:Complementary_usecase}

\myparagraphtitle{Changing existing rules \& adding a new rule.} We first change the existing rules by, on the one hand, optimizing numerical values of existing conditions and on the other hand adding complementary conditions. We optimized on the training dataset the numerical values in the ``Years of Education'' condition, and we fixed it to $12$. 
Then, we added the conditions ``Is a School Prof'' in Rule 2. In fact, it seems to be an important feature as we find it in almost all other configurations. We also added ``Born In Peru'' in Rule 3 as we simply add the countries of South and Latin America (other such countries did not change performances so we did not include them). The accuracy went up to $83.66\%$ and $83.61\%$ respectively and the statistical parity to $0.823$ and $0.824$ respectively. 
These changes are done on the train, altogether leading to test the accuracy of $84.0\% \: (+0.4\%)$  and a statistical parity of $0.826 \: (+0.02)$. Surprisingly, age and job type do not influence the prediction. We also noticed that the ``Statistical Weight'' feature is used in most of the models. Therefore, we decided to add the following rule displayed in Figure~\ref{fig:casestudy} :
 $(\text{Age}<28) \lor \text{Farmer} \lor \text{HandlerCleaner} \lor \text{Machinist} \lor ((\text{Age}>60) \land (\text{StatisticalWeight}<300k)) $. The accuracy went further up to $84.6\% (+1\%)$, and the statistical parity to $0.838 + (+0.01)$ . This illustrates the possibility to add human knowledge to a CNN-based model.

\begin{figure}[htb!]
\caption[Adult dataset ]{An other example of a learned model on the Adult dataset. \label{fig:casestudy_appendix1} With this model, we reach an accuracy of 82.28\% but with much more conditions than the initial one presented in Figure~\ref{fig:casestudy} which reached an accuracy of 83.6\%. They have 29 conditions and 8 conditions respectively.}
\centering
\includegraphics[scale=0.40]{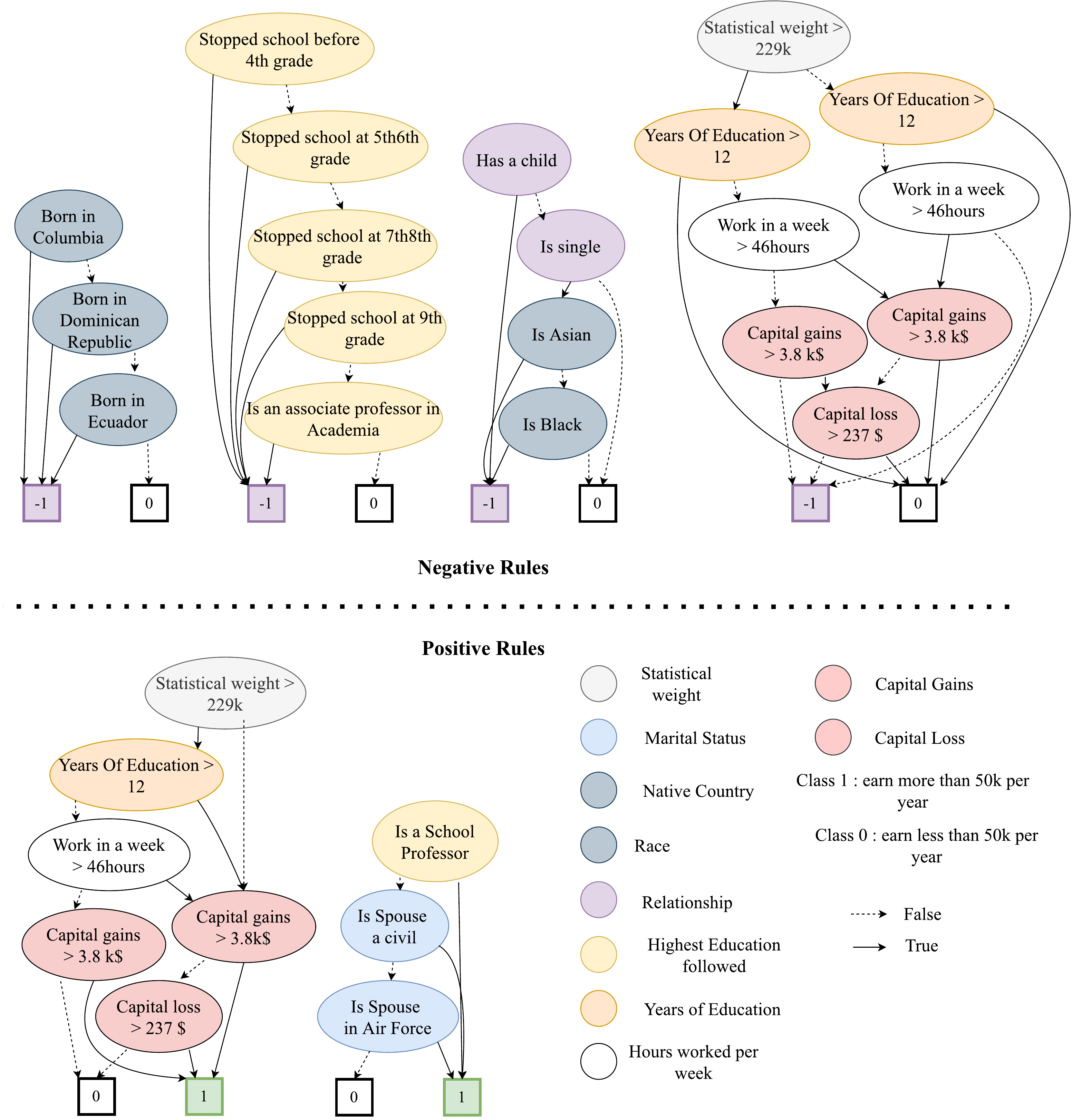}

\end{figure}

\myparagraph{Other model for Adult.} Regarding this specific model in Figure~\ref{fig:casestudy}, the rules are fully interpretable. Yet, depending on the features learnt by $\mathcal{TT}$net, it would be sometimes not fully interpretable: indeed, the feature Statistical Weight is a feature regarding the sampling weight of each group in the Adult census, see Figure~\ref{fig:casestudy_appendix1}. This feature does not give a clear indication about an individual. Moreover, some clear unfair rules regarding the Race attribute were deducted by the algorithm such as the Born in Columbia/Dominican Republic/Ecuador features. These features are not directly the Race attributes, but are correlated to the Race of the individual. Here, it would translate to a ghost rule describing a Latin American. This model even learnt the Black and Asian attribute with an accuracy of $82.28\%$. By removing these two conditions, we got an accuracy of $82.43\%$. These rules are clearly affected by inter-attributes correlation. As such, the original dataset and the features used have to be carefully analysed to ensure a  fair algorithm.

\section{Sound complete verification - complementary details}
\label{appendix:Sound}

\subsection{Detailed comparison results between $\mathcal{TT}$net and $\alpha-\beta$-Crown }\label{subsec:Appendix_beta_comp}

The detailed comparison results between $\mathcal{TT}$net and $\alpha-\beta$-Crown is presented Table~\ref{tab:Appendix_beta_comp}.

\begin{table*}[htb!]
\centering
\captionof{table}{Comparison of verification strategies: usage of a general DCNN to verify with $\alpha$-$\beta$-Crown \cite{xu2020fast, wang2021beta} or using specific $\mathcal{TT}$net with general SAT verification method. The comparison is based on the 7 benchmarks from the VNN competition.}
\label{tab:Appendix_beta_comp}
\resizebox{\columnwidth}{!}{
\begin{tabular}{@{}llllccclccc@{}}
\toprule
                          &                                  &                                                        & \multicolumn{4}{c}{DNN + $\alpha$-$\beta$-Crown}                                                                                                                                                                                                           & \multicolumn{4}{c}{$\mathcal{TT}$net + Classic Verification Pipeline}                                                                                                                                                                                                  \\ \midrule
Dataset                   & Dataset Name                     & \begin{tabular}[c]{@{}l@{}}Noise \\ level\end{tabular} & \begin{tabular}[c]{@{}l@{}}Model \\ verified\end{tabular} & \begin{tabular}[c]{@{}c@{}}Verified \\ Accuracy (\%)\end{tabular} & \begin{tabular}[c]{@{}c@{}}Verification \\ time (s)\end{tabular} & \begin{tabular}[c]{@{}c@{}}Timeout \\ (\%)\end{tabular} & \begin{tabular}[c]{@{}l@{}}Model \\ verified\end{tabular} & \begin{tabular}[c]{@{}c@{}}Verified \\ Accuracy (\%)\end{tabular} & \begin{tabular}[c]{@{}c@{}}Verification \\ time (s)\end{tabular} & \begin{tabular}[c]{@{}c@{}}Timeout \\ (\%)\end{tabular} \\ \midrule
\multirow{14}{*}{CIFAR10} & \multirow{2}{*}{cifar10\_resnet} & 2/255                                                  & resnet\_2b                                                & 27                                                                & 80                                                               & 21                                                      & $\mathcal{TT}$net C                                                   & 31 (+4\%)                                                         & 0.06 ($\times$1300)                                                & 0                                                       \\
                          &                                  & 1/255                                                  & resnet\_4b                                                & 23                                                                & 183                                                              & 29                                                      & $\mathcal{TT}$net C                                                   & 38 (+9\%)                                                        & 0.03 ($\times$6100)                                                & 0                                                       \\ \cmidrule(l){2-11} 
                          & \multirow{3}{*}{cifar2020}       & 2/255                                                  & cifar10\_2\_255                                           & -                                                                 & 52                                                               & 5                                                       & $\mathcal{TT}$net C                                                   & 31                                                                & 0.06 ($\times$870)                                                & 0                                                       \\
                          &                                  & 8/255                                                  & cifar10\_8\_255                                           & -                                                                 & 57                                                               & 9                                                       & $\mathcal{TT}$net A                                                   & 23                                                                & 0.04  ($\times$1400)                                               & 0                                                       \\
                          &                                  & 2/255                                                  & convBigRELU                                               & -                                                                 & 70                                                               & 6                                                       & $\mathcal{TT}$net C                                                   & 31                                                                & 0.06  ($\times$1200)                                               & 0                                                       \\ \cmidrule(l){2-11} 
                          & \multirow{6}{*}{marabou}         & 0.012                                                  & large                                                     & -                                                                 & 139                                                              & 33                                                      & $\mathcal{TT}$net C                                                   & 30                                                                & 0.03 ($\times$4600)                                                & 0                                                       \\
                          &                                  & 0.024                                                  & large                                                     & -                                                                 & 45                                                               & 8                                                       & $\mathcal{TT}$net A                                                   & 23                                                                & 0.04  ($\times$1100)                                               & 0                                                       \\
                          &                                  & 0.012                                                  & small                                                     & -                                                                 & 226                                                              & 50                                                      & $\mathcal{TT}$net C                                                   & 30                                                                & 0.03 ($\times$7500)                                                & 0                                                       \\
                          &                                  & 0.024                                                  & small                                                     & -                                                                 & 108                                                              & 25                                                      & $\mathcal{TT}$net A                                                   & 23                                                                & 0.04  ($\times$2700)                                               & 0                                                       \\
                          &                                  & 0.012                                                  & medium                                                    & -                                                                 & 139                                                              & 33                                                      & $\mathcal{TT}$net C                                                   & 30                                                                & 0.03 ($\times$4600)                                                & 0                                                       \\
                          &                                  & 0.024                                                  & medium                                                    & -                                                                 & 44                                                               & 8                                                       & $\mathcal{TT}$net A                                                   & 23                                                                & 0.04  ($\times$1100)                                               & 0                                                       \\ \cmidrule(l){2-11} 
                          & \multirow{3}{*}{oval21}          & max. 13/255                                            & cifar\_base                                               & -                                                                 & 262                                                              & 30                                                      & $\mathcal{TT}$net A                                                   & 13                                                                & 0.5 ($\times$520)                                                & 0                                                       \\
                          &                                  & max. 13/255                                            & cifar\_wide                                               & -                                                                 & 489                                                              & 60                                                      & $\mathcal{TT}$net A                                                   & 13                                                                & 0.5 ($\times$990)                                                & 0                                                       \\
                          &                                  & max. 13/255                                            & cifar\_deep                                               & -                                                                 & 496                                                              & 60                                                      & $\mathcal{TT}$net A                                                   & 13                                                                & 0.5 ($\times$1000)                                               & 0                                                       \\ \midrule
\multirow{10}{*}{MNIST}   & \multirow{2}{*}{eran}            & 0.015                                                  & relu\_9                                                   & -                                                                 & 151                                                              & 6                                                       & $\mathcal{TT}$net C                                                   & 99                                                                & 0.07 ($\times$2000)                                                & 0                                                       \\
                          &                                  & 0.012                                                  & ffnn                                                      & -                                                                 & 58                                                               & 17                                                      & $\mathcal{TT}$net C                                                   & 99                                                                & 0.07 ($\times$830)                                                 & 0                                                       \\ \cmidrule(l){2-11} 
                          & \multirow{2}{*}{verivital}       & 0.02                                                   & Convnet                                                   & -                                                                 & 14                                                               & 0                                                       & $\mathcal{TT}$net C                                                   & 98                                                                & 0.07 ($\times$2000)                                                & 0                                                       \\
                          &                                  & 0.04                                                   & Convnet                                                   & -                                                                 & 18                                                               & 0                                                       & $\mathcal{TT}$net C                                                   & 97                                                                & 0.01  ($\times$1800)                                               & 0                                                       \\ \cmidrule(l){2-11} 
                          & \multirow{6}{*}{mnistfc}         & 0.03                                                   & net\_256x2                                                & -                                                                 & 23                                                               & 0                                                       & $\mathcal{TT}$net C                                                   & 98                                                                & 0.09 ($\times$260)                                               & 0                                                       \\
                          &                                  & 0.05                                                   & net\_256x2                                                & -                                                                 & 20                                                               & 0                                                       & $\mathcal{TT}$net C                                                   & 97                                                                & 0.02  ($\times$1000)                                               & 0                                                       \\
                          &                                  & 0.03                                                   & net\_256x4                                                & -                                                                 & 129                                                              & 20                                                      & $\mathcal{TT}$net C                                                   & 98                                                                & 0.09 ($\times$ 1430)                                             & 0                                                       \\
                          &                                  & 0.05                                                   & net\_256x4                                                & -                                                                 & 193                                                              & 33                                                      & $\mathcal{TT}$net C                                                   & 97                                                                & 0.02  ($\times$9700)                                               & 0                                                       \\
                          &                                  & 0.03                                                   & net\_256x6                                                & -                                                                 & 214                                                              & 33                                                      & $\mathcal{TT}$net C                                                   & 98                                                                & 0.09 ($\times$ 2380)                                             & 0                                                       \\
                          &                                  & 0.05                                                   & net\_256x6                                                & -                                                                 & 240                                                              & 43                                                      & $\mathcal{TT}$net C                                                   & 97                                                                & 0.02  ($\times$12000)                                              & 0                                                       \\ \bottomrule
\end{tabular}}
\end{table*}



\subsection{Global comparison between methods}

\begin{table}[]
\centering
\caption{Comparison of verifiable strategies according architectures, correctness, type of solver, verification time, verified accuracy and timeouts.}
\label{tab:verification}
\resizebox{0.8\columnwidth}{!}{\begin{tabular}{@{}lcccccc@{}}
\toprule
           & \begin{tabular}[c]{@{}c@{}}Verifiable\\ Architecture\end{tabular} & Correctness & Type of solver        & \begin{tabular}[c]{@{}c@{}}Verification \\ Time\end{tabular} & \begin{tabular}[c]{@{}c@{}}Verified \\ Acc.\end{tabular} & Timeouts \\ \midrule
$\alpha$-$\beta$-Crown \cite{xu2020fast, wang2021beta} & All ReLU DNNs                                                      & No          & Only Beta-Crown       & High                                                         & High                                                     & Many     \\
BNN \cite{jia2020efficient}       & Only BNN                                                          & Yes         & Only well crafted SAT & Very Low                                                     & Medium                                                   & Any      \\
Ours       & Only $\mathcal{TT}$net                                                        & Yes         & Any SAT solver        & Very Low                                                     & Medium                                                   & Any      \\ \bottomrule
\end{tabular}}
\end{table}

In Table~\ref{tab:verification}, we are presenting a comparison of three different approaches for the formal verification of neural networks. The first approach, $\beta$-Crown~\cite{xu2020fast, wang2021beta}, is a general method that can be applied to any ReLU-based DNN but does not provide any guarantee of correctness. The second approach is specific to BNNs and provides a guarantee of correctness but requires a well-crafted specific SAT solver. Our proposed $\mathcal{TT}$net architecture constitutes a third approach in that it can be verified using any SAT solver, while providing a guarantee of correctness. In terms of verification time, our approach is significantly faster than $\alpha-\beta$-Crown and slightly faster than BNN, except for low noise CIFAR-10 case. On the other hand, it should be noted that our method may show a lower verified accuracy, particularly in high-noise scenarios.

\subsection{Overall architecture}
\label{sec:Aann_Modeldescrption}

In this study, we considered the three architectures shown in Table~\ref{table:models}. All the paddings are set to 0.

\begin{table*}[htb!]
\centering
\caption{ \label{table:models} \textbf{Different studied model architectures with $\mathcal{TT}$net.}}
\renewcommand\arraystretch{1.2}
\resizebox{\columnwidth}{!}{
\begin{tabular}{@{}lcccccccccccc@{}}
\toprule
  \textbf{Dataset}                    & \textbf{Name} & \textbf{Layers} & \textbf{\# of blocks} & \textbf{Filters sizes}   &  \textbf{Amplification}    & \textbf{Kernels} & \textbf{Groups}    & \textbf{Strides} & \textbf{Features} & \textbf{Parameters} & \textbf{FLOPS} & \textbf{Patch size} \\ \midrule
\multirow{2}{*}{MNIST}    & Model A(6,6) & 4      & 2               & 60-48-384-48  & 1.25-8 & 3-1-2-1 & 1-1-24-24 & 3-1-2-1 & 768      & 15488      &   0.63 m & (6,6)  \\
                          & Model B(7,7) & 4      & 2               & 60-48-384-48 & 1.25-8 & 3-1-3-1 & 1-1-48-48 & 2-1-2-1 & 2352     & 30780      &   0.77 m &  (7,7)  \\
                                                    & Model C(7,7) & 4      & 2               & 144-48-384-48 & 3-8 & 3-1-3-1 & 1-1-48-48 & 2-1-1-1 & 8112     & 86580      &   0.98 m &  (7,7)  \\
                          \hline 
\multirow{2}{*}{CIFAR 10} & Model A(6,6) & 4      & 2               & 60-48-384-48 & 1.25-8 & 3-1-2-1 & 3-3-24-24 & 3-1-2-1 & 1200     & 17890      &    0.89 m &  (6,6) \\
                          & Model B(7,7) & 4      & 2               & 60-48-384-48 & 1.25-8 & 3-1-3-1 & 3-3-48-48 & 2-1-2-1 & 2352     & 30780      &   0.87 m &     (7,7)    \\
                                                                              & Model C(7,7) & 4      & 2               & 144-48-384-48 & 3-8 & 3-1-3-1 & 1-1-48-48 & 2-1-1-1 & 8112     & 86580      &   1.12 m &  (7,7)  \\
                                                                              \bottomrule
\end{tabular}}
\end{table*}

\subsection{Training  details}

\myparagraphtitle{Training method.}  We built our training method on top of \cite{jia2020efficient} project and we refer to their notations for this section. We trained the networks using the Adam optimizer \cite{kingma2014adam} for 90 epochs with a minibatch size of 128. The mean and variance statistics of batch normalization layers are recomputed on the whole training set after training finishes.

Learning rate is 0.0005. We use Projected Gradient
Descent (PGD~\cite{madry2017towards}) with adaptive gradient cancelling to train robust networks, where the perturbation bound $\epsilon$ is increased linearly from 0 to the desired value in the first 50 epochs and the number of PGD iteration steps grows linearly from 0 to 10 in the first 23 epochs.

The parameter $\alpha$ in adaptive gradient cancelling is chosen to maximize the PGD attack success rate
evaluated on 40 minibatches of training data sampled at the first epoch. Candidate 
values of $\alpha$
are between 0.6 to 3.0 with a step of 0.4. Note that $\alpha$  is a global parameter shared by all neurons. We do not use any data augmentation techniques for training. Due to limited computing resources and significant differences between the settings we considered, data in this paper are reported based on one evaluation run.

\myparagraph{Weights initialization.} 
Weights for the final connected layers are initialized from a Gaussian distribution with standard deviation 0.01, and the mask weights in BinMask are enforced to be positive by taking the absolute value during initialization.

\myparagraph{Post-tuning parameters.} 
\label{sec:Aann_Modeldescrption2}
We use a proportion ratio of 0.1 and a probability $p = 0.05$ for MNIST low noise, we double it for high noise and a proportion ratio of 0.05 and a probability $p = 0.01$ for CIFAR-10 low noise and we double it for high noise.

\myparagraph{Other hyperparameters.} 
\label{sec:Aann_Modeldescrption3}
The input quantization step s is set to be $0.61 = 0.3 \times 2 + 0.01$ for training on the MNIST dataset, and 0.064 $\approx$ 16.3/255 for CIFAR-10, which are chosen to be slightly greater than twice the
largest perturbation bound we consider for each dataset. Except for CIFAR-10 for which we double the training noise, the training noise level is equal to the testing noise level. The CBD loss is applied on MNIST high noise only and $\nu$ is set to be $5e-4$, 0 otherwise. We apply a weight decay of $1e-7$
on the binarized mask weight of BinMask. We use the encoding proposed in \cite{abio2011bdds} for the linear regression encoding into SAT. In Table~\ref{table:Results_TTDCNN}, we use Model C(7,7) except for the high noise model with CIFAR-10, we use Model A. 

\subsection{Example of an adversarial sound and complete verification using SAT}
\label{ANN:example}

Figure~\ref{fig:Example_Noise_Propagation} is an example of an adversarial sound and complete verification using SAT. $LTT_{B2}$ outputs $ (\overline{U_1} \lor U_2) \land (U_1 \lor \overline{U_1}) $ that represents the $\CNF$ relation between the unknown input $U_1$ and the unknown input $ U_2 $ according the filter SAT equation $x_1 \land \overline{x3} $ for the input $\begin{bmatrix} x_0 & x_1 \\ x_2 & x_3 \end{bmatrix} = \begin{bmatrix} 1 & U_1 \\ 0 & 0 \end{bmatrix}$. In this example, we need to encode $U_1=U_2$ in $\CNF$: as all the input/output relations have been pre-computed in $LTT_{B2}$ we simply need to give the input and save the output.


\section{Differentiable logic gates CNNs - complementary details}
\label{appendix:Differentiable}

\subsection{Training  details}
Same as for Sound and Complete Verification section, without the PGD.

\subsection{Architecture  details}

\myparagraphtitle{MNIST.} This model is composed of one LTT block of kernel size $(3,2)$ and stride 2 with no padding. The input is resized to $20*20$ before entering the LTT block. It is followed then by a linear layer of $1600$ features to $10$ classes. The final linear layer is on 4 bit.

\myparagraph{CIFAR10.} This model is composed of one LTT block multihead  of kernel size $4$ and stride 1 with no padding. The input is resized to $20*20$ before entering the LTT block. It is followed then by a linear layer of $1600$ features to $10$ classes. The final linear layer is on 1 bit. There is also a first binary CNN layer of kernel 4, sparse at 92\%.

\section{Glossary}
\label{glossary}

\begin{sidewaysfigure*}[htbp]
    \centering
   \caption{Example of an adversarial sound and complete verification using SAT with class $0$ as the correctly predicted class for the image. After the preprocessing block, only the pixel $U_1$ can switch value under the norm $l_{\infty}$ and noise level $\epsilon=0.1$. As all input/outputs have been computed and stored, the first LTT block outputs $1$ (which means that the perturbation does not influence the first LTT block), and the second one outputs two values: $U_2$ and the SAT equation that linked $U_1$ and $U_2$: $U_1 = U_2$. After the linear regression, two possible conditions can lead to an attack: either the attack leads to class $1$ or class $2$ as the final prediction. The SAT solver outputs that the first case is impossible and the second case is possible: there is an attack that leads to the prediction being switched from class 0 to class 2.}
  \label{fig:Example_Noise_Propagation}
    \includegraphics[width=1\textwidth]{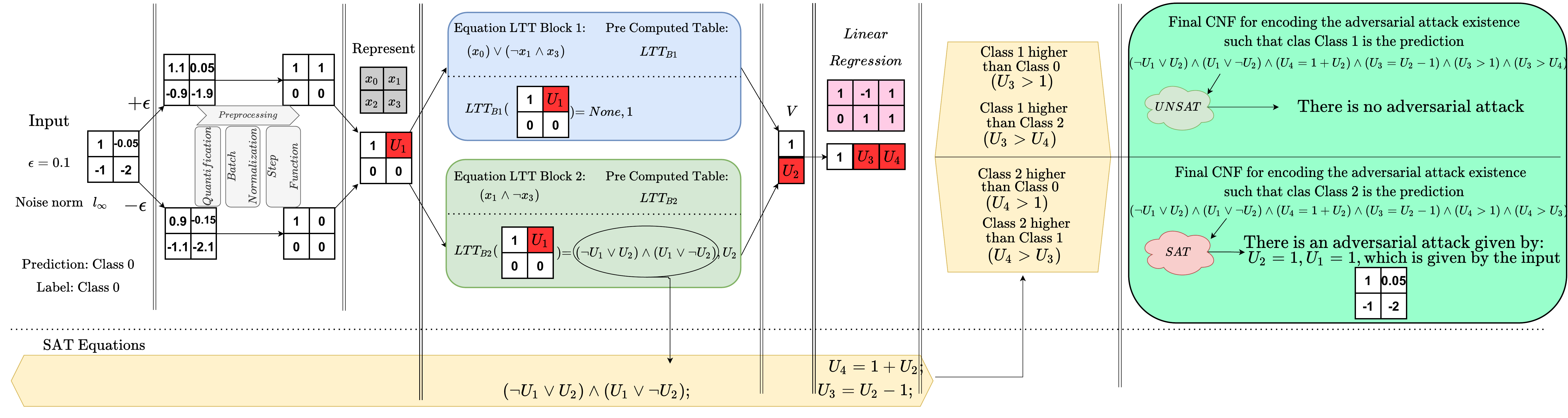}
\end{sidewaysfigure*}

\myparagraphtitle{Notations glossary:}
\begin{itemize}
\setlength\itemsep{-0.0em}
\item \textbf{$Bin$}: Heaviside step function defined as $Bin(x) = \frac{1}{2} + \frac{sgn(x)}{2}$ with  $x \in \mathbb{R}$
\item \textbf{$c$}: number of input channels
\item \textbf{$c_j = l_1 \land \cdots \land l_k$}: conjonction $j$ with $j \in \mathbb{N}$
\item \textbf{$d_j = l_1 \lor \cdots \lor l_k$}: disjunction $j$ with $j \in \mathbb{N}$
\item \textbf{$f$}: filter number
\item \textbf{$g$}: group parameter
\item \textbf{$k$}: kernel size
\item \textbf{$l_i$}: literal $i$, $l_i = x_i$ or $l_i = \overline{x_i}$  with $i \in \mathbb{N}$
\item $l_\infty$-norm: infinite norm
\item \textbf{$n$}: number of inputs of 2D-CNN corresponding of pixel size of an input patch, $n = \frac{k^2c}{g}$
\item \textbf{$p$}: padding parameter
\item \textbf{$prec$}: precondition on inputs (in the case of robustness, the $\epsilon$-ball, which is concretely in Figure~\ref{fig:Example_Noise_Propagation} the first $U_1$)
\item \textbf{$prop$}: property on outputs. In our case, the property is that the prediction should not change under the perturbation. In Figure~\ref{fig:Example_Noise_Propagation}, $\overline{prop}$ is "Class 1 higher than Class 0 and Class 1 higher than Class 2" for one of the SAT equation. 
\item \textbf{$r_{i,j} = y_i - y_j > 0$}: reified cardinality
\item \textbf{$s$}: stride parameter
\item \textbf{$sgn$}: sign function
\item \textbf{$U$}: "Unknown" (undetermined if 0 or 1). Literal $l$ in SAT equation.
\item \textbf{$V$}: vector of features obtained in the penultimate layer after the flatten operation but before the final classification layer.
\item \textbf{$x_i$}: Boolean variable $i$ with $i \in \mathbb{N}$
\item \textbf{$\overline{x_i}$}: complement of Boolean variable $x_i$ with  $i \in \mathbb{N}$
\item \textbf{$(x_1, \cdots ,x_n)$}: set of $n$ Boolean variables with  $n \in \mathbb{N}$
\item \textbf{$\Phi$}: Boolean expression
\end{itemize}

\myparagraph{Acronyms glossary:}
\begin{itemize}
\setlength\itemsep{-0.0em}
\item \textbf{BNN}: Binary Neural Network
\item \textbf{CNF}: conjunctive Normal Form. $\Phi=(d_1 \land \cdots \land d_m)$ with $m \in \mathbb{N}$
\item \textbf{DCNN}: Deep Convolutional Neural Network
\item \textbf{DNF}: disjunctive Normal Form. $\Phi=(c_1 \lor \cdots \lor c_m)$ with $m \in \mathbb{N}$
\item \textbf{DNN}: Deep Neural Network
\item \textbf{LTT}: Learning Truth Table
\item \textbf{MIP}: Mixed Integer Programming
\item \textbf{SMT}: Satisfiability Modulo Theory
\item \textbf{$\mathcal{TT}$net}: Truth Table Deep Convolutional Neural Network
\item \textbf{2D-CNN}: 2-Dimensional Convolutional Neural Network, given filter $f$, stride $s$, padding $p$, kernel size $k$, number of input channels $c$. With $(i,j) \in \mathbb{N}^{2} $, $x_X$ the $X^{th}$ pixel of the considered image patch, $(x_1^{(i,j)}, \cdots, x_{n}^{(i,j)})$ the set of $n$ pixels in the considered patch which top-left pixel is located at position $(i,j)$ in the input image.
\begin{equation*}
\Phi_{(f,s,p,k,c)}: \mathbb{R}^{n} \rightarrow \mathbb{R}\\
\end{equation*}
\begin{equation*}
(x_1^{(i,j)}, \cdots, x_{n}^{(i,j)}) 	\mapsto y_f ^{(i,j)} = \Phi_f(x_1^{(i,j)}, \cdots, x_{n}^{(i,j)})
\end{equation*}

In our specific case, a LTT block has the form: 

\begin{equation*}
(x_1^{(i,j)}, \cdots, x_{n}^{(i,j)}) 	\mapsto y_f ^{(i,j)} = \Phi^{2}_{f}(GELU(\Phi^{1}_{8}(x_1^{(i,j)}, \cdots, x_{n}^{(i,j)}))
\end{equation*}

with $\Phi^{1}_{(8f,s,0,k,c)}: \{0,1\}^n \rightarrow \mathbb{R}^8$ and $\Phi^{2}_{(f,1,0,1,8c)}: \mathbb{R}^{8} \rightarrow \{0,1\}$.
\end{itemize}

\end{document}